\documentclass[11pt]{article}

\usepackage{pcks}
\newcommand{\abs}[1]{\left\lvert#1\right\rvert}
\DeclareMathOperator*{\argmax}{argmax}
\DeclareMathOperator*{\argmin}{argmin}

\newcommand{\cA}{\mathcal{A}}
\newcommand{\cB}{\mathcal{B}}
\newcommand{\cC}{\mathcal{C}}
\newcommand{\cE}{\mathcal{E}}
\newcommand{\cF}{\mathcal{F}}
\newcommand{\cL}{\mathcal{L}}
\newcommand{\cN}{\mathcal{N}}
\newcommand{\cS}{\mathcal{S}}
\newcommand{\cV}{\mathcal{V}}
\newcommand{\disc}{\ensuremath{\delta}}
\DeclareMathOperator{\dist}{dist}
\DeclareMathOperator{\E}{\mathbb{E}}
\DeclareMathOperator{\diag}{diag}

\newcommand{\lng}{\ensuremath{\mathrm{lg}}}
\newcommand{\modulus}{\ensuremath{\gamma}}
\newcommand{\norm}[1]{\left\lVert#1\right\rVert}
\newcommand{\N}{\mathbb{N}}

\newcommand{\R}{\mathbb{R}}
\newcommand{\rk}[1]{\mbox{rk}(#1)}

\DeclareMathOperator{\supp}{Supp}

\newcommand{\snorm}[1]{\left\lVert#1\right\rVert_{\sigma,2}}
\newcommand{\supnorm}[1]{\left\lVert#1\right\rVert_{\infty}}
\newcommand{\twonorm}[1]{\left\lVert#1\right\rVert_{2}}
\newcommand{\tnorm}[1]{\left\lVert#1\right\rVert_{\tau,\infty}}
\newcommand{\tg}{\tilde{g}}
\newcommand{\tl}{\tilde{l}}
\newcommand{\tx}{\tilde{x}}

\newcommand{\vin}{v_\mathrm{src}}
\newcommand{\vter}{v_\mathrm{tgt}}

\newtheorem{theorem}{Theorem}[section]
\newtheorem*{app_thm*}{Theorem}
\newtheorem{lemma}[theorem]{Lemma}
\newtheorem*{app_lemma*}{Lemma}
\newtheorem{prop}[theorem]{Proposition}
\newtheorem*{app_prop*}{Proposition}
\newtheorem{coro}[theorem]{Corollary}
\newtheorem*{app_coro*}{Corollary}
\newtheorem{definition}[theorem]{Definition}
\newtheorem{assumption}{Assumption}
\theoremstyle{remark}
\newtheorem{remark}[theorem]{Remark}
\newtheorem{example}{Example}[section]

\newcounter{savedexample}
\makeatletter
\newenvironment{continuedexample}[1]{%
  \begingroup
  \setcounter{example}{\value{savedexample}}%
  \begin{example}%
}{%
  \end{example}%
  \endgroup
}
\makeatother

\definecolor{orange}{rgb}{1,0.5,0}
\definecolor{deepgreen}{RGB}{0,168,107}

\allowdisplaybreaks

\title{Heuristics for Combinatorial Optimization via Value-based Reinforcement Learning: A Unified Framework and Analysis}

\author{%
  Orit Davidovich\textsuperscript{1,}\thanks{Equal contribution.}
  , Shimrit Shtern\textsuperscript{2,*}
  , Segev Wasserkrug\textsuperscript{1}
  , Nimrod Megiddo\textsuperscript{1}\\[0.5ex]
  \textsuperscript{1}IBM Research\\
  \textsuperscript{2}Faculty of Data and Decision Sciences, Technion -- Israel Institute of Technology\\[0.5ex]
  \texttt{\small\textbf{\{orit.davidovich@,segevw@il.,megiddo@us.\}ibm.com}}\\\texttt{\small\textbf{shimrits@technion.ac.il}}
}

\begin{document}

\maketitle

%%%%%%%%%%%%%%%%%%%%%%%%%%%%%%%%%%%%
%%%%%%%%%%%%%%%%%%%%%%%%%%%%%%%%%%%%

\begin{abstract}
    Since the 1990s, considerable empirical work has been carried out to train statistical models, such as neural networks (NNs), as learned heuristics for combinatorial optimization (CO) problems. When successful, such an approach eliminates the need for experts to design heuristics per problem type. Due to their structure, many hard CO problems are amenable to treatment through reinforcement learning (RL). Indeed, we find a wealth of literature training NNs using value-based, policy gradient, or actor-critic approaches, with promising results, both in terms of empirical optimality gaps and inference runtimes. Nevertheless, there has been a paucity of theoretical work undergirding the use of RL for CO problems. To this end, we introduce a unified framework to model CO problems through Markov decision processes (MDPs) and solve them using RL techniques. We provide easy-to-test assumptions under which CO problems can be formulated as equivalent undiscounted MDPs that provide optimal solutions to the original CO problems. Moreover, we establish conditions under which value-based RL techniques converge to approximate solutions of the CO problem with a guarantee on the associated optimality gap. Our convergence analysis provides: (1) a sufficient rate of increase in batch size and projected gradient descent steps at each RL iteration; (2) the resulting optimality gap in terms of problem parameters and targeted RL accuracy; and (3) the importance of a choice of state-space embedding. Together, our analysis illuminates the success (and limitations) of the celebrated deep Q-learning algorithm in this problem context.
\end{abstract}

%%%%%%%%%%%%%%%%%%%%%%%%%%%%%%%%%%%%
%%%%%%%%%%%%%%%%%%%%%%%%%%%%%%%%%%%%

\section{Introduction}

A combinatorial optimization (CO) problem is an optimization problem where one optimizes a given function $g$ dependent on a parameter $c$ over a finite set $\Pi$ of feasible solutions ($\abs{\Pi} < \infty$). Ideally, one looks for the optimal solutions $x^*$,
\begin{equation}
\label{eq:cop}
    x^* \in \argmax_{x \in \Pi} g(x;c).
\end{equation}
CO problems are ubiquitous in many logistic, transportation and scheduling applications (see \cite{paschos2014applications} and references therein). This family of problems includes NP-Hard problems such as the Knapsack problem, the traveling salesman problem, the vehicle routing problem, and their variations, as well as polynomially solvable problems such as the shortest path problem and network flow problems.

The NP-hardness of some CO problems implies that, unless $P=NP$, there does not exist a polynomial-time algorithm to solve them exactly for the general case. 
Thus, while off-the-shelf solvers utilizing classical approaches, such as branch and bound, can provide exact solutions to CO problems, these solvers do not perform well when problem size increases significantly. In particular, these solvers may fail to obtain a feasible solution with a sufficient optimality guarantee in a reasonable amount of time.

Extensive research has been carried out into designing or learning algorithms that may provide sufficiently good {feasible} solutions to these problems, such as polynomial-time approximation algorithms~\cite{williamson2011design} that provide guarantees on the quality of the solution obtained or search heuristics and meta-heuristics (see e.g., \cite{colorni1996heuristics,hertz2003guidelines} and references therein), which do not provide such guarantees but perform well in practice. These two approaches address the issues of running time and storage as the problem scales up, paying a price in the quality of {feasible} solutions obtained (optimality gap). There are many known designed heuristics and approximation algorithms for specific CO problems such as the traveling salesman problem.
However, these methods are designed by experts per problem type, which constitutes their main drawback. For this reason researchers turned to statistical modeling to learn heuristics. Considerable empirical work has been carried out since the 1990s to train statistical models, like neural networks (NNs) \cite{hopfield_neural_1985}, to solve CO problems or incorporate these tools with standard operations research methodologies to provide improved heuristics \cite{Bengio2021Survey}.

In the recent decades, reinforcement learning (RL) frameworks have been combined with NNs to solve CO problems by treating CO problems as sequence generation tasks (see, for example, \cite{Mazyavkina2021Survey,chung_neural_2025} and references therein). Specifically, a variety of approaches rely on transformer architectures to construct a stochastic solver that learns heuristics to solve CO problems and show promising empirical performance, both in terms of optimality gap as well as in terms of time needed to generate a solution, for various families of CO problems \cite{Khalil2017,Drori2020,Bello2017NCO,Barrett2020CO,Kool2019Attention}. 

The {latent} starting point of these RL methods, however, is a Markov decision process (MDP) formulation tailored to each problem type. There is scarce research on characterizing the structure needed for a general CO problem to be formulated as an MDP with guaranteed convergence to an {optimal} value function $V^*$ that maps to the optimal solution of the CO problem. Moreover, to the best of our knowledge, there is no general framework to translate a given CO problem to such an MDP. In this work, we provide such characterization and corresponding framework {based on Karp and Held's work \cite{Karp1967Programming}}. Moreover, this framework allows us to translate any approximate of $V^*$ to a feasible solution of the CO problem, with optimality gap derived directly from the distance of the approximate value function to the optimal one.

To learn a stochastic solver from the MDP associated with the CO problem, some value-based RL approaches attempt to learn probability distributions over the space of state-action-state transitions associated with promising solution heuristics using approximate value functions of the MDP as a guide. Specifically, an exploration-exploitation framework is used to simultaneously learn {a useful}
% the ``optimal'' 
probability distribution and retrieve an optimal value function {approximate} under this distribution. Moreover, due to the high dimensionality of the state-action space {concerned}, the {optimal} value function is approximated by a parameterized family of functions (e.g., NNs), and the per-iteration computational cost is reduced by sampling, leading to methods such as deep Q-learning (DQL) \cite{Mnih2015DQN}. In a policy-gradient RL algorithm such as REINFORCE, as used for example in \cite{Kool2019Attention,Berto2023RL4CO}, the algorithm generates complete solution rollouts for each sampled CO instance and updates the policy based on entire trajectories, rather than on individual one-step transitions as in DQL. In this work we focus on transition-based value methods, such as DQL, that operate on sampled transitions. We do not analyze policy-gradient approaches such as REINFORCE, PPO, or A2C.

Specifically, in this work, we focus on fitted Q-iteration (FQI) \cite{Riedmiller2005FittedQ}, which can be thought of as a subroutine of DQL for a given learned distribution. We first show that, due to the deterministic nature of the underlying CO MDP, one can switch instead to fitted value iteration (FVI) \cite{Munos2008FVI}, which {reduces memory requirements} by handling value functions over the state space (rather than Q-functions on the state-action space). FVI fits in the literature into the more general framework of approximate policy and value iteration. 
{We provide a full analysis of FVI, including conditions for convergence, sample, and iteration complexity.}

FVI accumulates errors, stemming either from the restriction to a parameterized family of value functions, i.e., approximation errors, or from the {loss minimization} procedure, i.e., estimation errors. Error propagation in FVI and, more generally, in approximate value iteration (AVI) and approximate policy iteration (API), has been studied in the literature. In \cite{Fan2020DQL}, the authors derive bounds on the value function error in terms of the maximum per-iteration error, but do not provide sample or iteration complexity results. By contrast, \cite{Munos2008FVI} analyzes error propagation in terms of both sample and iteration complexity, under the assumption of no approximation error. In this work, we take a different approach to analyzing FVI. Specifically, we establish an equivalence between FVI and applying sample average approximation (SAA) to the {loss established at each} iteration of {a projected variant} of value iteration. This equivalence enables us to decouple the cumulative error into two independent components: the approximation error of the optimal value function, arising from the use of projected value iteration, and the estimation error accumulated from applying SAA to distance minimization at each projection step. 

While it has been shown that projected value iteration does not converge {for the general MDP} \cite{Farias2000FixedPoints}, we introduce an assumption to guarantee its convergence and use it to derive a bound on the approximation error. Furthermore, we provide empirical evidence, based on our modified analysis, demonstrating that our assumption holds in practice and that the resulting bound is tight.

Finally, leveraging existing SAA consistency theory of \cite{shapiro2021lectures}, we derive sample and iteration complexity bounds for the per-iteration estimation error, and consequently for FVI. In contrast to prior work, we further assume that each SAA subproblem is solved approximately via the projected gradient method, leading to a first-order iteration complexity characterization of FVI. 

%%%%%%%%%%%%%%%%%%%%%%%%%%%%%%%%%%%%
%%%%%%%%%%%%%%%%%%%%%%%%%%%%%%%%%%%%

\paragraph{Contribution}

We view the contribution of this paper as twofold. First, we provide a unified framework that translates CO problems into equivalent undiscounted MDPs and establish conditions under which applying value iteration to the derived MDP converges to an optimal value function from which an optimal solution to the original CO problem can be obtained. Specifically, we utilize the general framework of Karp and Held \cite{Karp1967Programming}, originally designed for discrete problems with possibly infinite feasible sets, to the CO setting. We show that a CO problem can be associated with a finite-state, finite-depth, deterministic {undiscounted} MDP possessing a unique absorbing state. This construction enables us to specify conditions on both the CO problem and the choice of constructed MDP that permit the application of a modified version of convergence results in \cite{Tseng1990Convergence}. Consequently, we prove that the MDP admits an optimal value function corresponding to optimal solutions of the CO problem, which can be obtained through the VI procedure. Using this unified translation framework, we further show that any $\epsilon$-approximation of the CO MDP’s optimal value function can be converted into a feasible and $O(\epsilon)$-optimal solution to the original CO problem.

Second, we show how {value-based} RL methods can be used to approximate the {optimal} value function for the CO MDP and analyze their complexity. Our analysis focuses on FVI, whose applicability stems from the deterministic structure of the underlying MDP. We propose a novel interpretation of FVI as projected value iteration with SAA applied at each iteration. Under this interpretation, we present the first error-propagation analysis that clearly separates the approximation error of FVI, stemming from restricting the {hypothesis} class of value functions, from the estimation error introduced by sampling in {loss minimization}. We do not assume that the optimization problem solved at each FVI iteration is solved exactly; instead, we assume it is solved approximately using first-order methods. This allows us to analyze the sample and first-order iteration complexity of FVI required to achieve a prescribed estimation error.

Taken together, these two contributions establish a comprehensive framework for addressing CO problems with RL-based methods and offer new insights into the conditions under which these methods are both applicable and effective.

\begin{figure}
  \centering
  \resizebox{\linewidth}{!}{
\begin{tikzpicture}[
  node distance=12mm and 16mm,
  every node/.style={font=\sffamily},
  box/.style={draw=black,very thick,rounded corners=4pt,minimum width=28mm,minimum height=10mm,align=center,fill=black!20,text=black},
  smallbox/.style={draw=black,very thick,rounded corners=4pt,minimum width=22mm,minimum height=8mm,align=center,fill=black!20,text=black},
  ellipseBox/.style={draw=black,very thick,ellipse,minimum width=15mm,minimum height=14mm,align=center,fill=black!20,text=black},
  hex/.style={draw=black,very thick,regular polygon,regular polygon sides=6,minimum width=40mm,minimum height=5mm,inner sep=2pt, align=center,fill=black!20,text=black},
  circ/.style={draw=black,very thick,circle,minimum size=1mm,align=center,fill=white,text=black},
  arrow/.style={-Latex,black,very thick},
  dashedbox/.style={draw=black,dashed,rounded corners=4pt,very thick,inner sep=8pt}
]

\node[box,anchor=west] (CO) at (-9,0.2) {CO (DDP)};

% Top dashed group
\begin{scope}[on background layer]
\node[dashedbox,minimum width=19.5cm,minimum height=2.55cm,anchor=north west,fill=gray!10] (topgroup) at (-8,4.4) {};
\end{scope}
\node[font=\sffamily\bfseries,text=black,anchor=north] at (topgroup.north) {Unified Framework from CO to MDP \S\ref{sec:co_as_mdp}};
% Top row nodes: SDP -> MDP -> VI (ellipse)
\node[smallbox,anchor=west] (SDP) at (-5,2.7) {SDP};
\node[smallbox,anchor=west] (MDP) [right=40mm of SDP] {MDP};
\node[ellipseBox,anchor=west] (VI) [right=40mm of MDP] {VI};
% Bottom dashed group
\begin{scope}[on background layer]
\node[dashedbox,minimum width=19.5cm,minimum height=3.7cm,anchor=north west,fill=gray!10] (botgroup) at (-8,-1.5) {};
\end{scope}
\node[font=\sffamily\bfseries,text=black,anchor=south] at (botgroup.south) {Solution through RL
% +NN 
\S\ref{sec:rl}
};
% bottom row nodes: DQL -> FQI -> FVI -> SAA+PGD -> PVI
\node[ellipseBox,anchor=west] (DQL) at (-6,-3.1) {DQL};
\node[ellipseBox,anchor=west] (FQI) [right=26mm of DQL] {FQI};
%\node[ellipseBox,anchor=west] (FVI_OLD) [right=16mm of FQI] {FVI};
\node[ellipseBox,anchor=west] (SAAPGD) [right=28mm of FQI] {\shortstack{SAA +\\ PGD}};
\node[ellipseBox,anchor=west] (PVI) [right=18mm of SAAPGD] {\shortstack{Projected\\ VI}};

\begin{scope}[on background layer]
\node[
    draw,
    smallbox,
    fit=(PVI)(SAAPGD),
    inner sep=16pt,
    draw=black,
    fill=gray!20,
](FVI) {};
\end{scope}
\node[anchor=north] at (FVI.north) {FVI};

%drawing the arrow on top
\draw[arrow] (CO.north) |- (SDP.west)node[above,xshift=-30pt,font=\small,text=black] {\shortstack{Karp-Held
% \\Theorem~\ref{thm:KH}
}};

\draw[arrow] (SDP) -- node[above,font=\small,text=black] {\shortstack{Assumptions~\ref{assum:x_extension}--\ref{assum:s_infinity} \\ Theorem \ref{thm:tseng_conditions}}} (MDP);

\draw[arrow] (MDP) -- node[above,font=\small,text=black] { Corollary \ref{coro:co_mdp_equiv}} (VI);

% Approximation scheme arrow coming from VI to right
\draw[arrow] (VI.east) -- ++(20mm,0) node[midway,above,font=\small,text=black,align=center] {\shortstack{Approximation\\ Scheme
%\\ Definition~\ref{def:approx_scheme}
}} |- (PVI.east);

% Middle: CO (left) and Approximated Value Function (right)
\node[hex,xscale=1.2,yscale=0.6] (ApproxVal)  at (0.4,0.2) {};
\node at (ApproxVal.center){\shortstack{Approximated \\ Value \\ Function}};

\draw[arrow] (ApproxVal.west)   -- (CO.east) node[midway,above,font=\small,text=black] {\shortstack{Feasible Solution \\
Assumption~\ref{assum:partial_to_feasible}\\
Theorem \ref{thm:approximation}}};

% small connector and circled cross for approximation error
\node[circ] (cross) at (7,0.2) {+};
\draw[arrow] (cross.west)--(ApproxVal.east);

% vertical arrow from PVI (lower right) to cross
\path let 
  \p1 = (SAAPGD.north),
  \p2 = (cross.south)
in
  coordinate (mid) at (\x1, {(\y1+\y2)/2});

\draw[arrow] (SAAPGD.north) -- (mid)node[above,right,yshift=12pt,xshift=-4pt, font=\small,text=black] {\shortstack{Estimation Error\\Theorem~\ref{thm:fvi}}} -| (cross.south); 

% arrow from PVI to cross (approximation error path)
\draw[arrow] (PVI.north) |-(cross.east)node[right,yshift=-5pt,xshift=1mm,font=\small,text=black]{\shortstack{Approximation Error\\
Assumption~\ref{assum:contraction}\\
Proposition~\ref{prop:pvi}}};

% arrows bottom
\draw[arrow] (DQL) -- node[above,font=\small,text=black] {\shortstack{Given sampling\\ probability%Oracle $p_\sigma$
% Single\\ iteration
}} (FQI);
\draw[arrow] (FQI) -- node[above,font=\small,text=black] {\shortstack{CO MDP \S\ref{sec:bias_variance}
% Lemma~\ref{lem:bias_var}+\\deterministic\\ transitions
}} (FVI);
%\draw[arrow] (SAAPGD) -- (FVI);
\draw[arrow] (PVI) -- node[above,font=\small,text=black] {Sampling} (SAAPGD);

% arrow from CO down to DQL (left vertical)
\draw[arrow] (CO.south) |- (DQL.west);

\end{tikzpicture}
}%
  \caption{Paper overview}%
  \label{fig:overview}
\end{figure}
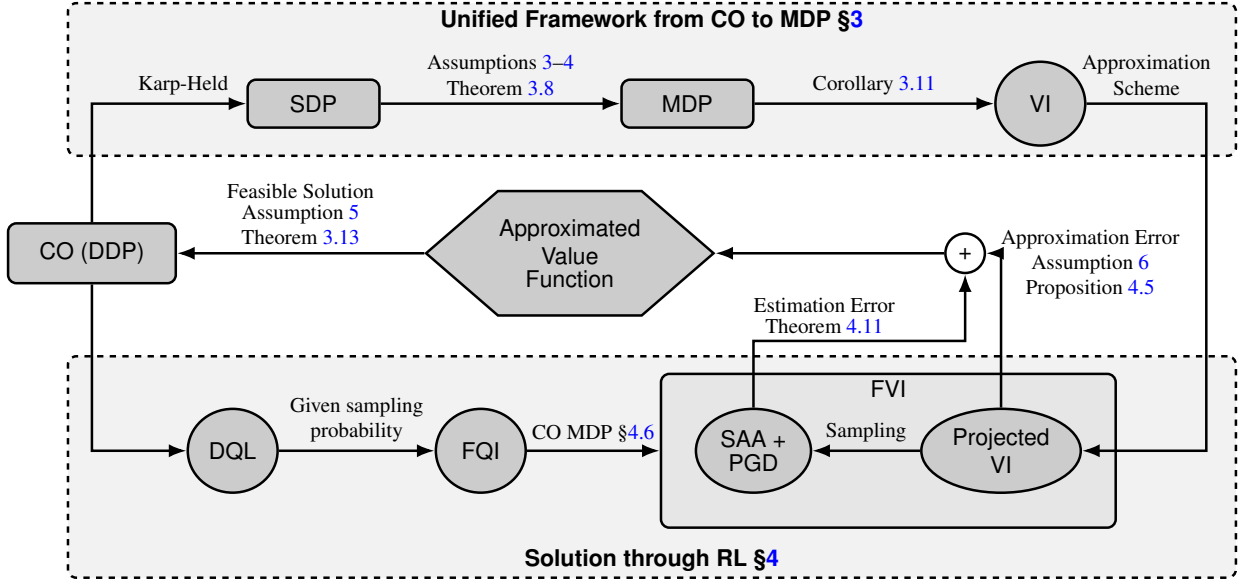

\paragraph{Overview}

Figure~\ref{fig:overview} provides an overview of the paper.
Following our two main contributions, this paper is split into two major parts: \S\ref{sec:co_as_mdp} is dedicated to the translation from CO problems to MDPs, while \S\ref{sec:rl} is dedicated to RL-based methods. Following a brief review of Karp-Held~\S\ref{sec:ddp}--\ref{sec:KHT} and MDP~\S\ref{sec:mdp}--\ref{sec:undiscounted} theory sufficient for this work, we provide our main results for the first part of the paper in \S\ref{sec:co_repn}, while \S\ref{sec:epsilon_approx} establishes how they apply to $\epsilon$-approximations. We finish this part by spelling out the MDP formulations of standard CO examples \S\ref{sec:TSP}--\S\ref{sec:SPP} resulting from our framework. Turning to part \S\ref{sec:rl}, we begin by laying the groundwork for the use of parametrized value functions \S\ref{sec:approx_scheme}, and then introduce projected value iteration \S\ref{sec:pvi} in this context, which allows us to isolate out the approximation error. Now we can introduce estimation procedures \S\ref{sec:pvi} on top of that. As we turn to concrete RL-based methods from the literature, we justify our focus on SAA using bias-variance decomposition \S\ref{sec:bias_variance}. We conclude this second part with our main analytical results for FVI in \S\ref{sec:saa}. Finally, we provide a summary and outlook of future work in \S\ref{sec:summary}. Proofs for the results stated in parts \S\ref{sec:co_as_mdp} and \S\ref{sec:rl} are provided in Appendix~\ref{app:proofs_sec3} and \ref{app:proofs_sec4} respectively.

%%%%%%%%%%%%%%%%%%%%%%%%%%%%%%%%%%%%
%%%%%%%%%%%%%%%%%%%%%%%%%%%%%%%%%%%%

\subsection{Mathematical Notation}

%%%%%%%%%%%%%%%%%%%%%%%%%%%%%%%%%%%%
%%%%%%%%%%%%%%%%%%%%%%%%%%%%%%%%%%%%

\subsubsection{Set Notation}
\label{sec:set_notation}

We will use either script letters ($\cA, \mathcal{B}$, etc.) or capital Greek letter ($\Theta, \Pi$, etc.) to denote abstract sets. The product $\cA \times \cdots \times \cA$ for $n$ copies of $\cA$ is denoted by $\cA^n$. For two sets $\mathcal{X}$ and $\mathcal{Y}$ we will denote the power set $\mathcal{Y}^{\mathcal{X}} \triangleq \Set{f: \mathcal{X} \rightarrow \mathcal{Y}}$. For two disjoint sets $\mathcal{X},\mathcal{Y} \subseteq \mathcal{Z}$, $\mathcal{X} \cap \mathcal{Y} = \emptyset$, we will use the notation $\mathcal{X} \sqcup \mathcal{Y}$ for their disjoint union.
As for specific sets, we will use $\R$ for the set of real numbers and 
$\langle n \rangle \triangleq \{0,\ldots,n-1\} \subset \N$ for the aforementioned subset of the integers $\N$ (including zero).
For a metric space $\left(\mathcal{Z},\dist\right)$ and $\mathcal{X} \subseteq \mathcal{Z}$, the distance between a point $z \in \mathcal{Z}$ and $\mathcal{X}$ is denoted by $\dist(z,\mathcal{X}) \triangleq \inf_{x \in \mathcal{X}} \dist(z,x)$. The size of a finite set $\cA$ will be denoted $\abs{\cA}$. We denote a ball in $\mathcal{Z}$ of radius $\rho$ centered at $z$ by $\mathcal{B}_\rho(z)$.

%%%%%%%%%%%%%%%%%%%%%%%%%%%%%%%%%%%%
%%%%%%%%%%%%%%%%%%%%%%%%%%%%%%%%%%%%

\subsubsection{Equivalence Relations}
\label{sec:equivalence_relations}

Here, we let $\cA$ denote a finite set, which we will consider as a finite alphabet. We will denote the set of all finite strings in the alphabet $\cA$ by $\cA^*$. The latter has a distinguished identity element $e\in \cA^*$ with respect to concatenation, i.e., the empty string. Note that $e \notin \cA$. We let $\lng(x)$ denote the length of a string $x\in \cA^*$, in particular, $\lng(e)=0$. 

Consider an equivalence relation $\sim$ defined over $\cA^*$. An equivalence relation induces a partition of $\cA^*$. We will use $\mathcal{A}/\sim$ to denote the derived set of equivalence classes and $[x]$ to denote the equivalence class of $x \in \cA^*$. An equivalence relation $\sim$ defined on $\cA^*$ is of {\em finite rank} if it has finitely many equivalence classes. One equivalence relation {\em refines} another when each equivalence class of the former is contained in an equivalence class of the latter. In that case, we say that the first equivalence relation is a {\em refinement} of the other. An equivalence relation $\sim$ defined on $\cA^*$ is a {\it right congruence} if, for any pair $x,y\in \cA^*$, $x \sim y$ implies $xu \sim yu$ for any $u \in \cA^*$.

%%%%%%%%%%%%%%%%%%%%%%%%%%%%%%%%%%%%
%%%%%%%%%%%%%%%%%%%%%%%%%%%%%%%%%%%%

\section{Prologue: Sequence Generation}
There is a natural sense in which we can think of solving a CO problem \eqref{eq:cop} as a sequence generation task. For example, in the traveling salesman problem with $d$ cities each city $a_i$ can be thought of as a distinct token drawn from a finite set $\cA$ of possible tokens, i.e., $\cA=\langle d \rangle $ represents all cities in a complete route. A string of tokens $x_0 \cdots x_{l-1}$ represents a route (not necessarily complete or efficient). To generate routes, especially complete and non-repetitive ones, we can look for a way to sequentially generate the next token, or city, $x_l$, following the current (partial) route,
\begin{equation*}
    x_l \mid x_0 \cdots x_{l-1}, \quad x_0,\ldots,x_l \in \cA,
\end{equation*}
where $x_0=0$ is chosen to be the home base. Generation should clearly depend on the parameters of the problem, namely, on the city map denoted by $c$, which can be encoded using a statistical model $q_\theta$ parameterized by weights $\theta\in\Theta$. The next token is then derived from a conditional probability distribution provided by a decoder $h_\theta$,
\begin{equation*}
    x_l \sim h_\theta\left(x_l | x_0 \cdots x_{l-1};q_\theta(c)\right),
\end{equation*}
where $\theta \in \Theta$ now represents the combined weights of the model in this encoder-decoder-type setting. Multiplying the next-token generation probabilities, we end up with a $p_\theta$ that we can use to generate solutions sequentially,
\begin{align}
    p_\theta(x;c) \triangleq \prod_{l=1}^d h_\theta\left(x_l | x_0 \cdots x_{l-1};q_\theta(c)\right), \quad x = x_0 \cdots x_d.
\label{eq:transformer_solver}
\end{align}

We can show that, in theory at least, a perfect $p^*(\cdot;c)$ does exist, meaning one that can be used to generate an optimal solution. Specifically, we can show there exists a function $V^*$ defined on sequences (e.g., routes) -- we call it the optimal value function and justify this terminology later on -- that factors through equivalence classes of routes, namely, $V^*(\cdot ; c) : \cA^* / \sim \rightarrow \R$, and another function $r$ defined on routes and cities -- we call it the immediate reward function -- that factors similarly, $r(\cdot,\cdot; c) : \cA^* / \sim \times \cA \rightarrow \R$, which we can put together to form a perfect $p^*(\cdot;c)$ using softmax. 

\begin{theorem}[Informal]
\label{thm:informal}
    There exists an optimal value function $V^*(\cdot ; c): \cA^* / \sim \rightarrow \R$ and an immediate reward function $r(\cdot,\cdot; c) : \cA^* / \sim \times \cA \rightarrow \R$ such that the following 
    \begin{equation*}
    \label{eq:factorization}
        p^*(x;c) = \prod_{l=1}^d h^*\left(x_l | x_0 \cdots x_{l-1};c\right), \quad x = x_0 \cdots x_d
    \end{equation*}        
    will generate an optimal solution using Beam Search \cite{Reddy1977Beam,Vaswani2017Attention,Janner2021RL} of width $B=1$ for
    \begin{align*}
    \label{eq:softmax}
         h^*\left(x_l \mid x_0 \cdots x_{l-1}; c\right) = 
        \frac{e^{r([x_0 \cdots x_{l-1}], x_l;c) + {V^*}([x_0 \cdots x_l];c)}}{\sum_{x_l' \in \cA} e^{r([x_0 \cdots x_{l-1}], x_l';c) + {V^*}([x_0 \cdots x_{l-1}x_l'];c)}}.
    \end{align*}
\end{theorem}

We will see, that this procedure arises naturally in cases where the CO problem is equivalent to an undiscounted MDP with appropriate states, actions, transitions, and immediate rewards, and present a framework for this translation. The Bellman functional equations of this MDP will then guarantee the generation of an optimal solution to the CO problem using Beam Search with width $B=1$, i.e., at each stage $l = 1,\ldots,d$, generate $x_l \in \cA$ by maximum likelihood. The reward structure would need to be designed so that generating an infeasible extension from a feasible partial solution -- in the case of the traveling salesman problem, a partial route without repeated cities -- would incur significantly low reward.

By standard MDP theory, finding the optimal value function can be done through value iteration, which is a successive applications of the Bellman map applied to any initial value function, defined here as a real-valued function on $\cA^* / \sim$. Since the equivalent MDP to the CO problem is undiscounted, its Bellman map is undiscounted: we do not discount future rewards by some factor smaller than one -- we care about reward now just as much as we care about reward later.

While value iteration is known to converge for discounted MDPs, convergence is not guaranteed for undiscounted MDPs. Nevertheless, we will show that, owing to the specific structure induced by the CO problem, applying value iteration to the corresponding undiscounted MDP does converge to the optimal value function $V^*$. Although this above convergence result is theoretically encouraging, it has limited practical relevance since the corresponding MDP generally possesses an enormous state space -- exponential in $d$ for the traveling salesman problem. This renders the use of value iteration computationally impracticable.

Standard RL practice informs us that we should aim for the best approximate to $V^*$ over some hypothesis class parametrized by a set of weights $\Theta$. {A well-known and widely used value-based RL algorithm is DQL introduced by Mnih et al. in \cite{Mnih2015DQN}. %[I re-introduced this sentence. It is important to stress the centrality of DQL in RL to amplify the impact of our work]}
It has been used in the literature to solve CO problems \cite{Khalil2017,Drori2020}. The DQL routine incorporates sampling both to explore the space of MDP transitions as well as to approximate value iteration, which complicates theoretical analysis. In this paper, we therefore focus on a single iteration of this algorithm, in which the sampling probability is fixed and aim to understand whether and why it should yield high-quality solutions for CO problems, as empirically observed in previous works. 

%%%%%%%%%%%%%%%%%%%%%%%%%%%%%%%%%%%%
%%%%%%%%%%%%%%%%%%%%%%%%%%%%%%%%%%%%

\section{CO as an MDP}
\label{sec:co_as_mdp}

In this section, we provide a unified framework for representing CO problems as undiscounted MDPs, amenable to solution through VI. Specifically, we identify the assumptions required for CO problems to be represented as such. To this end, we first present an intermediary representation of a CO problem as discrete decision process (DDP). We use the theory established by Karp and Held \cite{Karp1967Programming} for conditions under which a DDP can be transformed into an additive sequential decision processes (SDP) with states defined by a refinement of the DDP equivalence sets. Next, using the work of Tseng \cite{Tseng1990Convergence}, we establish conditions under which the resulting additive SDP can be transformed into a deterministic undiscounted MDP with a contractive Bellman operator under an appropriate choice of norm. Finally, VI for this MDP converges to the optimal value function, which admits an optimal policy for the underlying CO.

% \OD{Condense?}

%%%%%%%%%%%%%%%%%%%%%%%%%%%%%%%%%%%%
%%%%%%%%%%%%%%%%%%%%%%%%%%%%%%%%%%%%

\subsection{Preliminaries} 

We begin by reviewing existing definitions and results that form the theoretical foundation of our unified framework. We first revisit Karp and Held’s theory \cite{Karp1967Programming} and highlight its distinctive properties in the context of CO problems. We then review the concept of undiscounted MDPs and extend Tseng’s convergence theory for this setting \cite{Tseng1990Convergence}.

%%%%%%%%%%%%%%%%%%%%%%%%%%%%%%%%%%%%
%%%%%%%%%%%%%%%%%%%%%%%%%%%%%%%%%%%%

\subsubsection{Discrete Decision Processes}
\label{sec:ddp}

Let $\cC$ denote a parameter set. A $\cC$-parametrized DDP \cite{Karp1967Programming} consists of a tuple $\left( \cA, \Pi, g \right)$ such that $\cA$ is a finite set, $\emptyset \neq \Pi \subseteq \cA^*$, and $g: \Pi \times \cC \rightarrow \R$ is a $\cC$-parametrized objective function. The subset $\Pi$ is viewed as the subset of feasible solutions. A $\cC$-parametrized DDP is {\em finite} when $\abs{\Pi} < \infty$.

\begin{definition}
\label{def:dpi}
    The {\it depth} of a $\cC$-parametrized DDP is defined via $d_\Pi \triangleq \max_{x \in \Pi} \lng(x)$
    which may assume the value $\infty$. 
\end{definition}

\begin{definition}
\label{def:simPi}
A $\cC$-parametrized DDP is associated with a unique equivalence relation $\sim_\Pi$ on $\cA^*$ defined via $x \sim_\Pi y$ iff $\forall w \in \cA^*: xw \in \Pi \iff yw \in \Pi$.
\end{definition}

A $\cC$-parametrized DDP determines a $\cC$-parametrized family of discrete optimization problems 
\begin{equation*}
\label{eq:ddp_problem}
    \max_{x \in \Pi} g(x;c), \quad c\in\cC.
\end{equation*}    
A $\cC$-parametrized family of CO problems is derived from a finite $\cC$-parametrized DDP. 

The equivalence relation $\sim_\Pi$ in Definition~\ref{def:simPi} has two distinguished equivalence classes:
\begin{align}
    s_e & = \{e\}; \label{eq:state_e} \\
    s_\infty & = \Set{x \in \cA^* | \nexists w \in \cA^*: xw \in \Pi}. \label{eq:state_infinity} 
\end{align}
In CO the corresponding DDP is finite. Thus $s_\infty$ is non-empty with $\Set{x \in \cA^* \mid \lng(x) > d_\Pi} \subseteq s_\infty$ resulting in $\sim_\Pi$ {being of} finite rank. {Under Karp and Held’s more general formalism, $s_\infty$ does not appear, as it may be empty, and $\sim_\Pi$ is not necessarily of finite rank.} 

\begin{example}[Knapsack Problem]
\label{ex:ksp}
    Let $\cC \subseteq \R^d$ be a parameter set, and let $w \in \R^{m \times d}_{\geq 0}$ and $b \in \R^m_{\geq 0}$. The (multi-constrained, bounded) knapsack problem (KSP) with multiple budget constraints associated with $c \in \cC$ is given by the following liner integer programming problem:
    \begin{align}
    \label{eq:KSP}
        \max_{x \in \langle n \rangle^d} & \sum_{j=1}^d c_j x_j \\
        \mbox{s.t.} & \sum_{j=1}^d w_{ij} x_j \leq b_i, \quad i=1,\ldots,m.\nonumber
    \end{align}
    Note that the constraints are unaffected by the parametrization $\cC$.
    The KSP \eqref{eq:KSP} has a $\cC$-parametrized DDP formulation $\left( \cA, \Pi, g \right)$ with $\cA = \langle n \rangle$, $\Pi$ the finite set of feasible solutions, i.e., all $x \in \langle n \rangle^d \subset \cA^*$ 
    that satisfy the constraints in (8), and $g(x;c) = \sum_{j=1}^d c_j x_j$. 
\end{example}

%%%%%%%%%%%%%%%%%%%%%%%%%%%%%%%%%%%%
%%%%%%%%%%%%%%%%%%%%%%%%%%%%%%%%%%%%

\subsubsection{Karp-Held Theory}
\label{sec:KHT}

Karp-Held theory was developed for DDP with a possibly infinite set $\Pi$ of feasible solutions. It introduced conditions on an equivalence relation $\sim$ that refines $\sim_{\Pi}$ under which one can derive an additive SDP. We focus on the CO case in which $\Pi$ is finite, and show the specific properties of the SDP resulting from their theory.

A $\cC$-parametrized SDP consists of a tuple $(\cA, \cS, s_e, \allowbreak \cF, \lambda, h)$ such that $\cA$ is a finite alphabet, $\cS$ is a finite set of states, $s_e \in \cS$ is a distinguished initial state, $\cF \subseteq \cS$ is a subset of final states, and $\lambda: \cS \times \cA \rightarrow \cS$ is a transition map. A reward structure, $h$, of the $\cC$-parametrized SDP is defined through $h : \R \times \cS \times \cA \times \cC \rightarrow \R$, with $h(\xi, s, a; c)$ interpreted as the reward for transitioning to $\lambda(s,a)$, having been rewarded $\xi$ for reaching $s$, under the setting specified by $c \in \cC$. \footnote{In the definition given in \cite{Karp1967Programming} there is an additional degree of freedom $k: \cC \rightarrow \R$ that represents the reward for the null sequence $e \in \cA^*$ under any given parametric setting. We will assume $k \equiv 0$.} A $\cC$-parametrized SDP is {\it additive} when its reward structure is defined via incremental reward $\iota: \cS \times \cA \times \cC \rightarrow \R$, that is, $h(\xi,s,a;c) = \xi + \iota(s,a;c)$. We think of $\iota$ as the immediate reward for the transition $\lambda(s,a)$. Both the transition map and the reward structure can be extended recursively to $\cA^*$ in the obvious manner. 

\begin{definition}
\label{def:sim_lambda}
A $\cC$-parametrized SDP is associated with a unique equivalence relation $\sim_\lambda$ over $\cA^*$ defined via $x \sim_\lambda y \iff \lambda(s_e,x) = \lambda(s_e,y), x,y \in \cA^*$.
\end{definition}

An equivalence $x \sim_\lambda y$ is established by Definition~\ref{def:sim_lambda} iff beginning from the initial state and following the transitions prescribed by $x$ results in the same terminal state as following the transitions prescribed by $y$.

\begin{definition}
Sharing an alphabet $\cA$ and a parameter set $\cC$, an SDP $\left( \cA, \cS, s_e, \cF, \lambda, h \right)$ is said to {\it represent} a DDP $\left( \cA, \Pi, g \right)$ if the following conditions hold:
\begin{enumerate}[(\roman*)]
    \item $\Pi = \Set{x \in \cA^* \mid \lambda(s_e,x) \in \cF}$;
    \item $g(x;c) = h(0,s_e,x;c)$.
\end{enumerate}
\end{definition}

\begin{theorem}[\cite{Karp1967Programming}, Theorem 3]
\label{thm:KH}
    Let $\left( \cA, \Pi, g \right)$ be a $\cC$-parametrized DDP, and let $\sim$ be an equivalence relation over $\cA^*$. Then there exists an additive $\cC$-parametrized SDP that represents the DDP with its $\sim_\lambda$ being $\sim$ iff the following conditions hold:
    \begin{enumerate}[(i)]
    \item $\sim$ is a right congruence of finite rank (\S\ref{sec:equivalence_relations}) which refines $\sim_\Pi$;
    \item if $x \sim y$ and $xu, xv \in \Pi$ for $u,v \in \cA^*$, then for any $c \in \cC$
    \begin{equation*}
        g(xu;c)-g(yu;c) = g(xv;c)-g(yv;c).
    \end{equation*}
\end{enumerate}
\end{theorem}

The construction in Theorem~\ref{thm:KH}, which establishes its `if' statement, first determines $\left( \cA, \cS, s_e, \cF, \lambda \right)$ as follows. The set of states is given by\begin{align}
\label{eq:state_space}
   \cS \triangleq \cA^*/\sim,
\end{align}
which is finite by Theorem~\ref{thm:KH}({\romannumeral 1}). The initial state is given by $s_e \triangleq [e]$, the final states are $\cF = \Set{[x]|x\in\Pi}$, and the transition map $\lambda: \cS \times \cA \rightarrow \cS$ is determined through concatenation, $\lambda([x],a)=[xa]$, which can be extended to $\lambda: \cS \times \cA^* \rightarrow \cS$ in the obvious manner. It is well defined since $\sim$ is a right congruence by Theorem~\ref{thm:KH}({\romannumeral 1}). 

To establish additive reward structure, Karp and Held \cite{Karp1967Programming} first construct an extension $\tilde{g}: \cA^* \times \cC \rightarrow \R$. For every $x\in\Pi$, they have 
\begin{align}
\label{eq:g_tilde_equals_g}
\tilde{g}(x;c) \triangleq g(x;c), \quad \forall c \in \cC.
\end{align}
For $x \in s_\infty$, they have $\tilde{g}(x;c) \triangleq g(y;c)$, where $y$ is the longest sub-string of $x$ such that $y \notin s_\infty$. Such a $y$ always exists (it may be $e$). Here, we will make the canonical choice
\begin{align}
\label{eq:gec_zero}
    \tilde{g}(e;c) = 0, \quad \forall c \in \cC,
\end{align}
which represents zero initial reward for starting at the initial state. 
For $x \in \cA^* \setminus \left( s_e \sqcup s_\infty \sqcup \Pi \right)$, there exists $u \in \cA^*$ s.t. $xu \in \Pi$, then $yu \in \Pi$ for any $y \sim x$. By condition ({\romannumeral 2}) of Theorem~\ref{thm:KH}, the difference $g(yu;c)-g(xu;c)$
is independent of $u$. Thus, there exists $\tilde{g}$ that satisfies
\begin{align}
\label{eq:g_def}
    \tilde{g}(y;c)-\tilde{g}(x;c) = g(yu;c)-g(xu;c),
\end{align}
determined up to a choice of constant (for non-initial states). The incremental reward function $\iota: \cS \times \cA \times \cC \rightarrow \R$ is constructed via $\tilde{g}$ by
\begin{align}
\label{eq:iota_def}
    \iota(s,a;c) \triangleq \tilde{g}(xa;c)-\tilde{g}(x;c), \quad s=[x].
\end{align}
which is shown to be well defined (see discussion in \S\ref{app:proofs_sec3}). 

\begin{continuedexample}{ex:ksp}
For Theorem~\ref{thm:KH} to apply to KSP, we require an equivalence relation $\sim$ over $\cA^*$ that satisfies conditions ({\romannumeral 1}) and ({\romannumeral 2}). In particular, it should refine $\sim_\Pi$.
As noted, the equivalence relation $\sim_\Pi$ (Definition~\ref{def:simPi}) over $\cA^*$  has two distinguished classes $s_e$ \eqref{eq:state_e} and $s_\infty$ \eqref{eq:state_infinity}. Specifically, $s_\infty$ consists of all strings that are either of length greater than $d$ or correspond to (partial or full) infeasible solutions. There is an additional distinguished class consisting of $\Pi$ in its entirety, which we will denote by $s_d$. 

While there are may be many $\sim_\Pi$ refinements satisfying conditions ({\romannumeral 1}) and ({\romannumeral 2}), we will consider those $\sim$ which maintain $s_\infty$ while (possibly) refining others.\footnote{Note that there will always be at least one such refinement: the finest possible, i.e., $[x]=\{x\}$ for each $x \notin s_\infty$.} Later we will see the importance of this choice. We will define $\sim$, therefore, by providing its remaining equivalence classes. Let $x,y\in \cA^*\setminus s_e \sqcup s_d \sqcup s_\infty$ and denote $x=x_1\cdots x_l$ and $y=y_1\cdots y_{l'}$. If $x \sim_\Pi y$ then, necessarily, $l=l' < d$. For every $x$ (and similarly for $y$), we denote $W_{il}(x) \triangleq \sum_{j=1}^{l} w_{ij} x_j, i=1,\ldots,m$. For every $u = u_{l+1} \cdots u_d$ such that $xu, yu \in \Pi$, we have $W_{il}(x) + \sum_{j=l+1}^d w_{ij} u_j \leq b_i$ and $W_{il}(y) + \sum_{j=l+1}^d w_{ij} u_j \leq b_i$. 
For both inequalities to hold simultaneously, irrespective of $u$, as dictated by Definition~\ref{def:simPi}, it is sufficient to require $W_{il}(x) = W_{il}(y)$. Note that the latter is a sufficient not a necessary condition for defining an equivalence relation $\sim$ that refines $\sim_\Pi$. Apart from $s_e, s_d$ and $s_\infty$, we have its equivalence classes determined by tuples $(l,w_1,\ldots,w_m)$, so that 
\begin{align*}
    \begin{array}{l}
         x \in \cA^*\setminus s_e \sqcup s_d \sqcup s_\infty \vspace{1ex}\\
         \left[ x \right] = s_{(l,w_1,\ldots,w_m)}
    \end{array}
     \iff 
     \begin{array}{l}
          x = x_1 \cdots x_l, l<d \vspace{1ex}\\
          W_{il}(x) = w_i \leq b_i, \forall i\in\{1,\dots,m\}
     \end{array}
\end{align*}
These are the equivalent classes of partial solutions that can be extended to feasible solutions, labeled by their length $l$ and partial capacities $w_i$. Consequently, the equivalence relation $\sim$ just defined satisfies conditions ({\romannumeral 1})-({\romannumeral 2}) of Theorem~\ref{thm:KH}, the former by construction and the latter trivially. The transition map 
$\lambda: \cS \times \cA \rightarrow \cS$ is determined for $s=s_{(l,w_1,\ldots,w_m)}$, $l < d$, by
\begin{align*}
    \lambda(s_{(l,w_1,\ldots,w_m)},a) = \begin{cases}
        s_{(l+1,w_1+w_{1,l+1}a,\ldots,w_m+w_{m,l+1}a)} & l+1<d,\ w_i+w_{i,l+1}a \leq b_i,\ i=1,\ldots,m \\
        s_{d} & l+1=d,\ w_i+w_{i,l+1}a \leq b_i,\ i=1,\ldots,m \\
        s_\infty & \text{otherwise}
    \end{cases}
\end{align*}
The additive reward structure coming from Theorem~\ref{thm:KH} is defined through the auxiliary function $\tg$ 
\eqref{eq:g_tilde_equals_g}-\eqref{eq:g_def} determined here (up to a choice of scalar) by
\begin{align}
    \tg(x;c) &= \sum_{j=1}^{\max_{l'\in \cL(x)} l'} c_j x_j, \quad x = x_1 \cdots x_l \in \cA^*\nonumber\\
    \cL(x) &= \Set{\tl | 1 \leq \tl \leq \min\{l,d\}, \sum_{j=1}^{\tl} w_{ij} x_j \leq b_i, i=1,\ldots,m} \neq \emptyset, \nonumber
\end{align}
with $\tg(e;c) = 0$ and $\tg(x;c) = 0$ when $\cL(x)=\emptyset$. By \eqref{eq:iota_def} for $s=s_{(l,w_1,\ldots,w_m)}$ and $a\in \cA$ the incremental reward function $\iota$ is given by
\begin{align}
\label{eq:iota_ksp}
    \iota(s,a;c) = \tg(xa;c) - \tg(x;c) = c_{l+1}a, \quad s=[x]
\end{align}
whereas $\iota(s_{e},a;c) = c_{1}a$, $\iota(s_{d},a;c) = 0$, and $\iota(s_{\infty},a;c) = 0$.
\end{continuedexample}

%%%%%%%%%%%%%%%%%%%%%%%%%%%%%%%%%%%%
%%%%%%%%%%%%%%%%%%%%%%%%%%%%%%%%%%%%

\subsubsection{Markov Decision Processes}
\label{sec:mdp}

% textbook reference

% use the language of parametrized reward structure somewhere in the intro on MDPs

A finite MDP consists of a tuple $\left( \cS, \cA, p, r \right)$ where $\cS$ is a finite set of states, $\cA$ is a finite set of actions enacted at each state, $p(s'|s,a)$ is the conditional probability of transition from state $s$ to state $s'$ under action $a$, and $r(s,a)$ is the immediate reward for taking action $a$ at state $s$. A policy is a map $\pi: \cS \rightarrow \cA$. It is associated with a value function $V^\pi: \cS \rightarrow \R$ providing the expected return $V^\pi(s)$, starting from state $s$ and enacting the policy $\pi$ repeatedly. It is also associated with a $Q$-function $Q^\pi: \cS \times \cA \rightarrow \R$ providing the expected return $Q^\pi(s,a)$, starting from state $s$ taking the action $a$ and only then enacting the policy $\pi$ repeatedly. The goal of MDP is to find an optimal policy $\pi^*$, or, equivalently, its optimal value function $V^* \equiv V^{\pi^*}: \cS \rightarrow \R$ or $Q$-function $Q^*\equiv Q^{\pi^*}: \cS \times \cA \rightarrow \R$. 

One can characterize the optimal value function or $Q$-function neatly via fixed point theory. Consider the space
\begin{align}
\label{eq:space_of_value_fns}
    \cV_{\cS} \triangleq \Set{V | V:\cS \rightarrow \R},
\end{align}
of value functions over $\cS$.\footnote{We apply the tern ``value function" broadly, though not every $V \in \cV_{\cS}$ can be associated with a policy.} Given $\disc \in (0,1]$, define the Bellman map $B: \cV_{\cS} \rightarrow \cV_{\cS}$ to be
\begin{align}
\label{eq:B}
    BV(s) = \max_{a \in \cA} \left\{ r(s,a) + \disc \sum_{s' \in \cS} p(s'|s,a) V(s') \right\}, \quad s \in \cS.
\end{align}
The Bellman map can also be defined for $Q$-functions via
\begin{align}
\label{eq:BQ}
    BQ(s,a) = r(s,a) + \disc \sum_{s' \in \cS} p(s'|s,a) \max_{a' \in \cA} \left\{Q(s',a')\right\}, \quad s \in \cS, a \in \cA.
\end{align}

When the MDP is {\it discounted}, namely, when $\disc \in (0,1)$ in \eqref{eq:B}, $B$ is easily seen to be a contraction of modulus $\disc$ w.r.t. the $\ell_\infty$ norm \cite[\S 6]{Puterman1994MDP}, namely,
\begin{align}
\label{eq:contraction}
    \norm{BV - BV'}_\infty \leq \disc \norm{V - V'}_\infty, \quad
    \norm{BQ - BQ'}_\infty \leq \disc \norm{Q - Q'}_\infty 
\end{align}
Thus, by the Banach fixed-point theorem, $B: \cV_{\cS} \rightarrow \cV_{\cS}$ has a unique fixed point, $BV^* = V^*$, which is the optimal value function, and similarly for the optimal $Q$-function, $BQ^* = Q^*$. One can derive an optimal policy $\pi^*$ from $V^*$ by
\begin{align}
\label{eq:optimal_policy}
    \pi^*(s) \in \argmax_{a \in \cA} \left\{ r(s,a) + \disc \sum_{s' \in \cS} p(s'|s,a) V^*(s') \right\}.
\end{align}

Through \eqref{eq:contraction}, one has
\begin{align}
\label{eq:VI}
    V^* = \lim_{t \rightarrow \infty} B^tV_0,
\end{align}
converging at a linear rate for any initial $V_0 \in \cV_S$. The iterative process of improvement to optimality in \eqref{eq:VI} is known as {\it value iteration} (VI).

%%%%%%%%%%%%%%%%%%%%%%%%%%%%%%%%%%%%
%%%%%%%%%%%%%%%%%%%%%%%%%%%%%%%%%%%%

\subsubsection{Undiscounted MDPs}
\label{sec:undiscounted}

When an MDP is {\em undiscounted}, namely, when $\disc=1$ in \eqref{eq:B}, one is not guaranteed contraction for $B$ in sup-norm. For the undiscounted setting, Tseng \cite{Tseng1990Convergence} introduced the following two assumptions.

\begin{assumption}
\label{assum:absorbing}
    There is a distinguished state $s_{\infty} \in \cS$ such that $p(s_{\infty}|s_{\infty},a)=1$ and $r(s_\infty,a) = 0$ for all $a \in \cA$.
\end{assumption}

Given Assumption~\ref{assum:absorbing}, a policy $\pi: \cS \rightarrow \cA$ is said to be {\it proper} if $(p^\pi)^t(s_\infty|s) \rightarrow 1$ for $t \rightarrow \infty$, where $(p^\pi)^t(s'|s)$ is the conditional probability of getting from $s$ to $s'$ under $\pi$ in $t$ steps.     

\begin{assumption}
\label{assum:proper}
    All policies $\pi: \cS \rightarrow \cA$ of the MDP are proper.
\end{assumption}

Assumption~\ref{assum:proper} makes $s_{\infty}$ into a unique absorbing state while all other states are transient: they eventually transition to the absorbing state under any policy. Assumptions~\ref{assum:absorbing}-\ref{assum:proper} induce a state-space partition
\begin{align}
\label{eq:partition}
    \cS = \cS_0 \sqcup \cdots \sqcup \cS_{d} \sqcup \{s_\infty\}.
\end{align}
For $l=0,\ldots,d$, $s \in \cS_l$, and $a \in \cA$, we have $p(s'|s,a)>0$ iff $s' \in  \cS_{l+1} \sqcup \cdots \sqcup \cS_d \sqcup \{s_\infty\}$.

Following \cite{Tseng1990Convergence}, consider a strictly-positive, decreasing sequence
\begin{align}
\label{eq:taus}
    \tau_0 \geq \tau_1 \geq \ldots \geq  \tau_{d} \geq \tau_{\infty} > 0.
\end{align}
and denote
\begin{align}
\label{eq:delta_tau}
\modulus^\tau \triangleq \max_{l=0,\ldots,d} \frac{\tau_{l+1}}{\tau_l} \leq 1,
\end{align}
where we identify the $d+1$ index with $\infty$.
We have $\modulus^\tau < 1$ if and only if the sequence \eqref{eq:taus} is strictly decreasing.
We define $\tau: \cS \rightarrow (0,\infty)$ via
\begin{align}
\label{eq:tau_fn}
    \tau(s) & = \tau_l, \quad s \in \cS_l, \quad l=0,\ldots,d \\
    \tau(s_\infty) & = \tau_{\infty} \nonumber
\end{align}
For such $\tau$, we define the $\tau$-weighted $\ell_\infty$ norm
\begin{align}
\label{eq:tau_norm}
    \tnorm{V} \triangleq \max_{s \in \cS} \frac{\abs{V(s)}}{\tau(s)},
\end{align}
or, equivalently, $\tnorm{V \vphantom{\diag(\tau)^{-1}}} \triangleq \norm{\diag(\tau)^{-1} V}_\infty$, where $\diag(\tau)$ is the $\abs{\cS} \times \abs{\cS}$ diagonal matrix whose diagonal entries are given by $\tau(s)$.

% run $l$ up to $d-1$?

We note that $BV(s_\infty) = V(s_\infty)$ and $V^*(s_\infty)=0$ for an MDP respecting Assumptions~\ref{assum:absorbing}-\ref{assum:proper}. Therefore, let 
\begin{align}
\label{eq:VS0}
    & \cV_{\cS}^0 \triangleq \Set{V \in \cV_{\cS}| V(s_\infty)=0}.
\end{align}
We can thus restrict the undiscounted Bellman map \eqref{eq:B} to $B: \cV_{\cS}^0 \rightarrow \cV_{\cS}^0$. The following proposition appeared in \cite{Tseng1990Convergence} while we extended its proof to include a more general $\modulus^\tau$ \eqref{eq:delta_tau} and $\tau$ \eqref{eq:tau_fn} and allow  for zero transition probabilities, as would be the case with CO.

\begin{restatable}{prop}{contraction}
\label{prop:contraction}
    Let $(\cS,\cA,p,r)$ be a given undiscounted MDP with an associated Bellman map $B$. Suppose Assumptions~\ref{assum:absorbing} and \ref{assum:proper} hold, and that
    $\modulus^\tau<1$. Then $B: \cV_{\cS}^0 \rightarrow \cV_{\cS}^0$ is a contraction of modulus $\modulus^\tau$ w.r.t. the $\tau$-weighted $\ell_\infty$ norm.
\end{restatable}

By Proposition~\ref{prop:contraction} and the Banach fixed-point theorem, the undiscounted Bellman map $B: \cV_{\cS}^0 \rightarrow \cV_{\cS}^0$ has a unique fixed point, $BV^* = V^*$, which is the optimal value function in this case. 

\begin{remark}
\label{rem:sdp_mdp}
    Assuming $|\Pi|<\infty$, the $\cC$-parametrized additive SDP presented in \S\ref{sec:KHT} gives rise to a deterministic, undicounted MDP with $\cC$-parametrized reward. The state $\cS$ and action $\cA$ spaces are, respectively, the quotient set \eqref{eq:state_space} and alphabet, deterministic transitions are given by $\lambda$, namely, $p(s'|s,a) = \delta_{s'=\lambda(s,a)}$, and the reward is given by the incremental reward function $\iota$ \eqref{eq:iota_def}.
\end{remark}

%%%%%%%%%%%%%%%%%%%%%%%%%%%%%%%%%%%
%%%%%%%%%%%%%%%%%%%%%%%%%%%%%%%%%%%

\subsection{From Finite DDP to Undiscounted MDP}
\label{sec:co_repn}

In this section, we identify the properties of the CO problem, namely, its DDP representation, and of the chosen equivalence relation $\sim$ that enable the construction of a deterministic, undiscounted MDP satisfying Assumptions~\ref{assum:absorbing} and \ref{assum:proper}. This provides a unified framework for transforming CO problems into MDPs that guarantees (a) the convergence of value iteration (VI) to a unique optimal value function $V^*$ and (b) the equivalence between $V^*$ and an optimal solution of the original CO problem.

Let $\left( \cA, \Pi, g \right)$ be a finite $\cC$-parametrized DDP and let $\sim$ be an equivalence relation over $\cA^*$ that satisfies conditions ({\romannumeral 1}) and ({\romannumeral 2}) of Theorem~\ref{thm:KH}. 

% Add explanation of the assumptions.

\begin{assumption}
\label{assum:x_extension}
    For every $x \in \Pi$ there exists $u \in \cA^*$ such that $xu \in \Pi$, $\lng(xu)=d_\Pi$ (Definition~\ref{def:dpi}), and $g(xu;c) \geq g(x;c)$.
\end{assumption}

\begin{assumption}
\label{assum:s_infinity}
    Let $[x]$ denote the equivalence class of $x \in \cA^*$ w.r.t. $\sim$ over $\cA^*$. Then $[x] = s_\infty$ for any $x \in s_\infty$ \eqref{eq:state_infinity}.
\end{assumption}
Assumption~\ref{assum:x_extension} states that any string in $\Pi$ can be extended to a length $d_{\Pi}$ string in $\Pi$ without reducing the reward $g$. This assumption can easily be satisfied for various CO problems with linear objective functions given that $0\in \cA$. Assumption~\ref{assum:s_infinity} simply means that the $\sim$, which recall is a refinement of $\sim_{\Pi}$, does not refine the equivalence class $s_\infty$.

\begin{restatable}{theorem}{tsengConditions}
\label{thm:tseng_conditions}
    Let $\left( \cA, \Pi, g \right)$ be a finite $\cC$-parametrized DDP that satisfies Assumption \ref{assum:x_extension}, and let $\sim$ be an equivalence relation over $\cA^*$ that satisfies conditions ({\romannumeral 1}) and ({\romannumeral 2}) of Theorem~\ref{thm:KH} as well as Assumption~\ref{assum:s_infinity}. Then the additive $\cC$-parametrized SDP that represents it via Theorem~\ref{thm:KH} satisfies Assumptions~\ref{assum:absorbing} and \ref{assum:proper} in the sense of Remark~\ref{rem:sdp_mdp}. 
\end{restatable}

Following Theorem~\ref{thm:tseng_conditions}, we alter the reward structure of the resulting $\cC$-parametrized additive SDP, while maintaining Assumptions~\ref{assum:absorbing} and \ref{assum:proper}, to establish a connection to the optimal solution of the underlying CO problem.
Choosing a sufficiently large $M>0$, we re-define the immediate reward as follows: 
\begin{align}
\label{eq:reward_modified}
    r(s,a;c) \triangleq \left\{ 
    \begin{array}{ll}
        - M & s \notin \cF \sqcup \{s_\infty\}, \, \lambda(s,a)=s_\infty \\
        \iota(s,a;c) & \mbox{otherwise}
    \end{array}
    \right.
\end{align}
The change \eqref{eq:reward_modified} in incremental reward modifies the additive reward structure of the $\cC$-parametrized SDP, coming from Theorem~\ref{thm:tseng_conditions}, i.e., $h(\xi,s,a;c) \triangleq \xi + r(s,a;c)$. 
Given this modified reward structure, we define 
\begin{align}
\label{eq:optimal_value_function_general}
    V^*(s;c) \triangleq \max_{x \in \cA^*: \lambda(s,x)=s_\infty} h(0,s,x;c),
\end{align}
The set $\Set{x \in \cA^* 
\mid \lambda(s,x)=s_\infty}$ we are maximizing over is non-empty for every $s \in \cS$ due to Assumption~\ref{assum:proper}. Moreover, We have $r(s_\infty,a;c)=0$ 
for all $a$ (see \eqref{eq:s_inf_zero_reward} in \S\ref{app:proofs_sec3}). Adding additivity, we have $V^*(s_\infty;c) = 0$.

Note that a look-ahead $V^*(s;c)$ \eqref{eq:optimal_value_function_general} does not feature in Karp-Held theory \cite{Karp1967Programming}, 
because, unlike in CO,
the existence of a unique absorbing state is not guaranteed in their general discrete framework.

\begin{restatable}{prop}{VCOproblem}
\label{prop:V^*_co_problem}
    Under the assumptions of Theorem~\ref{thm:tseng_conditions},
    \begin{equation*}
        V^*(s_e;c) = \max_{x \in \Pi} g(x;c),
    \end{equation*}
    for every $c \in \cC$.
\end{restatable}

The functional equations of dynamic programming (DP) hold for $V^*(\cdot;c)$. Thus, the following proposition provides us with a greedy mechanism that yields an optimal solution for the CO problem for any parameter $c \in \cC$ \eqref{eq:ddp_problem}, beginning at the initial state $s_e$ and looking for the $\argmax$ at each stage.

\begin{restatable}{prop}{Vfunctional}
\label{prop:V^*_functional_equations}
    Under the assumptions of Theorem~\ref{thm:tseng_conditions}, the following functional equations hold for $V^*$ \eqref{eq:optimal_value_function_general}:
    \begin{align}
    \label{eq:dp_functional_equations}
        V^*(s;c) = \left\{ \begin{array}{ll}
            0 &  s=s_\infty \\
            \max_{a \in \cA} \{r(s,a;c) + V^*(\lambda(s,a);c)\} & \mbox{otherwise}
        \end{array} \right.
    \end{align}
\end{restatable}

The additive $\cC$-parametrized SDP resulting from Theorem~\ref{thm:tseng_conditions} together with its modified reward structure \eqref{eq:reward_modified} provides us with a $\cC$-parametrized family of MDPs $\left( \cS, \cA, p, r(\cdot,\cdot;c) \right)$ whose immediate reward is parametrized by $c \in \cC$. The state $\cS$ and action $\cA$ spaces are the quotient set \eqref{eq:state_space} and alphabet, respectively, deterministic transitions are given by $\lambda$, namely, $p(s'|s,a) = \delta_{s'=\lambda(s,a)}$, and the reward is given by \eqref{eq:reward_modified}. 

The undiscounted ($\delta=1$) Bellman map \eqref{eq:B} applied to $V \in \cV_\cS^0$ \eqref{eq:VS0} now becomes
\begin{align}
\label{eq:co_bellman}
    BV(s;c) = \max_{a \in \cA} \left\{ r(s,a;c) + V(\lambda(s,a);c) \right\}, \quad s \in \cS.
\end{align}
By Theorem~\ref{thm:tseng_conditions}, Assumptions~\ref{assum:absorbing} and \ref{assum:proper} hold for the resulting $\cC$-parametrized family of MDPs. By \cite{Tseng1990Convergence}, Assumptions~\ref{assum:absorbing} and \ref{assum:proper} guarantee the point-wise convergence of $\{B^tV_0(\cdot;c)\}_{t=0}^\infty$ to a unique fixed point for each $c \in \cC$ and  initial $V_0(\cdot;c) \in \cV_\cS^0$ (\S\ref{sec:undiscounted}). That fixed point is $V^*(\cdot;c)$ by Proposition~\ref{prop:V^*_functional_equations}.

\begin{coro}
\label{coro:co_mdp_equiv}
    Under the assumptions of Theorem~\ref{thm:tseng_conditions},
    \begin{enumerate}[label=(\roman*)]
        \item the resulting $\cC$-parametrized family of MDPs is {\em equivalent} to its underlying $\cC$-parametrized family of CO problems in the sense of Proposition~\ref{prop:V^*_co_problem}; 
        \item VI \eqref{eq:VI} converges for the resulting undiscounted $\cC$-parametrized family of MDPs to $V^*(\cdot;c)$ \eqref{eq:optimal_value_function_general} for any $c \in \cC$.
    \end{enumerate}
\end{coro}

We have thus established the informal Theorem~\ref{thm:informal}. Corollary~\ref{coro:fvi} established the MDP in question with $r(\cdot;c)$ its immediate reward structure. Proposition~\ref{prop:V^*_co_problem} and \ref{prop:V^*_functional_equations} concerning txhe optimal value function $V^*(\cdot;c)$ ensured that greedy generation results in an optimal solution for the CO problem parametrized by $c$. 

\begin{continuedexample}{ex:ksp}
Continuing with KSP \eqref{eq:KSP}, recall that we defined $\sim$ over $\cA^*$ that satisfies conditions ({\romannumeral 1}) and ({\romannumeral 2}) of Theorem~\ref{thm:tseng_conditions}. Assumption~\ref{assum:x_extension} is satisfied trivially for this $\cC$-parametrized DDP, since the only $u \in \cA^*$ that applies is $u=e$ with $d_\Pi = d$. Assumption~\ref{assum:s_infinity} is satisfied by definition: $\sim$ does not refine $s_\infty$. 

The modified reward structure $r: \cS \times \cA \times \cC \rightarrow \R$ \eqref{eq:reward_modified} maintains $\iota$ \eqref{eq:iota_ksp}, apart from  $s=s_{(l,w_1,\ldots,w_m)}$ and $a \in \cA$ such that $\lambda(s,a)=s_\infty$. In that case, $r(s_{(l,w_1,\ldots,w_m)},a;c) = - M$ if $\exists i=1,\ldots,m$ s.t. $w_i + w_{i,l+1}a > b_i$, and any $x \in s_{(l,w_1,\ldots,w_m)}$ extends to an infeasible solution by $a$. Here, we can pick, for example, $M = n \sum_{i=1}^d \abs{c_i}$. 

Figure~\ref{fig:ksp_state_space} illustrates the state space $\cS$ for the KSP
\begin{align}
    \max_{x \in \langle 5 \rangle^d} & x_1+x_2+x_3 \label{eq:ksp_example} \\
    \mbox{s.t. } & 2x_1+x_2+2x_3 \leq 4 \nonumber
\end{align}
As an example, the state $s_{(2,2)}$ consists of strings $x=x_1x_2$ of length $2$ whose partial capacity is $2$, namely, $2x_1+x_2=2$ -- there are two such strings, `$10$' and `$02$', i.e., $s_{(2,2)} = \Set{\text{`}10\text{'},\text{`}02\text{'}}$.
The actions $\cA=\langle 5 \rangle$ label the transitions we see in the transition graph (with some transitions to $s_\infty$ omitted for clarity). In particular, $\lambda(s_{(2,4)},1)=s_\infty$ since $4+2\cdot 1 > 4$, but $\lambda(s_{(2,4)},0)=s_3$, which is the state indicating the set of feasible solutions.
\end{continuedexample}

\begin{figure}
    \centering
    \begin{center}
        \tikzstyle{vertex}=[circle,draw=black!65,thick,fill=black!65,minimum size=2pt]
        \tikzstyle{block} = [fill=white, rectangle, minimum height=1.5em, minimum width=4em, rounded corners=3pt]
        \begin{tikzpicture}[arr/.style = {-Stealth, black!45}]
        \begin{scope}[scale=1.65]
        
            \foreach \pos/\name/\title in {{(0,3)/in/{$s_e$}}, {(-1,2)/obj10/$s_{(1,0)}$}, {(0,2)/obj12/$s_{(1,2)}$}, {(1,2)/obj14/$s_{(1,4)}$}, {(-2,1)/obj20/$s_{(2,0)}$}, {(-1,1)/obj21/$s_{(2,1)}$}, {(0,1)/obj22/$s_{(2,2)}$}, {(1,1)/obj23/$s_{(2,3)}$}, {(2,1)/obj24/$s_{(2,4)}$}, {(0,0)/ter/$s_3$}, {(0,-1)/inf/{$s_\infty$}}}
            \node[block] (\name) at \pos {\title};
    
            \foreach \source / \dest / \pos / \angle / \lbl in {in/obj10///$0$, in/obj12///$1$, in/obj14///$2$, obj10/obj20///$0$, obj10/obj21///$1$, obj10/obj22///$2$,  obj10/obj23/bend right/10/$3$, obj10/obj24/bend right/10/$4$, obj12/obj22///$0$, obj12/obj23///$1$, obj12/obj24///$2$, obj14/obj24///$0$, obj20/ter/bend right/10/$0$, obj20/ter/bend right/30/$1$, obj20/ter/bend right/50/$2$, obj20/inf/bend right/35/$3$, obj20/inf/bend right/50/$4$, obj21/ter/bend left/20/$1$, obj21/ter/bend right/20/$0$, obj22/ter/bend right/20/$0$, obj22/ter/bend left/20/$1$, obj23/ter///$0$, obj24/ter///$0$, obj24/inf///$1$
                                        }
                \draw[arr] (\source) to [\pos=\angle] node[sloped, black] {\lbl} (\dest);

            \draw[arr] (ter) to[bend right=30] coordinate[pos=.5] (sd0) node[sloped, black] {$0$} (inf);
            \draw[arr] (ter) to[bend left=30] coordinate[pos=.5] (sd4) node[sloped, black] {$4$} (inf);
            \draw[dotted] (sd0)--(sd4);
            
            \draw[arr] (obj24) to[bend left=20] coordinate[pos=.5] (s242) node[sloped, black] {$2$} (inf); %, /inf/bend left/20/$2$
            \draw[arr] (obj24) to[bend left=50] coordinate[pos=.5] (s244) node[sloped, black] {$4$} (inf); %, /inf/bend left/50/$4$
            \draw[dotted] (s242)--(s244);
            
        \end{scope}
        \end{tikzpicture}    
    \end{center}
    \caption{States and action-labeled transitions for the KSP \eqref{eq:ksp_example}. Some actions in $\cA=\langle 5 \rangle$ leading to $s_\infty$ were omitted for clarity.}
    \label{fig:ksp_state_space}
\end{figure}
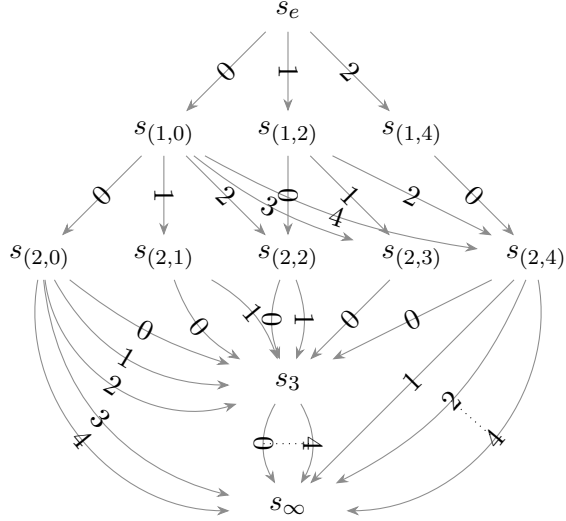

%%%%%%%%%%%%%%%%%%%%%%%%%%%%%%%%%%%
%%%%%%%%%%%%%%%%%%%%%%%%%%%%%%%%%%%

\subsection{\texorpdfstring{$\epsilon$}{epsilon}-Approximation of \texorpdfstring{$V^*$}{V*}}
\label{sec:epsilon_approx}

Following Corollary~\ref{coro:co_mdp_equiv}, the optimal value function $V^*$ of the derived $\cC$-parametrized MDP allows us to construct an optimal solution for the underlying $\cC$-parametrized CO problem. 

\begin{definition}
\label{def:epsilon_approx}
Let $V \in \cV_{\cS}^0$ and $\epsilon > 0$. We say that $V$ is an {\it $\epsilon$-approximation} of $V^*$ w.r.t. a $\tau$-weighted $\ell_\infty$ norm if $\tnorm{V-V^*} < \epsilon$.     
\end{definition}

Under the following additional Assumption~\ref{assum:partial_to_feasible}, any $\epsilon$-approximation of $V^*$ can be translated into an $O(d_{\pi}\epsilon)$-optimal solution. Assumption~\ref{assum:partial_to_feasible} guarantees that every partially feasible solution can be extended to a fully feasible solution; in particular, it implies that $V^*(s;c) > -M$ for any $s \in \cS$. It relies on our choice of equivalence relation $\sim$ over $\cA^*$ that satisfies conditions ({\romannumeral 1}) and ({\romannumeral 2}) of Theorem~\ref{thm:KH}.

\begin{assumption}
\label{assum:partial_to_feasible}
    For any $x \in \cA^* \setminus s_\infty$ there exists $u \in \cA^*$ such that $[xu] \in  \Set{[y]|y\in\Pi}$. 
\end{assumption}

Given any $V \in \cV_{\cS}^0$, define the following greedy generation sequence $\{(s_i,a_{i+1})\}_{i=0}^{k}$ recursively, beginning with $s_0=s_e$,
\begin{align}
    & a_{i+1} \in \argmax_{a \in \cA} \left\{ r(s_i,a;c) + V(\lambda(s_i,a)) \right\}; \label{eq:approx_solution} \\
    & s_{i+1} = \lambda(s_{i},a_{i+1}). \nonumber
\end{align}
The sequence terminates at the first $k$ such that $\lambda(s_{k},a_{k+1}) = s_\infty$. 

\begin{restatable}{theorem}{Approximation}
\label{thm:approximation}
    Let ${V}\in \cV_{\cS}^0$ be an $\epsilon$-approximation of $V^*$ w.r.t. a $\tau$-weighted $\ell_\infty$ norm \eqref{eq:tau_norm}. Let $\{(s_i,a_{i+1})\}_{i=0}^{k}$ be as defined \eqref{eq:approx_solution}. Under the assumptions of Theorem~\ref{thm:tseng_conditions} and Assumption~\ref{assum:partial_to_feasible}, for a sufficiently small $\epsilon > 0$, the following holds:
    \begin{enumerate}[label=(\roman*)]
        \item $y=a_1 \cdots a_{k} \in \cA^*$ is a feasible solution, namely, $y \in \Pi$;
        \item $\max_{x \in \Pi} g(x;c) \leq 2 \epsilon \tau_0 (d_\Pi + 1) + g(y;c)$.
    \end{enumerate}
\end{restatable}

%%%%%%%%%%%%%%%%%%%%%%%%%%%%%%%%%%%%
%%%%%%%%%%%%%%%%%%%%%%%%%%%%%%%%%%%%

\subsection{Additional Examples}

To show the generality of our approach, we provide additional well-known CO problems with their derived MDP formulations.

%%%%%%%%%%%%%%%%%%%%%%%%%%%%%%%%%%%%
%%%%%%%%%%%%%%%%%%%%%%%%%%%%%%%%%%%%

\subsubsection{Traveling Salesman Problem}
\label{sec:TSP}

Let $\cC \subseteq \R_{\geq 0}^{d \times d}$ be a parameter set. The TSP has a $\cC$-parametrized DDP formulation $\left( \cA, \Pi, g \right)$ with $\cA = \langle d \rangle$ the set of indexed cities ($0$ denoting the home base), $\Pi$ the set of valid routes of length $d+1$, i.e., $x = x_0 \cdots x_d \in \cA^*$, $x_j \in \cA$, where $x_0 = x_d =0$ and $\Set{x_0,\ldots,x_d}=\cA$, and a $\cC$-parametrized objective $g(x;c) = - \sum_{j=0}^{d-1} c[{x_j, x_{j+1}}] = - \left(c[{0,x_1}] + \ldots + c[{x_{d-1},0}] \vphantom{\sum}\right)$.
Each $c \in \cC$ is a matrix such that $c[{x_i,x_j}] \geq 0$ is the weight associated with the transition from $x_i$ to $x_j$. Here, $d_\Pi=d+1$ (Definition~\ref{def:dpi}).

For Theorem~\ref{thm:KH} to apply, we require an equivalence relation $\sim$ over $\cA^*$ that satisfies conditions ({\romannumeral 1}) and ({\romannumeral 2}). First, consider $\sim_\Pi$ (Definition~\ref{def:simPi}). As noted, it has two distinguished classes $s_e$ \eqref{eq:state_e} and $s_\infty$ \eqref{eq:state_infinity}, where $s_\infty$ consists of all routes that are either of length $> d+1$ or correspond to illegal (partial or full) routes, for example, routes that do not start at the home base. It has an additional distinguished class consisting of $\Pi$ in its entirety, which we will denote by $s_{d+1}$. All other equivalence classes of $\sim_\Pi$ that partition $\cA^* \setminus s_e \sqcup s_{d+1} \sqcup s_\infty$ are characterized by subsets $\emptyset \subseteq \cB \subseteq \cA \setminus \{0\}$ (for details, here and below, see \S\ref{app:TSP}). 

Now, to establish an appropriate $\sim$ it is sufficient to finitely refine $\sim_\Pi$ (apart from $s_e, s_{d+1}, s_\infty$) by pointed sets, i.e., pairs $(\cB,b)$, where $\emptyset \subsetneq \cB \subseteq \cA \setminus \{0\}$ and $b \in \cB$. The set $\cB=\emptyset$ characterizes the $\sim_\Pi$ equivalence class $[0]=\{0\}$ -- it cannot and does not need to be refined as ({\romannumeral 2}) holds trivially. To conclude,
\begin{align*}
    \begin{array}{l}
         x \in \cA^*\setminus s_e \sqcup s_{d+1} \sqcup s_\infty \vspace{1ex}\\
         \left[ x \right] = s_{(\cB,b)} \text{ or } s_\emptyset \vspace{1ex}\\
        \emptyset \subsetneq \cB \subseteq \cA \setminus \{0\}, b \in \cB    
    \end{array}
     \iff 
     \begin{array}{l}
          x = x_0 \cdots x_l, l = \abs{\cB} \text{ or } l=0 \vspace{1ex}\\
          x_0 = 0  \vspace{1ex}\\
          \{x_1,\ldots,x_l\} = \cB, x_l = b, l>0
     \end{array}
\end{align*}

Following the construction of Theorem~\ref{thm:KH}, $\cF = \{s_{d+1}\}$ and the transition map 
$\lambda: \cS \times \cA \rightarrow \cS$ is determined by
\begin{align*}
    \lambda(s,a) = \begin{cases}
        s_\emptyset & s=s_e, a=0 \\
        s_{(\{a\},a)} & s=s_\emptyset, a\neq 0 \\
        s_{(\cB \cup \{a\},a)} & s = s_{(\cB,b)}, \cB \subsetneq \cB \cup \{a\}, a\neq 0 \\
        s_{d+1} & s = s_{(\cB,b)},a=0, 0\notin \cB,|\cB|=d-1\\
        s_\infty & \text{otherwise}
    \end{cases}
\end{align*}
Figure~\ref{fig:tsp_state_space} illustrates $\cS$ for the TSP with $d=3$ including action-labeled transitions. As an example, $s_{(\{1\},1)}=\{\text{`}01\text{'}\}$ and $s_{(\{1,2\},1)}=\{\text{`}021\text{'}\}$, while $s_{4} = \Set{\text{`}0120\text{'},\text{`}0210\text{'}}$.

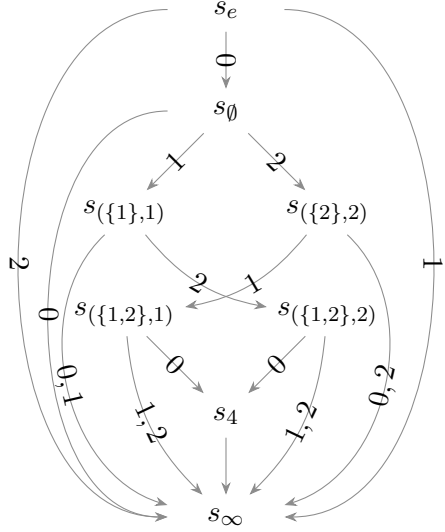
\begin{figure}
    \centering
    \begin{center}
        \tikzstyle{vertex}=[circle,draw=black!65,thick,fill=black!65,minimum size=2pt]
        \tikzstyle{block} = [fill=white, rectangle, minimum height=1.5em, minimum width=4em, rounded corners=3pt]
        \begin{tikzpicture}[arr/.style = {-Stealth, black!45}]
        \begin{scope}[scale=1.35]
        
            \foreach \pos/\name/\title in {{(0,3)/in/{$s_e$}}, {(0,2)/obj0/$s_{\emptyset}$}, {(-1,1)/obj11/$s_{(\{1\},1)}$}, {(1,1)/obj22/$s_{(\{2\},2)}$}, {(-1,0)/obj121/$s_{(\{1,2\},1)}$}, {(1,0)/obj122/$s_{(\{1,2\},2)}$}, {(0,-1)/ter/$s_4$}, {(0,-2)/inf/$s_\infty$}}
            \node[block] (\name) at \pos {\title};
    
            \foreach \source / \dest / \pos / \angle / \lbl in {{in/obj0///$0$}, {in/inf/bend left/90/$1$}, {in/inf/bend right/90/$2$}, {obj0/obj11///$1$}, {obj0/obj22///$2$}, {obj0/inf/bend right/90/$0$}, {obj11/obj122/bend right/20/$2$}, {obj11/inf/bend right/60/$0,1$}, {obj22/obj121/bend left/20/$1$}, {obj22/inf/bend left/60/$0,2$}, {obj121/ter///$0$}, {obj121/inf/bend right/20/$1,2$}, {obj122/ter///$0$}, {obj122/inf/bend left/20/$1,2$}, {ter/inf///}} 
            \draw[arr] (\source) to [\pos=\angle] node[sloped, black] {\lbl} (\dest);

        \end{scope}
        \end{tikzpicture}    
    \end{center}
    \caption{States and action-labeled transitions for the TSP with $d=3$.}
    \label{fig:tsp_state_space}
\end{figure}

The reward is derived from the incremental reward function $\iota: \cS \times \cA \times \cC \rightarrow \R$ given for $s=s_{(\cB,b)}$ or $s_\emptyset$ by $\iota(s,a;c) = - c[x_l,a]$ where $s=[x]$,
whereas $\iota(s_{e},a;c) = 0$, $\iota(s_{d+1},a;c) = 0$, and $\iota(s_{\infty},a;c) = 0$.
Assumptions~\ref{assum:x_extension}--\ref{assum:partial_to_feasible} hold trivially so that  Theorems~\ref{thm:tseng_conditions} and \ref{thm:approximation} apply.
The modified reward structure $r: \cS \times \cA \times \cC \rightarrow \R$ \eqref{eq:reward_modified} maintains $\iota$, apart from  $s \in \cS$ and $a \in \cA$ such that $\lambda(s,a)=s_\infty$. Thus, $r(s,a;c) = - M$ iff $s=s_e, a\neq 0$ or $s=s_\emptyset, a=0$ or $s=s_{(\cB,b)}, \cB \cup \{a\} = \cB \text{ or } |\cB|<d-1, a=0$. Here, we can pick, for example, $M = \sum_{i,j=0}^d c[i,j]>0$. 

%%%%%%%%%%%%%%%%%%%%%%%%%%%%%%%%%%%%
%%%%%%%%%%%%%%%%%%%%%%%%%%%%%%%%%%%%

\subsubsection{Shortest Path Problem}
\label{sec:SPP}

The single-source, single-target {\it shortest path problems} (SPPs) we consider are characterized, each, by a weighted finite directed graph comprising a finite set of vertices $\cA$, a finite set of edges $\cE \subseteq \cA \times \cA$, a cost function $c: \cE \rightarrow \R_{\geq 0}$, and a choice of distinct source $\vin \in \cA$ and target $\vter \in \cA$ vertices. A path in the graph from the source to the target vertex is a sequence $x=x_0 \cdots x_l \in \cA^*$ such that $x_0=\vin$, $x_l=\vter$, and $(x_i,x_{i+1}) \in \cE$ for $i=0,\ldots,l-1$. The CO problem minimizes the total path cost $\sum_{i=0}^{l-1} c((x_i,x_{i+1}))$ over all such $x$, which can be assumed to satisfy $\lng(x)\leq\abs{\cA}$ due to the non-negativity of the edge weights. Moreover, we may as well assume $\cE = \cA \times \cA$, since we can always introduce edges with sufficiently large weights.

Let $\cC \subseteq \R_{\geq 0}^{\abs{\cA} \times \abs{\cA}}$ be a parameter set and denote $d=\abs{\cA}-1$. The SPP, as described above, has a $\cC$-parametrized DDP formulation $\left( \cA, \Pi, g \right)$ with $\cA$ the alphabet, $\Pi$ the set of strings $x = x_1 \cdots x_l \in \cA^*$,$l \leq d$, s.t. $x_l = \vter$, and the $\cC$-parametrized objective
\begin{align}
\label{eq:g_spp}
    g(x;c) &= - \left( c[{\vin,x_1}] + \sum_{i=1}^{l-1} c[x_i,x_{i+1}]\right).
\end{align}
Each $c \in \cC$ is a matrix such that $c[{x_i,x_j}] \geq 0$ is the weight associated with the edge $(x_i,x_j) \in \cE$. Here, $d_\Pi=d$ (Definition~\ref{def:dpi}). One may easily conceive of an alternative, but equivalent, DDP formulation in which $\Pi$ is the set of all strings $x = x_0 \cdots x_l$ s.t. $x_0=\vin$ and $x_l=\vter$ and $g(x;c) = - \sum_{i=0}^{l-1}c[x_i,x_{i+1}]$. In the latter DDP formulation $d_\Pi=\abs{\cA}$.

Our choice of $\Pi$ results in the equivalence relation $\sim_\Pi$ over $\cA^*$ with $s_e$ and $s_\infty$, the latter consisting of all $x\in\cA^*$ such that either $\lng(x) \geq d+1$ or $\lng(x)=d$ while $x_d \neq \vter$, but, also, $d$ additional equivalence classes that partition $\cA^* \setminus s_e \sqcup s_\infty$ according to $\lng(x)$. 

The coarsest finite rank refinement to guarantee conditions ({\romannumeral 1}) and ({\romannumeral 2}) of Theorem~\ref{thm:KH} has the following equivalence classes, on top of $s_e$ and $s_\infty$,
\begin{align*}
    \begin{array}{l}
         x \in \cA^*\setminus s_e \sqcup s_\infty \vspace{1ex}\\
         \left[ x \right] = s_{(l,a)} \text{ or } s_d \vspace{1ex}\\
        0 < l < d, a \in \cA
    \end{array}
     \iff 
     \begin{array}{l}
          x = x_1 \cdots x_l \vspace{1ex}\\
          0< l < d, x_l=a \text{ or } l=d, x_l=\vter \vspace{1ex}\\
          \\
     \end{array}
\end{align*}
and it is clearly a right congruence (for details, see \S\ref{app:SPP}). Unlike in the case of the KSP of Example~\ref{ex:ksp} or the TSP \S\ref{sec:TSP}, the set $\Pi$ of full and feasible solutions is partitioned in the SPP state space, namely, $\Pi = s_d \sqcup \bigsqcup_{l=1}^{d-1} s_{(l,\vter)}$.

The resulting additive SDP of Theorem~\ref{thm:KH} involves the tuple $\left( \cA, \cS, s_e, \cF, \lambda \right)$ with $\cS = A^*/\sim = \Set{s_e,s_d,s_\infty} \sqcup \Set{s_{(l,a)} \mid 0<l<d, a \in \cA}$ and $\cF = \Set{s_d} \sqcup \Set{s_{(l,\vter)} \mid 0<l<d}$.
Thus, the state space is $O(\abs{\cE})$. The state transitions are given by
\begin{align*}
    \lambda(s,a) = \left\{ 
    \begin{array}{ll}
        s_{(1,a)} & s=s_e \\
        s_{(l,a)} & s=s_{(l-1,a')}, l<d-1 \\ 
        s_d & s=s_{(d-1,a')}, a=\vter \\ 
        s_\infty & \text{otherwise}
    \end{array}
    \right.
\end{align*}
Figure~\ref{fig:spp_state_space} illustrates the state space $\cS$ for the SPP with $d=3$, including action-labeled transitions. As an example, $s_{(1,1)}$ corresponds to the paths $\{\text{`}01\text{'}\}$ while $s_{(2,1)}$ corresponds to the paths $\Set{\text{`}001\text{'},\text{`}011\text{'},\text{`}021\text{'},\text{`}031\text{'}}$. Transitions, such as $\lambda(s_{(1,2)},2)=s_{(1,2)}$ or $\lambda(s_{(2,3)},3)=s_{3}$, that correspond to self loops along the path are allowed. Self loops do not increase the size of the state space.\footnote{Limiting the MDP to state-specific actions is challenging from a transformer-type sequence generation perspective: the set of tokens should remain fixed, while conditional probability determine their relevance.}

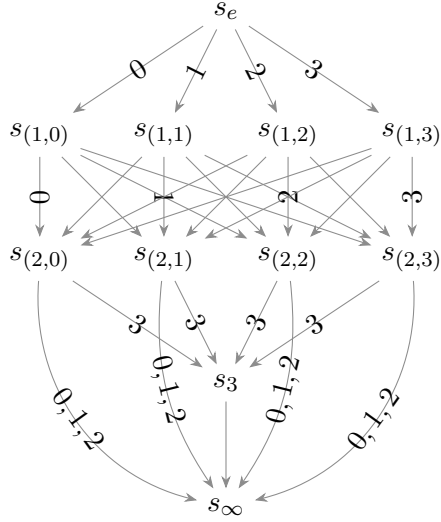
\begin{figure}
    \centering
    \begin{center}
        \tikzstyle{vertex}=[circle,draw=black!65,thick,fill=black!65,minimum size=2pt]
        \tikzstyle{block} = [fill=white, rectangle, minimum height=1em, minimum width=1em, rounded corners=0pt]
        \begin{tikzpicture}[arr/.style = {-Stealth, black!45}]
        \begin{scope}[scale=1.65]

            \foreach \pos/\name/\title in {{(0,4)/init/$s_e$}, 
                                    {(-1.5,3)/obj10/$s_{(1,0)}$}, {(-0.5,3)/obj11/$s_{(1,1)}$}, {(0.5,3)/obj12/$s_{(1,2)}$}, {(1.5,3)/obj13/$s_{(1,3)}$},
                                    {(-1.5,2)/obj20/$s_{(2,0)}$}, 
                                    {(-0.5,2)/obj21/$s_{(2,1)}$}, 
                                    {(0.5,2)/obj22/$s_{(2,2)}$}, 
                                    {(1.5,2)/obj23/$s_{(2,3)}$}, 
                                    {(0,1)/ter/$s_3$},
                                    {(0,0)/infis/$s_\infty$}}
            \node[block] (\name) at \pos {\title};
    
            \foreach \source / \dest / \pos / \angle / \lbl in {
            init/obj10//$0$, init/obj11//$1$, init/obj12//$2$, init/obj13//$3$, 
            obj10/obj20///$0$, obj10/obj21///, obj10/obj22///, obj10/obj23///,  
            obj11/obj20///, obj11/obj21///$1$, obj11/obj22///, obj11/obj23///, 
            obj12/obj20///, obj12/obj21///, obj12/obj22///$2$, obj12/obj23///, 
            obj13/obj20///, obj13/obj21///, obj13/obj22///, obj13/obj23///$3$, 
            obj20/ter///$3$, {obj20/infis/bend right/40/$0,1,2$}, 
            obj21/ter///$3$, {obj21/infis/bend right/20/$0,1,2$},
            obj22/ter///$3$, {obj22/infis/bend left/20/$0,1,2$},
            obj23/ter///$3$, {obj23/infis/bend left/40/$0,1,2$}, 
            ter/infis///
            }
            \draw[arr] (\source) to [\pos=\angle] node[sloped, black] {\lbl} (\dest);
            
        \end{scope}
        \end{tikzpicture}    
    \end{center}
    \caption{states and action-labeled transitions for the SPP with $d=3$ ($\cA=\Set{0,1,2,3}$, $\vin=0$, $\vter=3$). Some action labels were omitted for clarity.}
    \label{fig:spp_state_space}
\end{figure}

The additive reward structure coming from Theorem~\ref{thm:KH} is defined through the help function $\tg:\cA^* \times \cC \rightarrow \R$ \eqref{eq:g_tilde_equals_g}-\eqref{eq:g_def} determined here (up to a choice of scalar) by
\begin{align*}
    \tg(x;c) &= - \left(c[{\vin, x_{1}}] + \sum_{j=1}^{\min\{d,l\}-1} c[{x_j, x_{j+1}}]\right), \quad x=x_1 \cdots x_l \in \cA^*,
\end{align*}
with $\tg(e;c)=0$. Similarly to TSP, the incremental reward function $\iota: \cS \times \cA \times \cC \rightarrow \R$ for $s=s_{(l,a')}$ is given by
\begin{align}
\label{eq:iota_spp}
    \iota(s,a;c) = \tg(xa;c) - \tg(x;c) = -c[x_l,a], \quad s=[x]
\end{align}
whereas $\iota(s_{e},a;c) = -c[\vin,a]$, $\iota(s_{d},a;c) = 0$, and $\iota(s_{\infty},a;c) = 0$.

For Theorem~\ref{thm:tseng_conditions} to apply, we require the equivalence relation $\sim$ over $\cA^*$ to satisfy Assumptions~\ref{assum:x_extension} and \ref{assum:s_infinity}. For Assumption~\ref{assum:x_extension} to hold, it is sufficient to require $c[\vter,\vter]=0$, which is a standard requirement in DP formulations of SPP \cite{Pollack1960ShortestRoute}. However, we do not need to require other self loops to carry zero cost, i.e., $c[a,a] \geq 0$ for $a \neq \vter$. Assumption~\ref{assum:s_infinity} holds by construction. 

The modified reward structure $r: \cS \times \cA \times \cC \rightarrow \R$ \eqref{eq:reward_modified} maintains $\iota$ \eqref{eq:iota_spp}, apart from  $s$ and $a$ such that $\lambda(s,a)=s_\infty$. Thus, $r(s,a;c) = - M$ iff $s=s_{(d-1,a')}, a\neq\vter$. Here, we can pick, for example, $M = \sum_{i,j=1}^{d+1} c[i,j] > 0$. 
For Theorem~\ref{thm:approximation} to apply, we require Assumption~\ref{assum:partial_to_feasible}, which holds since, for any $x \in \cA^* \setminus s_e \sqcup s_{d} \sqcup s_\infty$, we can choose $u=\vter$ to get $[xu] \in \cF$.

A few notes are in order before we finish the discussion on the SPP. Karp and Held \cite{Karp1967Programming} discuss an alternative formulation of the SPP as a DDP. In their formulation $\Pi = \Set{x = x_1 \cdots x_l \in \cA^* \mid x_l = \vter}$, which, strictly speaking, considers the SPP as a discrete optimization problem. The equivalence classes of $\sim_\Pi$, in this case, consist of $\Set{s_e} \sqcup \Set{s_a \mid a \in \cA}$, in particular, there is no $s_\infty$ -- all strings are either in $\Pi$ or can be extended to strings in $\Pi$ -- and there is no absorbing state, thus, Assumption~\ref{assum:s_infinity} is violated. 

Lastly, changing the DDP formulation above to one in which $\Pi$ is the set of strings $x = x_0 \cdots x_l \in \cA^*$ such that $x_l = \vter$ and $1 \leq l \leq d+1$ and $g(x;c) = - \sum_{i=0}^{l-1} c[x_i,x_{i+1}]$ would have resulted in an MDP formulation that captures the single-target SPP. Proposition~\ref{prop:V^*_co_problem} would have made little sense, choosing the minimal cost among all paths that terminate in $\vter$. On the other hand, $V^*(s_{(1,a)};c)$ would then have captured the minimal cost path from any $a \in \cA$ to $\vter$.

%%%%%%%%%%%%%%%%%%%%%%%%%%%%%%%%%%%%
%%%%%%%%%%%%%%%%%%%%%%%%%%%%%%%%%%%%

\section{RL Training}
\label{sec:rl}

We now turn to discuss the RL framework aimed at producing $\epsilon$-approximations of $V^*$ (Definition~\ref{def:epsilon_approx}). For the sake of simplicity, in this part, we fix a problem instance $c \in \cC$ and suppress it, henceforth, from our notation. In particular, we replace $V(s;c)$ with $V(s)$ for any value function $V$ in the space of value functions
$\cV_{\cS}$ \eqref{eq:space_of_value_fns}.

%%%%%%%%%%%%%%%%%%%%%%%%%%%%%%%%%%%%
%%%%%%%%%%%%%%%%%%%%%%%%%%%%%%%%%%%%

\subsection{Preliminaries}

By Theorem~\ref{thm:tseng_conditions}, representable $\cC$-parametrized families of CO problems, constituting a special case of undiscounted MDPs that respect Assumptions~\ref{assum:absorbing}-\ref{assum:proper}, admit a state space partition 
\begin{align}
    \cS = \cS_0 \sqcup \cdots \sqcup \cS_{d_\Pi} \sqcup \{s_\infty\},
\end{align}
such that, for $l=0,\ldots,d_\Pi$, $s \in \cS_l$, and $a \in \cA$, one has $p(s'|s,a)>0$ iff $s' \in  \cS_{l+1} \sqcup \{s_\infty\}$.
Denote the maximal reward and maximal girth of the MDP by
\begin{align}
    R \triangleq \max_{s,a} r(s,a), \quad
    W \triangleq \max_{l} \abs{\cS_l}. \label{eq:RW}
\end{align}
Let $B: \cV_{\cS}^0 \rightarrow \cV_{\cS}^0$ be the undiscounted Bellman map \eqref{eq:co_bellman} with its unique optimal value function $V^*$ \S\ref{sec:undiscounted}. Using these definitions and $\modulus^{\tau}$ \eqref{eq:delta_tau}, we can characterize a compact set in which $V^*$ lies. 

\begin{restatable}{prop}{Vbound}
\label{prop:V^*_bound}
Let $\rho=\frac{R \sqrt{W} \modulus^\tau}{1-\modulus^\tau} \left( \frac{1}{\tau_\infty} - \frac{1}{\tau_0}\right)$. Then, $V^*\in \mathcal{B}_\rho^\tau(0)$ where the metric in $\cV_{\cS}^0$ is derived from $\tnorm{\cdot}$.
\end{restatable}

We will later use this characterization to define a parametrized set in which to search for $V^*$ approximates.

\subsection{Approximation Schemes}
\label{sec:approx_scheme}

Consider a finite set of $K$ state features $\phi_i: \cS \setminus \{s_\infty\} \rightarrow \R, i=1,\ldots,K$
assembled into a state feature vector $\phi(s)\triangleq(\phi_1(s),\ldots,\phi_K(s)), s \in \cS \setminus \{s_\infty\}$
that constitutes the embedding of the state $s$ into $\R^K$.  
An embedding is considered {\it efficient} (also known in the literature as ``compact'' \cite{Tsitsiklis1996Feature}) when $K\ll|\cS|$.

Let $\Theta$ be a compact and convex parameter set and let 
\begin{align}
    v: \R^K \times \Theta \rightarrow \R.
\end{align}
be continuous. This results in a $\Theta$-parametrized family of value functions $\cV_\Theta \triangleq \Set{V_\theta \mid \theta \in \Theta} \subset \cV_{\cS}^0$
with $V_\theta: \cS \rightarrow \R$ defined for every $\theta \in \Theta$ via $V_{{\theta}}(s) = v(\phi(s),{\theta})$ when $s \in \cS \setminus \{s_\infty\}$ while $V_{{\theta}}(s) = 0$ when $s = s_\infty$.

\begin{definition}
The compact subset $\cV_\Theta \subset \cV_{\cS}^0$ is an {\it approximation scheme} for value functions.\footnote{Though $\cV_\Theta$ depends on $v$, we suppress this dependence in our notation.}
\end{definition}

An approximation scheme is considered {\it efficient} when the states embedding $\phi$ is.

\begin{definition}
\label{def:sufficiently_expressive}
Following Proposition~\ref{prop:V^*_bound}, an approximation scheme is said to be {\it sufficiently expressive} w.r.t. $\tnorm{\cdot}$ if $\mathcal{B}_\rho^\tau(0) \cap \cV_\Theta \neq \emptyset$ for $\rho = \frac{R \sqrt{W} \modulus^\tau}{1-\modulus^\tau} \left( \frac{1}{\tau_\infty} - \frac{1}{\tau_0}\right)$.
\end{definition}

When an approximation scheme is sufficiently expressive, it can approximate $V^* \in \mathcal{B}_\rho^\tau(0)$ within $2\rho$. On the other extreme, $\cV_\Theta$ will be {\it fully expressive} when $V^* \in \cV_\Theta$.

%%%%%%%%%%%%%%%%%%%%%%%%%%%%%%%%%%%%
%%%%%%%%%%%%%%%%%%%%%%%%%%%%%%%%%%%%

\subsection{Projected Value Iteration}
\label{sec:pvi}

We now introduce projected value iteration: an iterative procedure that aims to approximate VI, where at each iteration we compute the weighted projection of the Bellman map applied to the current iterate onto the approximation scheme $\cV_{\Theta}$. As noted, the weights of the projection may be obtained by some learning mechanism employing exploration and exploitation. Hereafter, we will assume those to be given. We will provide assumptions under which projected value iteration converges and bound its approximation error with respect to $V^*$. While projected value iteration is not implementable in practice, it plays a key role in isolating out the  approximation error stemming from the approximation scheme.

For $\sigma \in \Delta_{S}$, define the $\sigma$-weighted $\ell_2$-norm on $\cV_S$ to be
\begin{align}
\label{eq:sigma_norm}
    \snorm{V} \triangleq \sqrt{\E_{s\sim\sigma}\left[ V(s)^2 \right]} = \norm{\diag(\sigma)^{1/2} V}_2.
\end{align}
When $\cV_\Theta$ is compact, for every $V \in \cV^0_S$ there exists $\theta_V^* \in \argmin_{\theta\in\Theta} \snorm{V_{\theta}-V}$ such that
\begin{align}
\label{eq:projection_min}
    \snorm{V_{\theta_V^*} - V} = \min_{\theta\in\Theta} \snorm{V_{\theta}-V}.
\end{align}
By the Axiom of Choice, i.e., an arbitrary choice, we get a well-defined notion of a projection, $\Pi^\sigma_\Theta: \cV_S^0 \rightarrow \cV_\Theta$,
w.r.t. $\snorm{\cdot}$ defined via
\begin{align}
\label{eq:proj}
    \Pi^\sigma_\Theta V \triangleq V_{\theta_V^*}.
\end{align}
When $\cV_\Theta$ is closed and convex, we have a uniquely determined continuous $\Pi^\sigma_\Theta$ \cite[Theorem 2.5]{conway1985functional}.

\begin{definition}
    Consider the undiscounted Bellman map $B: \cV_{\cS}^0 \rightarrow \cV_{\cS}^0$ \eqref{eq:co_bellman}. 
    \begin{enumerate}[label=(\roman*)]
        \item The {\it $\sigma$-projected Bellman map} in a given approximation scheme is defined to be
        \begin{align}
            \Pi^\sigma_\Theta B : \cV_S^0 \rightarrow \cV_\Theta.
        \end{align}
        \item The {\it $\sigma$-projected VI} (PVI)  initialized with $V_{\theta_0} \in \cV_\Theta$ is defined by the iterative update process
        \begin{align}
        \label{eq:pvi}
            V_{{\theta}_{t+1}} \triangleq \Pi^\sigma_\Theta BV_{{\theta}_{t}}.
        \end{align} 
    \end{enumerate}
\end{definition}

When the limit of PVI \eqref{eq:pvi} exists, we denote it by
\begin{align}
\label{eq:pvi_limit}
    \tilde{V}^* \triangleq \lim_{t \rightarrow \infty} V_{\theta_t} \equiv \lim_{t \rightarrow \infty} \left( \Pi^\sigma_\Theta B \right)^t V_{\theta_0} \in \cV_\Theta \subset \cV_S^0
\end{align}
For now, we do not claim it to be independent of $V_{\theta_0}$.

\begin{assumption}
\label{assum:contraction}
    There exists $\tau: \cS \rightarrow (0,\infty)$ \eqref{eq:tau_fn} such that $\Pi^\sigma_\Theta B$ is a contraction of modulus $\modulus <1$ w.r.t. $\tnorm{\cdot}$.
\end{assumption}

The following proposition provides a bound on the limit of $\sigma$-PVI under Assumption~\ref{assum:contraction}.

\begin{restatable}{prop}{PVI}
\label{prop:pvi}
Under Assumption~\ref{assum:contraction}, $\sigma$-PVI \eqref{eq:pvi} converges irrespective of initialization and its limit \eqref{eq:pvi_limit} satisfies
\begin{align}
\label{eq:proj_vi_rate}
    \tnorm{\tilde{V}^* - \Pi^\sigma_\Theta V^*} \leq \frac{\modulus}{1 - \modulus} \cdot \tnorm{V^* - \Pi^\sigma_\Theta V^* \vphantom{\lim_{t \rightarrow \infty} V_{{\theta}_t}}}.
\end{align}
In particular, the approximation scheme is fully expressive iff $\tilde{V}^* = V^*$.
\end{restatable}

The difference $\tnorm{V^* - \Pi^\sigma_\Theta V^*}$ on the RHS of \eqref{eq:proj_vi_rate} is the {\it approximation error}. The approximation error is a fundamental aspect of the given approximation scheme that we cannot eliminate.  Proposition~\ref{prop:pvi} allows us to separate out the approximation error while focusing, instead, on the $\sigma$-PVI limit. Given the conditions of Proposition~\ref{prop:pvi}, its limit $\tilde{V}^*$ \eqref{eq:pvi_limit} exists and is identified with 
the unique fixed point of the $\sigma$-projected Bellman map
\begin{align}
\label{eq:pvi_fixed_pt}
    \Pi^\sigma_\Theta B \tilde{V}^* = \tilde{V}^*.
\end{align}    

\begin{remark}
\label{rem:weaker_assumption_6}
    The bound in Proposition~\ref{prop:pvi} is established on the weaker assumption that, for $V_{\theta_0} \in \cV_\Theta$, the limit $\lim_{t \rightarrow \infty}(\Pi^\sigma_\Theta B)^t V_{{\theta}_0}$ exists and there exists $\modulus<1$ such that
    \begin{align*}
        \tnorm{\left(\Pi^\sigma_\Theta B\right)^{t+1} V_{{\theta}_0} - \Pi^\sigma_\Theta V^*} \leq \modulus \tnorm{\left(\Pi^\sigma_\Theta B\right)^{t} V_{{\theta}_0} - V^*}
    \end{align*}
    for all $t \geq 0$. Such weaker assumption does away with the uniqueness of the limit $\tilde{V}^*$. We will re-visit this weaker assumption in \S\ref{sec:pvi_experiments}. In general, it is known that Assumption~\ref{assum:contraction} does not hold for the general discounted MDP in sup-norm \cite[Section 3]{Farias2000FixedPoints}. However, varying the weights $\tau$ and restricting to CO MDPs, we only require contraction on the PVI iterates.
\end{remark}

%%%%%%%%%%%%%%%%%%%%%%%%%%%%%%%%%%%%
%%%%%%%%%%%%%%%%%%%%%%%%%%%%%%%%%%%%

\subsection{Affine Approximation}
\label{sec:pvi_experiments}

The importance of the affine approximation scheme transcends that of a basic or toy example: it serves as a natural choice of state embedding -- the first embedding layer of a transformer, for example \eqref{eq:transformer_solver}. 

Given $K \ll \abs{\cS}$, $K=K'+1$ and a closed and convex set $\Theta \subseteq \Set{\theta=(\theta^0,\theta^1,\ldots,\theta^{K'}) \in \R^{K} \mid \theta^0 = 1}$, one can define an {\it affine approximation scheme} by $v(x,\theta) = x^\top\theta$. Specifically, defining $\phi(s_\infty) = \bm{0}^\top$, $V_{{\theta}}(s) = \sum_{i=0}^{K'} \phi_i(s) \theta^i$, for all $s\in \cS$. Let $\Phi$ denote the $\abs{\cS} \times K$ matrix defined by
\begin{align}
\label{eq:Phi}
    \Phi_{s,i} = %\phi_i(s),\quad \forall s \in \cS, i\in \langle K\rangle.
    \begin{cases}
       \phi_i(s) & s \in \cS \setminus \{s_\infty\} \\
       0 & s = s_\infty
    \end{cases}
\end{align}
for $s \in \cS$ and $i=0,\ldots,K'$. We denote its $i$-th column by $\phi_i \in \R^{\abs{\cS}}$ and its $s$-row, constituting the state's embedding, 
by $\phi(s) \in \R^{K}$. The column vector $\phi_0$ serves as the bias. 

Consider $\sigma \in \Delta_S$ with $\supp(\sigma) = \cS \setminus \{s_\infty\}$. Under an %compact 
affine approximation scheme, $\cV_\Theta$ is closed and convex and $\Pi^\sigma_\Theta$ \eqref{eq:proj} is uniquely determined. It is well known \cite{Stewart1989Projection} that when $\Theta = \{\theta=(\theta^0,\theta^1,\ldots,\theta^{K'}) \in \R^{K} \mid \theta^0 = 1\}$ and $\rk{\Phi} = {K}$ then $\Phi^\top \diag(\sigma) \Phi$ is invertible and $\Pi^\sigma_\Theta V = V_{\Phi^\dagger_\sigma V}$ for the $\sigma$-weighted Moore-Penrose pseudoinverse 
\begin{align}
\label{eq:weigted_pseudoinverse}
    \Phi^\dagger_\sigma \triangleq \left(\Phi^\top \diag(\sigma) \Phi\right)^{-1} \Phi^\top \diag(\sigma).
\end{align}

%%%%%%%%%%%%%%%%%%%%%%%%%%%%%%%%%%%%
%%%%%%%%%%%%%%%%%%%%%%%%%%%%%%%%%%%%

Since our analytical scheme is based on the convergence of PVI \S\ref{sec:pvi}, we tested the validity of Assumption~\ref{assum:contraction} for the affine approximation scheme 
by running PVI 
using $\Pi^\sigma_\Theta$ defined thereof.\footnote{This uses $\Theta \subseteq \Set{\theta=(\theta^0,\theta^1,\ldots,\theta^{K'}) \in \R^{K} \mid \theta^0 = 1}$ contrary to our assumption of compactness for $\Theta$. However, compactness was necessary for $\Pi^\sigma_\Theta$ to be well-defined and is therefore not necessary here.
} To that end, we focused on two combinatorial optimization problem (COP) types: KSP (Example~\ref{ex:ksp}) and TSP \S\ref{sec:TSP}. We considered three problem dimensions per problem type (KSP: $d=10,14,18$; TSP: $d=8,10,12$).
Problem dimensions were limited by the prohibitive size of the state space, influencing both computational and memory requirements. However, though PVI is not intended be used in practice, PVI convergence is necessary in our analytical framework.

To run PVI in a variety of scenarios, we sampled $50$ problem instances for each COP type (KSP/TSP) and dimension $d$ (for further details on COP instance generation, see Appendix~\ref{app:experiments}). For each sampled COP instance, we considered two embedding dimensions $K=d/2$ and $d$. For each instance and a choice of $K$, we considered $50$ randomly generated distributions $\sigma$ on the relevant state space. Thus, we had a total of $2500$ scenarios to check for Assumption~\ref{assum:contraction} per problem type, problem dimension ($d$) and embedding dimension ($K$). For each one of the $2500$ generated scenarios $(\mathrm{COP},d,K,\sigma)$, we sampled $50$ triplets $\tau$ \eqref{eq:tau_norm}, $\Phi$ \eqref{eq:Phi}, and initial $\theta_0$ to test contraction in PVI.

The contraction modulus for each one of the $1.5$mil PVIs we ran was estimated as follows. We randomly generated an initial value function $V_{\theta_0}$ and let PVI run for $T$ iterations (depending on problem type and dimension). Setting $\epsilon=10^{-12}$, for each PVI sequence $\Set{V_{\theta_t}}_{t=0}^{T-1}$ \eqref{eq:pvi},
we first calculated the minimal $t_* \in [0,T-2]$ for which $\tnorm{V_{\theta_{t'+1}}-V_{\theta_{t'}}} \leq \epsilon$ for all $t' \in [t_*,T-2]$. When there was no such $t_*$, we set $t_*=T-2$. Following Remark~\ref{rem:weaker_assumption_6}, we calculated
\begin{align}
    \gamma = \max \left\{ \max_{1\leq t \leq t_*-1} \left\{ \frac{\tnorm{V_{\theta_{t+1}}-V_{\theta_{t}}}}{\tnorm{V_{\theta_{t}}-V_{\theta_{t-1}}}}\right\}, \max_{0\leq t \leq t_*-1} \left\{ \frac{\tnorm{V_{\theta_{t+1}}-\Pi_\Theta^\sigma V^*}}{\tnorm{V_{\theta_{t}}-V^*}} \right\}\right\}.
\label{eq:gamma_experiment}
\end{align}
Since a contractive $\Pi_\Theta^\sigma B$ in each PVI run necessitates $\gamma<1$, we %therefore 
took $\gamma$ in \eqref{eq:gamma_experiment} as indicative of contraction whenever it was found to be $<1$. Under Assumption~\ref{assum:contraction} per scenario $(\mathrm{COP},d,K,\sigma)$, we calculated the ratio $\chi \in \Set{\frac{k}{50}}_{k=0}^{50}$ of contractive PVIs. The statistics for $\chi$ are presented in Table~\ref{tab:experiment_summary}.

\begin{table}[t]
\centering
\footnotesize
% \captionsetup{font=footnotesize}
\setlength{\tabcolsep}{6pt}
\renewcommand{\arraystretch}{1.15}
\begin{tabular}{lllcccc}
\toprule
\textbf{COP} & \textbf{d} & \textbf{K} & \textbf{Mean} & \textbf{Min} & \textbf{Skewness} & \(\bm{(q_{0.95}, q_{0.50}, q_{0.25})}\) \\
\midrule
\multirow{7}{*}{KSP} & \multirow{2}{*}{10}  & 5 & 0.896 [0.893,                        0.899] & 0.62 & -0.731 [-0.818,-0.647]
                    & \((0.98,\,0.90,\,0.86)\) \\
                    &                       & 10 & 0.747 [0.742, 0.753] & 0.32 & -0.590 [-0.661,-0.520] & \((0.92,\,0.76,\,0.66)\) \\
\cmidrule(lr){3-7}
                     & \multirow{2}{*}{14}  & 7 & 0.908 [0.906, 0.911] & 0.62 & -1.016 [-1.102,-0.931]& \((1.00,\,0.92,\,0.86)\) \\
                     &                      & 14 & 0.800 [0.795, 0.805] & 0.36 & -0.652 [-0.722,-0.581] & \((0.96,\,0.84,\,0.70)\) \\
\cmidrule(lr){3-7}
                     & \multirow{2}{*}{18}  & 9 & 0.886 [0.882, 
                     0.891] & 0.40 & -1.444 [-1.556,-1.335] & \((1.00,\,0.92,\,0.84)\) \\
                     &                      & 18 & 0.764 [0.757, 0.770] & 0.16 & -0.939 [-1.014,-0.867] & \((0.96,\,0.80,\,0.66)\) \\
\midrule
\multirow{7}{*}{TSP} & \multirow{2}{*}{8}  & 4 & 0.952 [0.951, 
                     0.953] & 0.82 & -0.690 [-0.791,-0.585]
                     & \((1.00,\,0.96,\,0.94)\) \\
                     &                      & 8 & 0.893 [0.891, 
                     0.895] & 0.66 & -0.530 [-0.638,-0.422]
                     & \((0.96,\,0.90,\,0.86)\) \\
\cmidrule(lr){3-7}
                     & \multirow{2}{*}{10} & 5 & 0.988 [0.988, 
                     0.989] & 0.92 & -1.227 [-1.329,-1.120] & \((1.00,\,1.00,\,0.98)\) \\
                     &                     & 10 & 0.976 [0.976, 
                     0.977] & 0.86 & -0.873 [-1.002,-0.753] & \((1.00,\,0.98,\,0.96)\) \\
\cmidrule(lr){3-7}
                     & \multirow{2}{*}{12} & 6 & 0.997 [0.997, 
                     0.997] & 0.94 & -2.429 [-2.711, 
                     -2.185] & \((1.00,\,1.00,\,1.00)\) \\
                     &                      & 12 & 0.994 [0.993, 
                     0.994] & 0.94 & -1.589 [-1.721,-1.464] & \((1.00,\,1.00,\,0.98)\) \\
\bottomrule
\end{tabular}
\vspace{1ex}
\caption{For each $(\mathrm{COP}, d, K)$, we report the Mean, Min, and Skewness for $\chi$ -- the ratio of contractive PVIs per scenario $(\mathrm{COP}, d, K, \sigma)$; Skewness is Fisher–Pearson (bias-corrected); Mean and Skewness are reported with a 95\% percentile bootstrap CI (4k resamples). To capture concentration near~$1$, we further include empirical quantiles $q_{0.95}, q_{0.50}, q_{0.25}$. All summaries use $n=2500$ groups per $(\mathrm{COP}, d, K)$.}
\label{tab:experiment_summary}
\end{table}

Table~\ref{tab:experiment_summary} demonstrates that there were no scenarios $(\mathrm{COP},d,K,\sigma)$ in which Assumption~\ref{assum:contraction} failed, namely, in each scenario, we were able to generate an affine approximation scheme and a $\tau$ that produced contraction and hence (empirical) convergence of the sequence of iterates. Mean, skewness, and quantiles, $q_{0.95}, q_{0.50}, q_{0.25}$, suggest mass concentration near $1$, indicating that for almost all problem instances most of the generated tuples resulted in a contraction. Moreover, non-converging instances were made rarer as we decreased the precision $\epsilon$ down to $10^{-4}$ and rarer still as we increased problem dimension $d$ (KSP: $0.203\%$ ($d=10$), $0.008
\%$ ($d=14$), $0.000\%$ ($d=18$); TSP: $0.000\%$ in all dimensions).

If we think of $\chi$ as the random variable defined by the chance to sample contractive $(\Phi,\tau,\theta_0)$ per instance $(\mathrm{COP},d,K,\sigma)$, it depends on problem type, problem dimension, and embedding dimension in likely intricate ways. The summary statistics suggest a drift away from $1$ as $d$ and $K$ increase. The former is to be expected. The latter suggests that a lower $K$ increases contraction as 
it produces a greater ``crunch'' in the projected Bellman operator. We believe the improved statistics for TSP as compared to KSP can be explained by the relation $\abs{\cS}/K$ which is $O(d)$ for KSP and $O(2^d)$ for TSP. At the same time, when PVI did converge, we did not witness a decisive benefit for higher embedding dimensions $K$ in terms of the relative optimality gap (see \eqref{eq:rel_opt_gap} in \S\ref{app:experiments}).
We suggest that the decisive factor affecting the relative optimality gap is $O(\abs{\cS}/K)$, which is identical, in our experiments, for both high and low choices of $K$.

Finally, we used Proposition~\ref{prop:pvi} as a sanity check for the $\gamma$ value \eqref{eq:gamma_experiment} calculated in each contractive PVI run. Indeed, in each problem instance $(\mathrm{COP},d)$, we found not only that \eqref{eq:proj_vi_rate} holds, but it also provided a tight bound (see Fig.~\ref{fig:experiment_proposition} in \S\ref{app:experiments}).

%%%%%%%%%%%%%%%%%%%%%%%%%%%%%%%%%%%%
%%%%%%%%%%%%%%%%%%%%%%%%%%%%%%%%%%%%

\subsection{Estimation Procedures}
\label{sec:estimation}

Recall that we have $\sigma \in \Delta_S$ with $\supp(\sigma) = \cS \setminus \{s_\infty\}$ and consider the $\sigma$-projected Bellman map $\Pi^\sigma_\Theta B: \cV_S^0 \rightarrow \cV_\Theta$ for a given approximation scheme \S\ref{sec:pvi}. Under the conditions of Proposition~\ref{prop:pvi}, $\Pi^\sigma_\Theta B$ is assumed to be a contraction of modulus $\modulus$ w.r.t. $\tnorm{\cdot}$, and the $\sigma$-PVI converges to the fixed point $\tilde{V}^*$ \ref{eq:pvi_fixed_pt} of the $\sigma$-projected Bellman map irrespective of initialization.

We typically cannot calculate $\Pi^\sigma_\Theta B$ directly, due to state space size, and must therefore resort to some estimation procedure. Assume  we have a procedure that yields an estimate $\tilde{V}_{t+1} \triangleq V_{\tilde{\theta}_{t+1}}$ of $\Pi^\sigma_\Theta B \tilde{V}_t$ at each step $t$ of $\sigma$-PVI initialized with $\tilde{V}_0 \equiv V_{\theta_0}$. The $\tau$-normed difference between the $\sigma$-projected Bellman improvement and its estimate,
\begin{align}
\label{eq:estimation_error}
    \tnorm{\tilde{V}_{t+1}-\Pi^\sigma_\Theta B\tilde{V}_t},
\end{align}
is the revolving {\it estimation error} for the estimation procedure. Works such as \cite{Munos2005AVI} regard the difference $\tilde{V}_{t+1}-B\tilde{V}_t$, while the estimation procedure is applied to $B\tilde{V}_t$. Such an approach absorbs the approximation error into AVI error propagation. Considering $\Pi^\sigma_\Theta B$ instead, we keep approximation and estimation errors separate. The following proposition, adapted from \cite[Lemma 4]{Weng1991FixedPoint}, relates the revolving estimation errors to the convergence of $\sigma$-PVI estimates to $\tilde{V}^*$.

\begin{restatable}{prop}{estimationConvergence}
\label{prop:estimation_convergence}
    Denote the estimation error by $\epsilon_t := \tnorm{\tilde{V}_{t+1}-\Pi^\sigma_\Theta B\tilde{V}_t}$
    and assume $\epsilon_t \rightarrow 0$. Then, under Assumption~\ref{assum:contraction}, the sequence of estimates $\left\{ \tilde{V}_t\right\}$ converges to $\tilde{V}^*$ \eqref{eq:pvi_limit} in the $\tau$-weighted $\ell_\infty$ norm.
\end{restatable}

% \OD{A ubiquitous estimation procedure involves the use of stochastic gradient descent. Our analysis to follow will restrict to this case.} 

%%%%%%%%%%%%%%%%%%%%%%%%%%%%%%%%%%%%
%%%%%%%%%%%%%%%%%%%%%%%%%%%%%%%%%%%%

\subsection{SAA through the Lens of RL}
\label{sec:bias_variance}

\begin{algorithm}[t]
\caption{Fitted Q-Iteration \cite{Riedmiller2005FittedQ}}
\label{alg:fqi}
\begin{algorithmic}[1]
\State \textbf{input:} $T \in \N$, $\sigma\in\Delta_{\cS \times \cA}$, $\tilde{\theta}_0 \in \Theta$, $\delta \in (0,1]$.
\For{$t = 0,\ldots,T-1$}
    \State sample $\left\{(s_i,a_i,s_i')\right\}_{i=1}^n$ i.i.d. from $p_\sigma$ 
    \State $y_i := r(s_i,a_i) + \delta \max_{a' \in \cA} {Q}_{\tilde{\theta}_t}(s_i',a')$
    \State $f_{t,n}(\theta) := \frac{1}{n} \sum_{i=1}^n \left(y_i - Q_{\theta}(s_i,a_i)\right)^{2}$ \label{step:fqi_saa}
	\State estimate $\argmin_{\theta\in\Theta} f_{t,n}(\theta)$ to get $\tilde{\theta}_{t+1}$
\EndFor
\State \textbf{output:} $Q_{\tilde{\theta}_T}$
\end{algorithmic}
\end{algorithm}

So far we discussed estimation procedures without commitment to any particular one. One specific choice, namely SAA, arises when we bridge the gap between PVI and DQL with FVI and FQI forming the middle links. This joins in the second thread of this work that begins with empirical evidence in favor of RL for CO and culminates with FVI.

DQL couples value updates with an explicit exploration–exploitation mechanism: a continually updated behavior policy generates new transitions stored in a replay buffer, so the effective state–action distribution evolves over time and progressively concentrates on regions visited under -- and favored by -- the current high-value policy. In this work, however, we are not concerned with the resulting dynamics of this state–action distribution, but rather with the process by which the target policy is learned. Stripping away the exploration–exploitation machinery leaves us with FQI (Algorithm~\ref{alg:fqi}) \cite{Riedmiller2005FittedQ}, which assumes a given sampling distribution $\sigma$ over state–action pairs. This section is dedicated to the link between PVI and FQI via FVI from which SAA emerges.

Consider a general MDP $\left( \cS, \cA, p, r\right)$ with discount factor $\delta \in (0,1]$ together with a $Q$-function $\tilde{Q}: \cS \times \cA \rightarrow \R$.
Define the loss function $f: \Theta \times \cS \times \cA \times \cS \rightarrow \R$ in relation to $\tilde{Q}$ to be $f(\theta,s,a,s') \triangleq \left(y(s,a,s') - Q_{\theta}(s,a)\right)^{2}$, where $y(s,a,s') \triangleq r(s,a) + \delta \max_{a'\in\cA}\tilde{Q}(s',a')$ \cite{Fan2020DQL}. By definition \eqref{eq:BQ}, we have
\begin{align}
\label{eq:BQy}
    B\tilde{Q}(s,a) = \E_{s'\sim\sigma}[y(s,a,s')].
\end{align}
Pick $\sigma \in \Delta_{\cS \times \cA}$ and consider the distribution on $\cS \times \cA \times \cS$ determined by the density $p_\sigma(s,a,s') \triangleq \sigma(s,a) p(s'|s,a)$. 
By the bias-variance decomposition \cite{Fan2020DQL},
\begin{align}
\label{eq:q_bias_variance}
\E_{p_\sigma}[f(\theta,s,a,s')] = \E_{p_\sigma}\left[ \left(y(s,a,s')-B\tilde{Q}(s,a)\right)^{2}\right] + \snorm{B\tilde{Q}-Q_{\theta}}^{2}
\end{align}
The variance term on the RHS does not involve $\theta$, hence, 
\begin{align}
\label{eq:argmin_equality_q_function}
    \argmin_\theta \E_{p_\sigma}[f(\theta,s,a,s')] = \argmin_\theta \left\Vert B\tilde{Q}-Q_{\theta}\right\Vert _{\sigma}^{2}
\end{align}
By \eqref{eq:BQy}, FQI uses SAA (Line~\ref{step:fqi_saa}) to get a (biased) estimate of the RHS of \eqref{eq:argmin_equality_q_function} for $\tilde{Q} \equiv Q_{\tilde{\theta}_t}$ and, thus,  ultimately, of the expected loss. 

Switching from $Q$-functions to value functions, 
we consider a given value function $\tilde{V}: \cS \rightarrow \R$ and define the loss function $f: \Theta \times \cS \times \cS^{\cA} \rightarrow \R$ in relation to $\tilde{V}$ to be
\begin{align}
\label{eq:fvi_loss}
    f(\theta,s,\xi) \triangleq \left(y(s,\xi)-V_\theta(s)\right)^{2},
\end{align}
where $y(s,\xi) \triangleq \max_{a \in \cA} \left\{  r(s,a) + \delta \tilde{V}(\xi(a)) \right\}$. Now pick $\sigma \in \Delta_S$ and consider $p_\sigma$ to be the distribution over $\cS \times \cS^{\cA}$ whose marginalization over $\cS^{\cA}$ produces $\sigma$ such that the probability of $\xi \in \cS^{\cA}$ mapping $a$ to $s'$ conditioned on $s \in \cS$ is $\Pr[\xi(a) = s'|s] = p(s'|s,a)$.
One can similarly work out the loss function bias-variance decomposition.

\begin{restatable}{lemma}{BiasVarainace}
\label{lem:bias_var}
    For $\sigma \in \Delta_S$ and $p_\sigma$ and $f$
    defined above, we have
    \begin{align}
    \label{eq:bias_var}
        \E_{p_\sigma}[f(\theta,s,\xi)] &= \E_{p_\sigma} \left[ \left(y(s,\xi)-B\tilde{V}(s)\right)^{2}\right] + \nonumber\\ & \quad 2\E_{\sigma} \left[ \left(B\tilde{V}(s)-V_{\theta}(s)\right) \E_{\xi|s}\!\left[y(s,\xi)-B\tilde{V}(s)\right]\right] + \snorm{B\tilde{V}-V_{\theta}}^{2}
    \end{align}
\end{restatable}

Focusing on the middle term on the RHS of \eqref{eq:bias_var}, for $s \in \cS$, we have
\begin{align*}
    \E_{\xi|s}\!\left[y(s,\xi)-B\tilde{V}(s)\right]
     &  = \E_{\xi|s}\left[y(s,\xi)\right] - B\tilde{V}(s) \\
     &  = \E_{\xi|s}\!\left[ \max_{a \in \cA} \left\{  r(s,a) + \delta \tilde{V}(\xi(a)) \right\} \right] - B\tilde{V}(s) 
\end{align*}
where ${\xi|s \sim \prod_{a \in \cA} p(\cdot|s,a)}$. For that middle term to cancel, we need
\begin{align}
\label{eq:make_bias_var_work}
    \E_{\xi|s}\!\left[ \max_{a \in \cA} \left\{  r(s,a) + \delta \tilde{V}(\xi(a)) \right\} \right]
    & \overset{!}{=} \max_{a \in \cA} \left\{ r(s,a) + \delta \sum_{s' \in \cS} p(s'|s,a) \tilde{V}(s') \right\} 
\end{align}
Expectation and maximum generally do not commute. When transitions are deterministic, however, we have $\Pr_{p_\sigma}[\xi(a)=s'|s]=\delta_{\lambda(s,a),s'}$, which determines $\xi|s$ with probability $1$, and we get equality in \eqref{eq:make_bias_var_work}. At the same time, the value-function analogue of \eqref{eq:q_bias_variance} trivializes to
\begin{align}
\label{eq:argmin_equality_value_func}
\E_{p_\sigma}[f(\theta,s,\xi)] = {\E_{p_\sigma} \left[ \left(y(s,\xi)-B\tilde{V}(s)\right)^{2}\right]} + \snorm{B\tilde{V}-V_{\theta}}^{2} = \snorm{B\tilde{V}-V_{\theta}}^{2}
\end{align}

This allows us to consider FVI (Algorithm~\ref{alg:fvi}) \cite{Munos2008FVI} as the value-function analogue of FQI (Algorithm~\ref{alg:fqi}) suitable for MDPs with deterministic transitions such as CO MDPs (where $\delta=1$). Similarly, it uses SAA (Line~\ref{step:fvi_saa}) to get a (biased) estimate for the RHS of \eqref{eq:argmin_equality_value_func}, namely, the projected improvement in the approximation scheme. Importantly, the SAA optimal solution (Line~\ref{step:fvi_estimate}) and its distance from $\theta^*$ that minimizes \eqref{eq:argmin_equality_value_func} will determine the accuracy of $V_{\theta}$ obtained for this procedure.

\begin{algorithm}[t]
\caption{Fitted Value Iteration \cite{Munos2008FVI}}
\label{alg:fvi}
\begin{algorithmic}[1]
\State \textbf{input:} $T \in \N$, $\sigma\in\Delta_{S}$, $\tilde{\theta}_0 \in \Theta$, $\delta \in (0,1]$.
\For{$t = 0,\ldots,T-1$}
    \State sample $\{s_i\}_{i=1}^n$ i.i.d. from $\sigma$
    \State $y_i := \max_{a \in \cA} \left\{ r(s_i,a) + \delta{V}_{\tilde{\theta}_t}(s'_{i,a}) \right\}$ where $p(s'_{i,a}|s_i,a)=1$
    \State $f_{t,n}(\theta) := \frac{1}{n}\sum_{i=1}^n \left(y_i - V_{\theta}(s_{i})\right)^{2}$ \label{step:fvi_saa}
	\State estimate $\argmin_{\theta\in\Theta} f_{t,n}(\theta)$ to get $\tilde{\theta}_{t+1}$ \label{step:fvi_estimate}
\EndFor
\State \textbf{output:} $V_{\tilde{\theta}_T}$ 
\end{algorithmic}
\end{algorithm}

%%%%%%%%%%%%%%%%%%%%%%%%%%%%%%%%%%%%
%%%%%%%%%%%%%%%%%%%%%%%%%%%%%%%%%%%%

\subsection{Sample Average Approximation}
\label{sec:saa}

Consider a sufficiently expressive (Definition~\ref{def:sufficiently_expressive}), compact approximation scheme $\cV_\Theta$ with a continuous $v: \R^K \times \Theta \rightarrow \R$ such that $\Theta \subset \R^N$ is compact and convex. We assume, in addition, that $v(\phi(s),\cdot)$ is Lipschitz continuous over $\Theta$ for each $s \in \cS\setminus \{s_\infty\}$ with Lipschitz constant $L(s)$, namely,
\begin{align}
    \label{eq:L}
    \abs{V_\theta(s)-V_{\theta'}(s)} \equiv 
    \abs{v(\phi(s),\theta)-v(\phi(s),\theta')} 
    \leq L(s) 
    \dist\left(\theta,\theta'\right),
\end{align}
which extends, by definition, to $S$ with $L(s_\infty)=0$. Let $\sigma \in \Delta_S$ with $\supp(\sigma) = \cS \setminus \{s_\infty\}$. Following Proposition~\ref{prop:pvi}, assume $\Pi^\sigma_\Theta B$ is a contraction of modulus $\modulus < 1$ w.r.t. $\tnorm{\cdot}$ for some $\tau: \cS \rightarrow (0,\infty)$ \eqref{eq:tau_fn}. Under these assumptions (or, possibly, additional ones), our goal is to establish a convergence theorem for FVI (Algorithm~\ref{alg:fvi}) in the undiscounted case ($\delta=1$) which suits our CO use case.

As previously noted (\S\ref{sec:bias_variance}), at each iteration $t$, FVI uses SAA (Line~\ref{step:fvi_saa}) to estimate $f_t(\theta)$, defined as the $\sigma$-expectation
\begin{align}
\label{eq:dg_loss_fn}
    & f_t(\theta) \triangleq \E_{\sigma}[f_t(\theta,s)] \equiv \snorm{BV_{\tilde{\theta}_t}-V_\theta}^2
\end{align}
of the sample function
\begin{align}
\label{eq:sample_fn}
    & f_t(\theta,s) \triangleq \left(BV_{\tilde{\theta}_t}(s) - V_\theta(s)\right)^2.
\end{align}
SAA thus serves to estimate $\sigma$-projected improvement in the given compact approximation scheme,
\begin{align*}
    & \argmin_{\theta\in\Theta} f_t(\theta) = \argmin_{\theta\in\Theta} \snorm{BV_{\tilde{\theta}_t}-V_\theta}^2.
\end{align*}
We denote the sample average random variable by
\begin{align}
\label{eq:loss_sample_average}
    & {f}_{t,n}(\theta) = \frac{1}{n}\sum_{i=1}^{n} f_t(\theta,s_i).
\end{align}
for i.i.d. $\sigma$-samples $\{s_i\}_{i=1}^n$. Before we continue, we will establish some basic properties of $f_t(\theta)$ and $f_t(\theta,s)$ that will serve us later. 

\begin{restatable}{lemma}{ShapiroCond}
\label{lem:Shapiro_cond}
    Consider a compact approximation scheme $v: \R^K \times \Theta \rightarrow \R$ such that $\Theta$ is compact and convex, $v$ is continuous, and $v(\phi(s),\cdot)$ is Lipschitz continuous over $\Theta$ for each $s \in \cS$. Then,
    \begin{enumerate}[label=(\roman*)]
    \item $f_t(\theta)$ is finite for all $\theta\in \Theta$;
    \item for any $\theta,\theta'\in \Theta$ the r.v. $Y^{\theta,\theta'}_t\!(s) = \left(f_t(\theta,s)-f_t(\theta)\right) - \left(f_t(\theta',s)-f_t(\theta')\right)$ is $\omega_t$-sub-Gaussian for some $\omega_t$, namely, $\E_{\sigma}\left[e^{\zeta Y^{\theta,\theta'}_t\!(s)}\right]\leq e^{\frac{1}{2}\omega_t^{2}\zeta^{2}}{, \forall \zeta\in \R}$;
    \item there exists $\kappa_t:S\rightarrow\R_{\geq0}$ such that $\abs{f_t(\theta,s)-f_t(\theta',s)} \leq \kappa_t(s) d\left(\theta,\theta '\right)$, for every $\theta,\theta'\in\Theta$, and its moment-generating function $M_{\kappa_t}(\zeta):=\E_{\sigma}\left[e^{\zeta\kappa_t(s)}\right]$ is finite $\forall \zeta \in \R$;
    \item for any $\hat{\kappa}_t > \E_{\sigma}\left[\kappa_t(s)\right]$ there exists a corresponding $\beta_t:=\beta(\hat{\kappa}_t)$ such that $\Pr\left[\overline{\kappa}_{t,n} \geq \hat{\kappa}_t\right] \leq e^{-n\beta_t}$ for the sample average $\overline{\kappa}_{t,n} = \frac{1}{n} \sum_{i=1}^n \kappa_t(s_i)$. In particular, for $\hat{\kappa}_t=\kappa_t^*+\delta$, where $\delta>0$ and $\kappa_t^*:=\max_{s\in S}\kappa_t(s)$, we have $\beta_t(\hat{\kappa}_t)=\infty$.
    \end{enumerate}
\end{restatable}

Let us denote the optimal solutions of $f_t(\theta)$ and $f_{t,n}(\theta)$ in $\Theta$ and their optimal values by \begin{align*}
    & \theta_t^* \in \Theta_t^* \triangleq \argmin_{\theta\in \Theta} f_t(\theta), \quad f^*_t = f_t(\theta_t^*), \\
    & \theta_{t,n}^* \in \Theta_{t,n}^* \triangleq \argmin_{\theta\in\Theta} f_{t,n}(\theta), \quad f^*_{t,n} = f_{t,n}(\theta_{t,n}^*).
\end{align*}
We will also denote the compact subset of $\epsilon$-optimal solutions of $f_t(\theta)$ by
\begin{align*}
    \Theta_{t,\epsilon}^* \triangleq \Set{\theta\in\Theta \mid f_t(\theta) \leq f_t^* + \epsilon}
\end{align*}
for any $\epsilon > 0$. The optimal solution $\theta_t^*$ satisfies an equivalent form of the Pythagorean theorem for $\sigma$-projected Bellman improvement.

\begin{restatable}{lemma}{projectionAngle}
\label{lem:projection_angle}
    Assume $\cV_\Theta \subset \cV^0_S$ is convex. Then, for any $\theta \in \Theta$, we have $\snorm{V_{\theta} - V_{\theta_t^*}}^2 \leq f_t(\theta) - f_t^*$.
\end{restatable}

By Lemma~\ref{lem:Shapiro_cond}(\romannumeral 3), $f_t(\cdot,s)$ is Lipschitz continuous for each $s \in \cS$. It then follows that $\{f_t(\theta,s)\}_{\theta\in\Theta}$ is a uniformly integrable family of random variables. Thus, $f^*_{t,n}$ converges to $f^*_t$ as $n \rightarrow \infty$ w.p.$1$ \cite[Proposition 8]{kim2014guide}, and, in particular, $\E_{\sigma^n}\!\left[f^*_{t,n}\right]$ converges to $f^*_t$. Moreover, \cite[Theorem 5.3]{shapiro2021lectures} states that $d\left(\Theta_{t,n}^*,\Theta_t^*\right)$ converges to $0$ w.p.$1$ as $n \rightarrow \infty$. To establish a convergence theorem for FVI, however, we are interested in the rate -- established in terms of number of samples $n$ -- at which an estimate $\tilde{\theta}_{t+1}$ of $\theta_{t,n}^*$ (Line~\ref{step:fvi_estimate} of FVI), obtained by some method, converges to the optimal solution set $\Theta_t^*$.

Note that when the dimension of $\theta$ is large, SAA is often solved by using first-order methods. Here we take
{\it projected gradient descent} (PGD) as our estimation method in Line~\ref{step:fvi_estimate}. Accordingly, to establish our convergence result for FVI, we will determine the minimal number of samples $n$ and PGD steps $c$ required at each iteration. Having assumed the approximation scheme is sufficiently expressive, we will initialize Algorithm~\ref{alg:fvi} with $V_{\tilde{\theta}_0} \in \mathcal{B}_\rho^\tau(0)$, where $\rho$ is defined in Definition~\ref{def:sufficiently_expressive}.
% namely, with $V_{\tilde{\theta}_0} \in \cV_\Theta$ such that
% \begin{align}
% \label{eq:alg_init}
%     \tnorm{V_{\tilde{\theta}_0}} \leq \frac{R \sqrt{W} \modulus^\tau}{1-\modulus^\tau} \left( \frac{1}{\tau_\infty} - \frac{1}{\tau_0}\right),
% \end{align}
% which exists by the sufficient expressivity of $\cV_\Theta$ (Definition~\ref{def:sufficiently_expressive}). 
The following theorem merges the general result on estimation procedures from Proposition~\ref{prop:estimation_convergence} with the SAA  \cite{shapiro2021lectures} and PGD \cite{Beck2023} convergence properties.
\begin{restatable}{theorem}{FVI}
\label{thm:fvi}
    Assume $\cV_\Theta \subset \cV^0_S$ is convex and $f_t(\cdot,s)$ is convex for each $s \in \cS$. 
    Let $\{\epsilon_t\}_{t=0}^\infty$ be a monotonically decreasing sequence such that $0 < \epsilon_t < 1/4$ and $\lim_{t \rightarrow \infty} \epsilon_t = 0$. Let $\omega_t$, $\hat{\kappa}_t$, $\kappa_t^*$, $\bar{\kappa}_{t,n}$, and $\beta_t$ satisfy the properties in Lemma~\ref{lem:Shapiro_cond}, and let  $\rho$ be as stated in Definition~\ref{def:sufficiently_expressive}. Then, at each iteration $t = 0,\ldots,T-1$ of Algorithm~\ref{alg:fvi}, choosing a number of samples
    \begin{align}
    \label{eq:fvi_n}
        n_t \geq \max \left\{\frac{32\omega_t^2}{\epsilon_t^{2}} \left[N \ln\left(\frac{16 \nu D_{\Theta} \hat{\kappa}_t}{\epsilon_t}\right) + \frac{1}{2}\ln\left(\frac{1}{\epsilon_t}\right)\right], \frac{1}{2\beta_t} \ln\left(\frac{1}{\epsilon_t}\right)\right\},
    \end{align} and a number of PGD steps 
    \begin{align*}
        c_t \geq \frac{D_\Theta^2}{\eta_t \epsilon_t}
    \end{align*}
     with stepsize $\eta_t \in (0,1/{\kappa}^*_t]$ to obtain an estimate $\tilde{\theta}_{t+1} \in \Theta$, %at $t$, %will 
     guarantees
    \begin{align}
    \label{eq:fvi_error}
        \E\left[\tnorm{V_{\tilde{\theta}_{T}} -\tilde{V}^*}\right] 
        & \leq \min_{t=0,\ldots,T-1}\left\{\modulus^{T-t} \rho + \frac{\sqrt{\epsilon_t}}{(1-\modulus)} \left(2\tnorm{L} D_\Theta + \frac{1}{c(\sigma,\tau)}\right)\right\}.
    \end{align}
    In particular, Algorithm~\ref{alg:fvi} output, $V_{\tilde{\theta}_{T}}$, converges to $\tilde{V}^*$ in expectation as $T \rightarrow \infty$.
\end{restatable}

One important takeaway of Theorem~\ref{thm:fvi} is that as $\epsilon_t \rightarrow 0$ we get $n_t \rightarrow \infty$. In particular, for a sufficiently small $\epsilon_t$, we get $n_t = O(\abs{\cS})$, implying that the number of samples becomes prohibitive, effectively lower bounding the estimation error guarantee. It is also important to note that \eqref{eq:fvi_error} provides a guarantee on the expected error, which, nevertheless, can be translated into a guarantee in probability {by Markov's inequality}.

% \Pr[\tnorm{X_T-X} > \epsilon] \leq \frac{1}{\epsilon}\E[\tnorm{X_T-X}] \rightarrow 0 for every \epsilon and as T \rightarrow \infty

\begin{remark}
The lower bound on the number of samples $n_t$ \eqref{eq:fvi_n} in Theorem~\ref{thm:fvi} depends on a constant $\nu$ characterized by the geometry of $\Theta$. Explicitly, for $\Theta \subset \R^N$, $\nu$ is defined as a constant such that $\abs{\mathcal{N}_\delta(\Theta)} \leq \left(\nu D_\Theta\delta^{-1}\right)^N$
for a $\delta$-net $\mathcal{N}_\delta(\Theta)$ of $\Theta$. As an example, for $\Theta = [a,b]^N \subset \R^N$, {$D_\Theta=\sqrt{N}(b-a)$, and for} $0 < \delta < (b-a)$, we can take $\nu = 2/\sqrt{N}$, using the bound $\delta+(b-a)<2(b-a)$. For $\Theta = \mathcal{B}^2_\rho(\theta) \subset \R^N$, {$D_{\Theta}=2\rho$, and for} $0 < \delta < 2\rho$, since $B_{\rho}^2(0)\subset[-\rho,\rho]^N$,
we can take $\nu = 2$, following \cite[Lemma 5.2]{Vershynin2012} and using the bound $\delta+2\rho 
< 4\rho$.     
\end{remark}

\begin{remark}
\label{rem:fvi_consts}
    In Theorem~\ref{thm:fvi}, one can make a global choice of $\omega_t, \hat{\kappa}_t, \kappa_t^*$, $\beta_t$, and $\eta_t$ %in Theorem~\ref{thm:fvi} 
    that is independent of $t$. In fact, there exists $\kappa:S\rightarrow\R_{\geq0}$ such that $\left|f_t(\theta,s)-f_t(\theta',s)\right|\leq\kappa(s) \dist\left(\theta,\theta '\right)$ for every $\theta,\theta'\in\Theta$. This can be seen by calculating directly
    \begin{align*}
        \left|f_t(\theta,s)-f_t(\theta',s)\right| & = \abs{\left(V_\theta(s) - BV_{\tilde{\theta}_t}(s)\right) + \left(V_{\theta'}(s) - BV_{\tilde{\theta}_t}(s)\right)} \abs{V_\theta(s)-V_{\theta'}(s)}\\
        & \leq \abs{\left(V_\theta(s) - BV_{\tilde{\theta}_t}{(s)}\right) + \left(V_{\theta'}(s) - BV_{\tilde{\theta}_t}{(s)}\right)} L(s) \dist\left(\theta,\theta'\right)
    \end{align*}
    The expression $\abs{\left(V_\theta(s) - BV_{\tilde{\theta}_t}(s)\right) + \left(V_{\theta'}(s) - BV_{\tilde{\theta}_t}(s)\right)}$ can be bounded in $\Theta$, irrespective of $\tilde{\theta}_t$.

    \begin{restatable}{lemma}{VthetaStateBound}
    \label{lem:Vtheta_state_bound}
        $\abs{V_\theta(s) - BV_{\tilde{\theta}_t}(s)} \leq \left(1+\frac{\tau_0 \modulus^\tau}{\tau_{d_\Pi}}\right)\norm{L}_\infty D_\Theta + \frac{\tau_0 \modulus \left(1+\modulus^\tau\right)}{1-\modulus} \tnorm{V^* - \Pi^\sigma_\Theta V^*}$.
    \end{restatable}

    Thus, we can choose $\kappa(s) \triangleq  \zeta_\kappa L(s)$ for 
    \begin{align*}
        \zeta_\kappa \triangleq 2 \left[\left(1+\frac{\tau_0 \modulus^\tau}{\tau_{d_\Pi}}\right)\norm{L}_\infty D_\Theta + \frac{\tau_0 \modulus \left(1+\modulus^\tau\right)}{1-\modulus} \tnorm{V^* - \Pi^\sigma_\Theta V^*}\right]
    \end{align*}
    whose moment generating function, $M_{\kappa}(\zeta):=\E_{\sigma}\left[e^{\zeta\kappa(s)}\right]$, is also finite $\forall \zeta \in \R$. For any $\delta>0$, choosing $\kappa^*_t \equiv \kappa^* \triangleq \max_{s \in \cS} \kappa(s) = \zeta_\kappa \max_{s \in \cS} L(s) = \zeta_\kappa \norm{L}_\infty$ and $\hat{\kappa}_t \equiv \hat{\kappa} \triangleq \kappa^* + \delta$, we will have $\Pr\left[\overline{\kappa}_{n} \geq \hat{\kappa}\right] = 0$ for any sample average $\overline{\kappa}_{n} = \frac{1}{n} \sum_{i=1}^n \kappa(s_i)$, thus, $\beta_t \equiv \beta \triangleq \beta(\hat{\kappa}) = \infty$. Following Lemma~\ref{lem:Shapiro_cond}({\romannumeral 3}), we have $\left|f_t(\theta,s)-f_t(\theta',s)\right|\leq \zeta_\kappa L(s) D_\Theta$ and, consequently, $\abs{f_t(\theta)-f_t(\theta')} \leq \zeta_\kappa D_\Theta \E_\sigma[L(s)]$. Thus, $\abs{Y^{\theta,\theta'}_t\!(s)} \leq \zeta_\kappa D_\Theta \left(\norm{L}_\infty + \E_\sigma[L(s)] \vphantom{Y^{\theta,\theta'}_t}\right)$. As in Lemma~\ref{lem:Shapiro_cond}({\romannumeral 2}), we can use Hoeffding's Lemma to choose $\omega_t \equiv \omega \triangleq \zeta_\kappa D_\Theta \left(\norm{L}_\infty + \E_\sigma[L(s)] \vphantom{Y^{\theta,\theta'}_t}\right)$. Given all these choices, we can take
    \begin{align*}
        n_t {\leq  \frac{32 \omega^2}{\epsilon_t^{2}} \left[N \ln\left(\frac{16 \nu D_{\Theta} \hat{\kappa}}{\epsilon_t}\right) + \frac{1}{2}\ln\left(\frac{1}{\epsilon_t}\right)\right]+1.}
    \end{align*}
    {Moreover, for any $t$, one can always choose $\eta_t\in (\mu/\kappa^*,1/\kappa_t^*)$, for some $\mu\in(0,1)$ guaranteeing that
    \begin{align*}
        c_t\leq \frac{D_\Theta^2\kappa^*}{\mu \epsilon_t}+1.%\geq \frac{D_\Theta^2}{\eta \epsilon_t}
    \end{align*}}
\end{remark}

\begin{restatable}{coro}{CorFVI}
\label{coro:fvi}
    Let $\epsilon>0$. Following the global choice of constants $\omega_t \equiv \omega, \hat{\kappa}_t \equiv \hat{\kappa}, \kappa_t^* \equiv \kappa^*, \beta_t \equiv \beta$, and $\eta_t \equiv \eta \in (0,1/\kappa^*]$ in Remark~\ref{rem:fvi_consts}, while taking $\epsilon_t = {\epsilon_0}/{(1+\delta)^t}$, for $\epsilon_0 < 1/4$ and $\delta > 0$, and
    \begin{align*}
        T = \left\lceil a \log_{1+\delta}\left(\frac{\rho+\rho'}{\epsilon}\right) \right\rceil,
    \end{align*}
    for $\rho$ as in Theorem~\ref{thm:fvi} and
    \begin{align*}
        a & = - \left(\frac{2}{\log_{1+\delta}\modulus^*}\right) > 0, \quad
        \modulus^* = \max \left\{\modulus,\frac{1}{\sqrt{(1+\delta)}}\right\} < 1, \\
        \rho' &= \frac{\sqrt{\epsilon_0}}{(1-\modulus)} \left(2\tnorm{L} D_\Theta + \frac{1}{c(\sigma,\tau)}\right),
    \end{align*}
    Algorithm~\ref{alg:fvi} will output an $\epsilon$-approximation $V_{\tilde{\theta}_{T}}$ of $\tilde{V}^*$ \eqref{eq:pvi_limit} w.r.t. $\tnorm{\cdot}$ in expectation with sampling complexity
    \begin{align*}
        \sum_{t=0}^{T-1} n_t = O\left(\log_{1+\delta}\left(\frac{1}{\epsilon}\right)
        \left(\frac{1}{\epsilon}\right)^{2a}\right) 
    \end{align*}
    and first-order iteration complexity
    \begin{align}
    \label{eq:fvi_runtime}
        \sum_{t=0}^{T-1} c_t 
        = O\left(\left(\frac{1}{\epsilon}\right)^{a}\right).
    \end{align}      
\end{restatable}

{Following Corollary~\ref{coro:fvi} and Proposition~\ref{prop:pvi}, we find that, for each $\epsilon>0$, we can guarantee the following bound with high probability
\begin{align*}
    \tnorm{V_{\tilde{\theta}_{T}} -{V}^*} \leq \epsilon + \left(\frac{1}{1-\modulus}\right) \tnorm{V^* - \Pi^\sigma_\Theta V^* \vphantom{\lim_{t \rightarrow \infty} V_{{\theta}_t}}}.
\end{align*}
The second summand on the RHS cannot be made arbitrarily small without changing the approximation scheme itself. FVI runtime is given by \eqref{eq:fvi_runtime}. It is manifestly not polynomial in $1/\epsilon$, unless we can bound $\modulus$ away from $1$. Such a global bound, however, cannot be guaranteed for all CO problems in the same parametrized class. Moreover, the dependence of $\modulus$ on problem input size is unclear.
% For both these reasons, FVI does not constitute a fully polynomial-time approximation scheme (FPTAS). 
Nevertheless, $V_{\tilde{\theta}_{T}}$ is an approximate value function for the CO MDP that can be translated to an approximate solution to CO as stated in Theorem~\ref{thm:approximation}.}
% \textcolor{deepgreen}{[FPTAS losses its meaning as long as the latter part in not $0$. Also as far as I know FPTAS refers to deterministic algorithms while this is a not. There are randomized FPTAS which use randomization and provided expected value guarantee. I am not sure we want to get into these issues here. Also, I am not clear how our constants relate to the problem's input. I fear this paragraph can get a lot of criticism from people whose bread and butter is FPTAS/PTAS]} \OD{I get what you are saying. I did get a question in the ISDSA talk about how our work relates to FPTAS/PTAS. The purpose of of the reference to FPTAS is to be clear that this is not FPTAS. We can also add your comment that this is randomized and not deterministic. I do anticipate a question about the relation of our work to standard approximation schemes. But if you object, we can remove the sentence concerning FPTAS/PTAS.}

% if 0 < \modulus < 1-\epsilon', then a < -2/\log_{1+\delta}(1-\epsilon') for a proper choice of \delta.

%%%%%%%%%%%%%%%%%%%%%%%%%%%%%%%%%%%%
%%%%%%%%%%%%%%%%%%%%%%%%%%%%%%%%%%%%

\section{Summary and Outlook}
\label{sec:summary}

In this paper, we presented a unified framework to convert CO problems to equivalent undiscounted MDPs and analyze their solutions through value-based RL methods in terms of convergence, optimality gap, as well as sample and first-order complexity.

In our analysis, we did not discuss generalization, i.e., the problem of learning an approximate value function as a function of parameterized reward $c\in \cC$. Rather, our focus was the training processes, analyzing existing value-based RL methods for a fixed (possibly randomized) parameter realization $c$. Future work may build on our framework to analyze the generalization properties of methods such as DQL in solving CO problems. 

In our analysis (Theorem~\ref{alg:fvi}) we provide a bound on the expected weighted infinity norm of the error, i.e., the difference between the value function at the last iteration and the accumulation point. However, we do not necessarily need to be accurate on all states to get a good approximate solution. Future work may explore how the exploration-exploitation mechanism's focus on high-value states serves to mitigate the explosion in sample size necessary to decrease the error.

% embedding with contraction \S\ref{sec:pvi_experiments} will yield better results for algorithms modeled with NN architectures. 

% Stochastic gradient descent as opposed to SAA + PGD.

\clearpage

%%%%%%%%%%%%%%%%%%%%%%%%%%%%%%%%%%%%
%%%%%%%%%%%%%%%%%%%%%%%%%%%%%%%%%%%%

\bibliography{ref.bib}

@article{Karp1967Programming,
    author = {Karp, Richard M. and Held, Michael},
    title = {Finite-State Processes and Dynamic Programming},
    year = {1967},
    issue_date = {May 1967},
    publisher = {Society for Industrial and Applied Mathematics},
    address = {USA},
    volume = {15},
    number = {3},
    journal = {SIAM J. Appl. Math.},
    month = {May},
    pages = {693--718},
    numpages = {26}
}

@inproceedings{Berto2023RL4CO,
    title={{RL}4{CO}: A Unified Reinforcement Learning for Combinatorial Optimization Library},
    author={Federico Berto and Chuanbo Hua and Junyoung Park and Minsu Kim and Hyeonah Kim and Jiwoo Son and Haeyeon Kim and Joungho Kim and Jinkyoo Park},
    booktitle={NeurIPS 2023 Workshop: New Frontiers in Graph Learning},
    year={2023}
}

@techreport{Reddy1977Beam,
  author      = {Reddy, D. Raj},
  title       = {Speech Understanding Systems: Summary of Results of the Five-Year Research Effort at Carnegie-Mellon University},
  institution = {Department of Computer Science, Carnegie-Mellon University},
  address     = {Pittsburgh, PA},
  year        = {1976},
  month       = sep,
  note        = {CMU Computer Science Technical Report; also distributed as NTIS AD-A049 288},
}

@inproceedings{Kool2019Attention,
    author       = {Wouter Kool and Herke van Hoof and Max Welling},
    title        = {Attention, Learn to Solve Routing Problems!},
    booktitle    = {7th International Conference on Learning Representations, {ICLR} 2019, New Orleans, LA, USA, May 6-9},
    year         = {2019}
}

@article{Mnih2015DQN,
	author = {Mnih, Volodymyr and Kavukcuoglu, Koray and Silver, David and Rusu, Andrei A. and Veness, Joel and Bellemare, Marc G. and Graves, Alex and Riedmiller, Martin and Fidjeland, Andreas K. and Ostrovski, Georg and Petersen, Stig and Beattie, Charles and Sadik, Amir and Antonoglou, Ioannis and King, Helen and Kumaran, Dharshan and Wierstra, Daan and Legg, Shane and Hassabis, Demis},
	date = {2015/02/01},
	id = {Mnih2015},
	journal = {Nature},
	number = {7540},
	pages = {529--533},
	title = {Human-level control through deep reinforcement learning},
	volume = {518},
	year = {2015}
}

@article{Tseng1990Convergence,
    title={Solving {H}-horizon, stationary {M}arkov decision problems in time proportional to log({H})},
    author={Paul Tseng},
    journal={Operations Research Letters},
    year={1990},
    volume={9},
    pages={287--297}
}

@article{Tsitsiklis1996Feature,
    title={Feature-based methods for large scale dynamic programming},
    author={John N. Tsitsiklis and  Benjamin van Roy},
    journal={Machine Learning},
    year={1996},
    volume={22},
    issue={1},
    pages={59--94}
}

@article{Weng1991FixedPoint,
    title={Fixed Point Iteration for Local Strictly Pseudo-contractive Mapping},
    author={Xinlong Weng},
    journal={Proceedings of the American Mathematical Society},
    year={1991},
    volume={113},
    issue={3},
    pages={727--731}
}

@article{Pollack1960ShortestRoute,
    title={Solutions of the Shortest-Route Problem: A Review},
    author={Maurice Pollack and Walter Wiebenson},
    journal={Operations Research},
    year={1960},
    volume={8},
    issue={2},
    pages={224--230}
}

@article{Bengio2021Survey,
    title = {Machine learning for combinatorial optimization: A methodological tour d’horizon},
    journal = {European Journal of Operational Research},
    volume = {290},
    number = {2},
    pages = {405-421},
    year = {2021},
    author = {Yoshua Bengio and Andrea Lodi and Antoine Prouvost}
}

@inproceedings{Khalil2017,
    author = {Khalil, Elias B. and Dai, Hanjun and Zhang, Yuyu and Dilkina, Bistra and Song, Le},
    title = {Learning combinatorial optimization algorithms over graphs},
    year = {2017},
    publisher = {Curran Associates Inc.},
    address = {Red Hook, NY, USA},
    booktitle = {Proceedings of the 31st International Conference on Neural Information Processing Systems},
    pages = {6351--6361},
    numpages = {11},
    location = {Long Beach, California, USA},
    series = {NIPS'17}
}

@inproceedings{Drori2020,
    author={Drori, Iddo and Kharkar, Anant and Sickinger, William R. and Kates, Brandon and Ma, Qiang and Ge, Suwen and Dolev, Eden and Dietrich, Brenda and Williamson, David P. and Udell, Madeleine},
    booktitle={2020 19th IEEE International Conference on Machine Learning and Applications (ICMLA)}, 
    title={Learning to Solve Combinatorial Optimization Problems on Real-World Graphs in Linear Time}, 
    year={2020},
    pages={19--24},
}

@inproceedings{Bello2017NCO,
    author       = {Irwan Bello and Hieu Pham and Quoc V. Le and Mohammad Norouzi and Samy Bengio},
    title        = {Neural Combinatorial Optimization with Reinforcement Learning},
    booktitle    = {5th International Conference on Learning Representations, {ICLR} 2017, Toulon, France, April 24--26, 2017, Workshop Track Proceedings},
    year         = {2017}
}

@inproceedings{Janner2021RL,
 author = {Janner, Michael and Li, Qiyang and Levine, Sergey},
 booktitle = {Advances in Neural Information Processing Systems},
 editor = {M. Ranzato and A. Beygelzimer and Y. Dauphin and P.S. Liang and J. Wortman Vaughan},
 pages = {1273--1286},
 publisher = {Curran Associates, Inc.},
 title = {Offline Reinforcement Learning as One Big Sequence Modeling Problem},
 volume = {34},
 year = {2021}
}

@article{Mazyavkina2021Survey,
title = {Reinforcement learning for combinatorial optimization: A survey},
journal = {Computers \& Operations Research},
volume = {134},
pages = {105400},
year = {2021},
author = {Nina Mazyavkina and Sergey Sviridov and Sergei Ivanov and Evgeny Burnaev}
}

@inproceedings{Munos2005AVI,
author = {Munos, R\'{e}mi},
title = {Error bounds for approximate value iteration},
year = {2005},
publisher = {AAAI Press},
booktitle = {Proceedings of the 20th National Conference on Artificial Intelligence},
volumn = {2},
pages = {1006--1011},
numpages = {6},
location = {Pittsburgh, Pennsylvania},
series = {AAAI'05}
}

@article{Barrett2020CO, 
    title={Exploratory Combinatorial Optimization with Reinforcement Learning}, 
    volume={34}, 
    number={4}, 
    journal={Proceedings of the AAAI Conference on Artificial Intelligence}, 
    author={Barrett, Thomas and Clements, William and Foerster, Jakob and Lvovsky, Alex}, 
    year={2020}, 
    month={Apr.}, 
    pages={3243--3250}
}

@InProceedings{Riedmiller2005FittedQ,
    author="Riedmiller, Martin",
    editor="Gama, Jo{\~a}o and Camacho, Rui and Brazdil, Pavel B. and Jorge, Al{\'i}pio M{\'a}rio
    and Torgo, Lu{\'i}s",
    title="Neural Fitted Q Iteration -- First Experiences with a Data Efficient Neural Reinforcement Learning Method",
    booktitle="Proceedings of the 16th European Conference on Machine Learning",
    year="2005",
    location = {Porto, Portugal},
    series = {ECML'05},
    publisher="Springer Berlin Heidelberg",
    address="Berlin, Heidelberg",
    pages="317--328",
    numpages="12"
}

@InProceedings{Fan2020DQL,
  title = 	 {A Theoretical Analysis of Deep Q-Learning},
  author =       {Fan, Jianqing and Wang, Zhaoran and Xie, Yuchen and Yang, Zhuoran},
  booktitle = 	 {Proceedings of the 2nd Conference on Learning for Dynamics and Control},
  pages = 	 {486--489},
  year = 	 {2020},
  editor = 	 {Bayen, Alexandre M. and Jadbabaie, Ali and Pappas, George and Parrilo, Pablo A. and Recht, Benjamin and Tomlin, Claire and Zeilinger, Melanie},
  volume = 	 {120},
  series = 	 {Proceedings of Machine Learning Research},
  month = 	 {10-11 Jun},
  publisher =    {PMLR}
}

@article{kim2014guide,
  title={A guide to sample average approximation},
  author={Kim, Sujin and Pasupathy, Raghu and Henderson, Shane G.},
  journal={Handbook of simulation optimization},
  pages={207--243},
  year={2014},
  publisher={Springer}
}

@book{shapiro2021lectures,
  title={Lectures on stochastic programming: Modeling and theory},
  author={Shapiro, Alexander and Dentcheva, Darinka and Ruszczynski, Andrzej},
  year={2021},
  publisher={SIAM}
}

@book{Beck2023,
    author = {Beck, Amir},
    title = {Introduction to Nonlinear Optimization: Theory, Algorithms, and Applications with Python and MATLAB},
    publisher = {Society for Industrial and Applied Mathematics},
    year = {2014},
    edition  = {1st}
}

@incollection{Vershynin2012,
    place={Cambridge}, 
    title={Introduction to the non-asymptotic analysis of random matrices}, 
    booktitle={Compressed Sensing: Theory and Applications}, 
    publisher={Cambridge University Press}, author={Vershynin, Roman}, 
    editor={Eldar, Yonina C. and Kutyniok, Gitta}, 
    year={2012}, 
    pages={210--268}
}

@book{Puterman1994MDP,
    author       = {Martin L. Puterman},
    title        = {Markov Decision Processes: Discrete Stochastic Dynamic Programming},
    series       = {Wiley Series in Probability and Statistics},
    publisher    = {Wiley-Interscience},
    year         = {1994},
    address      = {New York}
}

@article{Stewart1989Projection,
title = {On scaled projections and pseudoinverses},
journal = {Linear Algebra and its Applications},
volume = {112},
pages = {189--193},
year = {1989},
author = {G. W. Pete Stewart},
}

@article{Munos2008FVI,
  author  = {R{{\'e}}mi Munos and Csaba Szepesv{{\'a}}ri},
  title   = {Finite-Time Bounds for Fitted Value Iteration},
  journal = {Journal of Machine Learning Research},
  year    = {2008},
  volume  = {9},
  number  = {27},
  pages   = {815--857}
}

@book{paschos2014applications,
  title={Applications of combinatorial optimization},
  author={Paschos, Vangelis Th.},
  volume={3},
  year={2014},
  publisher={John Wiley \& Sons}
}

@book{williamson2011design,
  title={The design of approximation algorithms},
  author={Williamson, David P. and Shmoys, David B.},
  year={2011},
  publisher={Cambridge university press},
  address={Cambridge}
}

@article{colorni1996heuristics,
  title={Heuristics from nature for hard combinatorial optimization problems},
  author={Colorni, Alberto and Dorigo, Marco and Maffioli, Francesco and Maniezzo, Vittorio and Righini, Giovanni and Trubian, Marco},
  journal={International Transactions in Operational Research},
  volume={3},
  number={1},
  pages={1--21},
  year={1996},
  publisher={Elsevier}
}

@article{hertz2003guidelines,
  title={Guidelines for the use of meta-heuristics in combinatorial optimization},
  author={Hertz, Alain and Widmer, Marino},
  journal={European Journal of Operational Research},
  volume={151},
  number={2},
  pages={247--252},
  year={2003},
  publisher={Elsevier}
}

@book{conway1985functional,
  author       = {John B. Conway},
  title        = {A Course in Functional Analysis},
  year         = {1985},
  publisher    = {Springer-Verlag},
  address      = {New York},
  series       = {Graduate Texts in Mathematics},
  volume       = {96},
  edition      = {1st}
}

@article{chung_neural_2025,
	title = {Neural combinatorial optimization with reinforcement learning in industrial engineering: A survey},
	volume = {58},
	pages = {130},
	number = {5},
	journal = {Artificial Intelligence Review},
	shortjournal = {Artif Intell Rev},
	author = {Chung, Kwok Tung and Lee, Carman and Tsang, Yung Po},
	year = {2025}
}

@article{hopfield_neural_1985,
  author    = {John J. Hopfield and David W. Tank},
  title     = {``\textnormal{Neural}'' computation of decisions in optimization problems},
  journal   = {Biol. Cybern.},
  year      = {1985},
  volume    = {52},
  number    = {3},
  pages     = {141--152},
  month     = jul
}

@inproceedings{Vaswani2017Attention,
author = {Vaswani, Ashish and Shazeer, Noam and Parmar, Niki and Uszkoreit, Jakob and Jones, Llion and Gomez, Aidan N. and Kaiser, \L{}ukasz and Polosukhin, Illia},
title = {Attention is all you need},
year = {2017},
publisher = {Curran Associates Inc.},
address = {Red Hook, NY, USA},
booktitle = {Proceedings of the 31st International Conference on Neural Information Processing Systems},
pages = {6000--6010},
numpages = {11},
location = {Long Beach, California, USA},
series = {NIPS'17}
}

@article{Farias2000FixedPoints,
  author  = {De Farias, Daniela Pucci and Van Roy, Benjamin},
  title   = {On the Existence of Fixed Points for Approximate Value Iteration and Temporal-Difference Learning},
  journal = {Journal of Optimization Theory and Applications},
  volume  = {105},
  pages   = {589--608},
  year    = {2000}
}

%%%%%%%%%%%%%%%%%%%%%%%%%%%%%%%%%%%%
%%%%%%%%%%%%%%%%%%%%%%%%%%%%%%%%%%%%

\clearpage
\appendix

%%%%%%%%%%%%%%%%%%%%%%%%%%%%%%%%%%%%
%%%%%%%%%%%%%%%%%%%%%%%%%%%%%%%%%%%%

\section{Proofs}

%%%%%%%%%%%%%%%%%%%%%%%%%%%%%%%%%%%%
%%%%%%%%%%%%%%%%%%%%%%%%%%%%%%%%%%%%

\subsection{Proofs of Section \ref{sec:co_as_mdp}}
\label{app:proofs_sec3}

\contraction*

\begin{proof}
    Let $V_1, V_2 \in \cV_{\cS}^0$. Note that $BV_1(s) = BV_2(s)$ for $s \in S_{d} \sqcup \Set{s_\infty}$. Consider $s \in \cS \setminus S_{d} \sqcup \Set{s_\infty}$. Hence, there exists $l=0,\ldots,d-1$ s.t. $s \in \cS_l$. Let $a^*_s \in A$ be the argmax for $BV_1$ at $s$ \eqref{eq:B}.
\begin{align*}
    BV_1(s) - BV_2(s) & = B^{a^*_{s}}V_1(s) - \max_{a \in \cA} \left\{ B^aV_2(s) \right\} \\
    & \leq B^{a_{s}^*}V_1(s) - B^{a^*_{s}}V_2(s) \\
    & = \sum_{s' \in \cS} p(s'|s,a^*_{s}) \left( V_1(s')-V_2(s')\right) \\
    & = \sum_{s' \in \cS} \tau(s' )p(s'|s,a^*_{s}) \left( \frac{V_1(s')-V_2(s')}{\tau(s')} \right) \\
    & \leq \tnorm{V_1-V_2} \sum_{s' \in \cS} \tau(s') p(s'|s,a^*_{s}) \\
    & = \tau_l \tnorm{V_1-V_2} \left[ \frac{\tau_{l+1}}{\tau_{l}} \sum_{s' \in \cS_{l+1}} p(s'|s,a^*_{s}) \right. + \\
    & \hspace{1.5cm} \left. \ldots + \frac{\tau_{d}}{\tau_{l}} \sum_{s' \in S_{d}} p(s'|s,a^*_{s}) + \frac{\tau_{d + 1}}{\tau_{l}} p(s_\infty|s,a^*_{s}) \right] \\
    & \leq \tau(s)\, \modulus^\tau \tnorm{V_1-V_2} \sum_{s' \in \cS} p(s'|s,a^*_{s}) \\
    & = \tau(s)\, \modulus^\tau \tnorm{V_1-V_2}
\end{align*}
We conclude that 
\begin{align*}
    BV_1(s) - BV_2(s) & \leq \tau(s)\, \modulus^\tau \tnorm{V_1-V_2}.
\end{align*}
The argument we employed is symmetric in $V_1$ and $V_2$, hence
\begin{align*}
    \abs{BV_1(s) - BV_2(s)} & \leq \tau(s)\, \modulus^\tau \tnorm{V_1-V_2}.
\end{align*}
Dividing by $\tau(s)$ and taking the maximum of the LHS, we get 
\begin{align*}
    \tnorm{BV_1 - BV_2} \leq \modulus^\tau \tnorm{V_1-V_2}.
\end{align*}
and the proof is complete. 
\end{proof}

To show that \eqref{eq:iota_def} is well-defined, first, consider $s=[x]$ and $a \in \cA$ such that $\lambda(s,a) = s_\infty$. By definition, 
\begin{align}
\iota(s,a;c) & = \tilde{g}(xa;c)-\tilde{g}(x;c) %\nonumber \\& 
=\tilde{g}(x;c)-\tilde{g}(x;c) %\nonumber \\& 
= 0 \label{eq:zero_incremental_reward}
\end{align}
The second equality results from the fact that $x$ and $xa$ share the same maximal sub-string $y$ that extends to a string in $\Pi$. For any other $s=[x]$, let $y \in [x]$ and $u' \in \cA^*$ s.t. $xau', yau' \in \Pi$ and denote $u = au'$. Then, by \eqref{eq:g_def}, $\tilde{g}(ya;c)-\tilde{g}(xa;c) = g(yau';c)-g(xau';c) = g(yu;c)-g(xu;c) = \tilde{g}(y;c)-\tilde{g}(x;c)$, which proves that \eqref{eq:iota_def} does not depend on the choice of $x$ s.t. $[x]=s$. Finally, the full reward structure $h: \R \times \cS \times \cA \times \cC \rightarrow \R$ is constructed additively via $h(\xi,s,a;c) \triangleq \xi + \iota(s,a;c)$.
The change \eqref{eq:reward_modified} in incremental reward modifies the additive reward structure of the $\cC$-parametrized SDP, coming from Theorem~\ref{thm:tseng_conditions}, $h(\xi,s,a;c) \triangleq \xi + r(s,a;c)$.
It extends recursively to $h: \R \times \cS \times \cA^* \times \cC \rightarrow \R$ via
\begin{align}
    & h(\xi,s,e;c) = \xi; \label{eq:h_e} \\
    & h(\xi,s,xa;c) = h(h(\xi,s,x;c),\lambda(s,x),a;c). \label{eq:h_recursive}
\end{align}

\tsengConditions*

\begin{proof}
We follow the construction in \S\ref{sec:KHT} to establish the additive $\cC$-parametrized SDP $\left( \cA, \cS, s_e, \cF, \lambda, h \right)$. By Assumption~\ref{assum:s_infinity}, the set of states $\cS$ has a distinguished class $s_\infty$ which is not refined. If $x \in s_\infty$ then there does {\it not} exist $w \in \cA^*$ such that $xw \in \Pi$, in particular, $\lambda(s_\infty,a)=s_\infty$.
By \eqref{eq:zero_incremental_reward}, we also have 
\begin{align}
\label{eq:s_inf_zero_reward}
    \iota(s_\infty,a;c)=0, \quad \forall a \in \cA   
\end{align}
This establishes Assumption~\ref{assum:absorbing}, with the deterministic transition probability $p$ determined via $\lambda$ and $\iota$ standing for $r$. In particular, $p(s;|s,a) = \delta_{s',\lambda(s,a)}$.
Note that we have not relied, so far, on $\abs{\Pi} < \infty$. 

As for Assumption~\ref{assum:proper}, consider the equivalence relation $\sim_{d_\Pi}$ defined via the following partition of $\cA^*$ by non-intersecting subsets
\begin{align}
    \cA^* & = s_\infty \sqcup \bigsqcup_{l=0}^{d_\Pi} \cA_l, \label{eq:A_filtration}\\
    \cA_l & \triangleq \Set{x \notin s_\infty|\lng(x) = l}, \nonumber
\end{align}
with $d_\Pi$ as in Definition~\ref{def:dpi}. We will show that $\sim_\Pi$ refines $\sim_{d_\Pi}$, and, since $\sim$ refines $\sim_\Pi$ by condition ({\romannumeral 1}) of Theorem~\ref{thm:KH}, it also refines $\sim_{d_\Pi}$. With $\sim$ refining $\sim_{d_\Pi}$, we will get a partition of $\cS$ via $\cS\setminus\{s_\infty\} = \bigsqcup_{l=0}^{d_\Pi} \cS_l$ for $\cS_l \triangleq \Set{s \in \cS\setminus\{s_\infty\} | s \subseteq \cA_l}$,
which constitutes a special case of \eqref{eq:partition} with $d\equiv d_\Pi$. For any $l=0,\ldots,d_\Pi$, $s \in \cS_l$, and $a \in \cA$, there will exist $s' \in  \cS_{l+1} \sqcup \{s_\infty\}$ such that $\lambda(s,a)=s'$ -- string length necessarily increases by concatenation. Such a partition of $\cS$ will thus establish Assumption~\ref{assum:proper}, and, specifically, $(p^\pi)^t(s_\infty|s)=1$ for $t \geq d_\pi+1$ for any $\pi : \cS \rightarrow \cA$.

To show that $\sim_\Pi$ refines $\sim_{d_\Pi}$, consider a pair $x \sim_\Pi y$. By Definition~\ref{def:simPi}, exactly one of the following two options holds: ({\romannumeral 1}) neither $x$ nor $y$ extend to a string in $\Pi$, or, ({\romannumeral 2}) both $x$ and $y$ extend to a string in $\Pi$. In scenario ({\romannumeral 1}), we have $x, y \in s_\infty$, which, by \eqref{eq:A_filtration}, is also an equivalence class of $\sim_{d_\Pi}$, therefore, to demonstrate the refinement of $\sim_{d_\Pi}$ by $\sim_\Pi$, we need only consider scenario ({\romannumeral 2}).  Thus, assuming scenario ({\romannumeral 2}) for $x \sim_\Pi y$ (which includes the case when both $x$ and $y$ are in $\Pi$), let $d(x) \triangleq \max_{u \in \cA^*: xu \in \Pi} \lng(u)$.
By Definition~\ref{def:simPi}, $d(x)=d(y)$, since the sets we are maximizing over are identical. Also, note that $d(x) \leq d_\Pi-\lng(x)$ by definition of $d_\Pi$ (and similarly for $d(y)$). 
Applying Assumption~\ref{assum:x_extension}, we can show that $d(x) \geq d_\Pi-\lng(x)$ and, similarly, that $d(y) \geq d_\Pi-\lng(y)$, for there exists $u \in \cA^*$ such that $xu \in \Pi$ and $\lng(xu)=d_\Pi$ implying that $d(x) \geq \lng(u) = d_\Pi - \lng(x)$. This establishes $\lng(x)=\lng(y)$ and, by definition \eqref{eq:A_filtration}, the refinement of $\sim_{d_\Pi}$ by $\sim_\Pi$.
\end{proof}

\begin{lemma}
\label{lem:recursive_additivity}
$h(\xi,s,ax;c) = \xi + r(s,a;c) + h(0,\lambda(s,a),x;c)$.
\end{lemma}

\begin{proof}
    We prove by induction on $\lng(x)$. The base case for $x=e$ is derived from both \eqref{eq:h_e} and \eqref{eq:h_recursive}. For induction step, consider $a \in \cA$ and $x'=xa' \in \cA^*$ with $s'=\lambda(s,a)$. Then,
    \begin{align*}
        h(\xi,s,ax';c) &= h(\xi,s,axa';c) \\
        &= h(h(\xi,s,ax;c),\lambda(s,ax),a';c) \\
        &= h(\xi,s,ax;c) + r(\lambda(s,ax),a';c) \\
        &= \xi + r(s,a;c) + h(0,\lambda(s,a),x;c) + r(\lambda(s,ax),a';c) \\
        &= \xi + r(s,a;c) + h(h(0,\lambda(s,a),x;c),\lambda(s,ax),a';c) \\
        &= \xi + r(s,a;c) + h(h(0,\lambda(s,a),x;c),\lambda(\lambda(s,a),x),a';c) \\
        &= \xi + r(s,a;c) + h(h(0,\lambda(s,a),x;c),\lambda(\lambda(s,a),x),a';c) \\
        &= \xi + r(s,a;c) + h(h(0,s',x;c),\lambda(s',x),a';c) \\
        &= \xi + r(s,a;c) + h(0,s',xa';c) \\
        &= \xi + r(s,a;c) + h(0,\lambda(s,a),x';c),
    \end{align*}
    where we used the induction step on the forth equality and a combination of \eqref{eq:h_recursive} and additivity for the rest. This completes the proof.
\end{proof}

\VCOproblem*

\begin{proof}
First, following \eqref{eq:optimal_value_function_general}, let $x \in \cA^*$ such that $\lambda(s_e,x)=s_\infty$. This means  $[x]=s_\infty$. Denote $\lng(x) = l \geq 1$ and let $x=x_1 \cdots x_l$, $x_j \in \cA$. Also denote $\tx_j = x_1 \cdots x_j \in \cA^*$ for $j= 1 \ldots l$ and $\tx_0=e$. Let $\tx_k$ be the longest sub-string of $x$ such that $\tx_k \notin s_\infty$, in particular, $\lambda(\tx_k,x_{k+1})=s_\infty$. We necessarily have $k < l$. Also note that $h(0,s_e,x;c) = h(0,s_e,\tx_{k+1};c)$. 

We may as well assume that $[\tx_k] \in \cF$, otherwise, by \eqref{eq:reward_modified} and \eqref{eq:h_recursive}, we have
\begin{align*}
    h(0,s_e,x;c) & = h(0,s_e,\tx_{k+1};c) \\
    & = h(h(0,s_e,\tx_k;c),[\tx_k],x_{k+1};c) \\ 
    & = h(0,s_e,\tx_k;c) + r([\tx_k],x_{k+1};c) \\
    & = h(0,s_e,\tx_k;c) - M \\
    & \approx - M
\end{align*}
and $x$ would not compete for the maximum  \eqref{eq:optimal_value_function_general}. Thus, we have
\begin{align*}
    h(0,s_e,x;c) & = h(0,s_e,\tx_{k+1};c) \\
    & = \sum_{j=0}^{k} r([\tx_j],x_{j+1};c) \\
    & = \sum_{j=0}^{k-1} \iota([\tx_j],x_{j+1};c) \\
    & = \sum_{j=0}^{k-1} (\tg(x_{j+1};c)-\tg(x_j;c)) \\
    & = \tg(\tx_k;c)-\tg(e;c) \\
    & = \tg(\tx_k;c)
\end{align*}
utilizing recursive additivity from Lemma~\ref{lem:recursive_additivity} in the 2nd equation, the definition \eqref{eq:reward_modified} as well as \eqref{eq:zero_incremental_reward} in the 3rd equality, the definition \eqref{eq:iota_def} in the 4th equation, and \eqref{eq:gec_zero} in the last equality. But if $[\tx_k] \in \cF$ then $\tx_k \in \Pi$ and $\tg(\tx_k;c)=g(\tx_k;c)$ \eqref{eq:g_tilde_equals_g}. Taking the maximum over all $x \in \cA^*$ such that $\lambda(s_e,x)=s_\infty$, we establish that $V^*(s_e;c) \leq \max_{x \in \Pi} g(x;c)$.

To establish the reverse inequality, let $x \in \Pi$. By Assumption~\ref{assum:x_extension}, there exists $u \in \cA^*$ such that $[xu] \in \cF$ and $\lambda([xu],a)=s_\infty$ for any $a \in \cA$.
Pick an $a$ and let $y=xua$, thus, $\lambda(s_e,y)=s_\infty$. Following the calculation above, with $xu$ being the longest sub-string of $y$ such that $xu \notin s_\infty$, we have $h(0,s_e,y;c) = \tg(xu;c) = g(xu;c)$, with the last equation established by \eqref{eq:g_tilde_equals_g}. Then, $V^*(s_e;c) \geq h(0,s_e,y;c) = g(xu;c) \geq g(x;c)$, 
with the right-most inequality given by Assumption~\ref{assum:x_extension}. Taking the maximum over all $x \in \Pi$, the reverse inequality is established and completes the proof.
\end{proof}

\Vfunctional*
%\begin{app_prop*}[\ref{prop:V^*_functional_equations}]
%    Under the assumptions of Theorem~\ref{thm:tseng_conditions}, the following functional equations hold for $V^*$ \eqref{eq:optimal_value_function_general}:
 %   \begin{align*}
 %       V^*(s;c) = \left\{ \begin{array}{ll}
%0 &  s=s_\infty \\
%            \max_{a \in \cA} \{r(s,a;c) + V^*(\lambda(s,a);c)\} & \mbox{otherwise}
 %       \end{array} \right.
 %   \end{align*}    
%\end{app_prop*}

\begin{proof}
Define a function 
\begin{align*}
    H(s;c) \triangleq \left\{ \begin{array}{ll}
        0 &  s=s_\infty \\
        \max_{a \in \cA} \{r(s,a;c) + V^*(\lambda(s,a);c)\} & \mbox{otherwise}
    \end{array} \right.
\end{align*}
First, we'll show that $V^*(s;c) \leq H(s;c)$. By definition \ref{eq:optimal_value_function_general}, $V^*(s;c) = h(0,s,x;c)$ for some $x \in \cA^*$ such that $\lambda(s,x)=s_\infty$. If $x=e$ then $s=s_\infty$ and $V^*(s;c) = 0 = H(s;c)$. If $x \neq e$, then $x=ay$ for some $a \in \cA$ and 
\begin{align*}
    V^*(s;c) & = h(0,s,ay;c) \\
    & = r(s,a;c) + h(0,\lambda(s,a),y;c) \\
    & \leq r(s,a;c) + V^*(\lambda(s,a);c) \\
    & \leq H(s;c)
\end{align*}
where the 2nd equality comes from recursive additivity in Lemma~\ref{lem:recursive_additivity}. Let's now show the reverse $V^*(s;c) \geq H(s;c)$. It holds for $s=s_\infty$. If $s \neq s_\infty$ then let's pick an argmax $a^*$ such that $H(s;c) = r(s,a^*;c) + V^*(\lambda(s,a^*);c)$. By definition, $V^*(\lambda(s,a^*);c) = h(0,\lambda(s,a^*),y;c)$ for some $y \in \cA^*$ such that $\lambda(\lambda(s,a),y)=s_\infty$. We can then show
\begin{align*}
    H(s;c) & = r(s,a^*;c) + h(0,\lambda(s,a^*),y;c) \\
    & = h(0,s,a^*y;c) \\
    & \leq V^*(s;c)
\end{align*}
by recursive additivity in Lemma~\ref{lem:recursive_additivity} and the definition \ref{eq:optimal_value_function_general}. This ends the proof.
\end{proof}

\Approximation*
%\begin{app_thm*}[\ref{thm:approximation}]
%    Let ${V}$ be an $\epsilon$-approximation of $V^*$ w.r.t. a $\tau$-weighted $\ell_\infty$ norm \eqref{eq:tau_norm}. Let $\{(s_i,a_{i+1})\}_{i=0}^{k}$ be as defined \eqref{eq:approx_solution}. Under the assumptions of Theorem~\ref{thm:tseng_conditions} and Assumption~\ref{assum:partial_to_feasible}, for a sufficiently small $\epsilon > 0$, the following holds:
 %   \begin{enumerate}[label=(\roman*)]
%        \item $y=a_1 \cdots a_{k} \in \cA^*$ is a feasible solution, namely, $y \in \Pi$;
%        \item $\max_{x \in \Pi} g(x;c) - g(y;c) \leq 2 \epsilon \tau_0 (d_\Pi + 1)$.
%    \end{enumerate}
%\end{app_thm*}

\begin{proof}
    Beginning with ({\romannumeral 1}), consider $s_k$. By construction, $s_k \neq s_\infty$ and $\lambda(s_k,a_{k+1}) = s_\infty$. Either $s_k \in \cF$ or not. If $s_k \notin \cF$, then $r(s_k,a_{k+1};c) = -M$. By Assumption~\ref{assum:partial_to_feasible}, there exists $a' \in \cA$ s.t. $s' = \lambda(s_k,a') \neq s_\infty$. By definition of $a_{k+1}$,
    \begin{align*}
        r(s_k,a_{k+1};c) + V(\lambda(s_k,a_{k+1})) & \geq r(s_k,a';c) + V(\lambda(s_k,a')),
    \end{align*}
    which implies $- M \geq r(s_k,a';c) + V(s')$.
    Since $V$ is an $\epsilon$-approximation of $V^*$ w.r.t. the $\tau$-weighted $\ell_\infty$ norm, $V(s') > V^*(s') - \epsilon \tau(s')$. Hence, $\epsilon \tau(s') > M + r(s_k,a';c) + V^*(s')$,
    which is contradictory for a sufficiently small $\epsilon$. Therefore, $s_k \in \cF$. Now, note that $s_k = [y]$, and since $\sim$ refines $\sim_\Pi$, this means that $y \in \Pi$.

    Focusing now on ({\romannumeral 2}), for each $s_i$ apply Proposition~\ref{prop:V^*_functional_equations} to pick $a_i^*$ such that
    \begin{align*}
        V^*(s_i;c) = r(s_i,a_i^*;c) + V^*(\lambda(s_i,a_i^*))
    \end{align*}
    Then, by $\epsilon$-approximation w.r.t. the $\tau$-weighted $\ell_\infty$ norm, for each $i=0,\ldots,k$,
    \begingroup
    \allowdisplaybreaks
    \begin{align*}
        r(s_i,a_{i+1};c) + V^*(s_{i+1};c) + \epsilon \tau(s_{i+1}) & \geq r(s_i,a_{i+1};c) + V(s_{i+1}) \\
        & \geq r(s_i,a_{i+1}^*;c) + V(\lambda(s_i,a_{i+1}^*)) \\
        & \geq r(s_i,a_{i+1}^*;c) + V^*(\lambda(s_i,a_{i+1}^*);c) - \epsilon \tau(\lambda(s_i,a_{i+1}^*)) \\
        & = V^*(s_i;c) - \epsilon \tau(\lambda(s_i,a_{i+1}^*)),
    \end{align*}
    \endgroup
    where the 2nd inequality results from \eqref{eq:approx_solution}. This implies, by \eqref{eq:taus}, that $2 \epsilon \tau_0 \geq V^*(s_i;c) - V^*(s_{i+1};c) - r(s_i,a_{i+1};c)$.
    Taking the sum, and recalling that $k\leq d_{\Pi}$ must hold, we get
    \begin{align*}
        2 \epsilon \tau_0 (d_\Pi + 1) & \geq \sum_{i=0}^k \left( V^*(s_i;c) - V^*(s_{i+1};c) - r(s_i,a_{i+1};c) \right) \\
        & = V^*(s_e;c) - \sum_{i=0}^k r(s_i,a_{i+1};c),
    \end{align*}
    which completes the proof by Proposition~\ref{prop:V^*_co_problem} and the additivity of the SDP.
\end{proof}

%%%%%%%%%%%%%%%%%%%%%%%%%%%%%%%%%%%%
%%%%%%%%%%%%%%%%%%%%%%%%%%%%%%%%%%%%

\subsection{Proofs of Section \ref{sec:rl}}
\label{app:proofs_sec4}

By Proposition~\ref{prop:contraction} and the Banach fixed-point theorem, the undiscounted Bellman map $B: \cV_{\cS}^0 \rightarrow \cV_{\cS}^0$ has a unique fixed point, $BV^* = V^*$, which is the optimal value function in this case. Let $\pi^*$ denote its corresponding optimal policy \eqref{eq:optimal_policy}. Let $P_*$ be its transition probability matrix and let $r_*$ be its immediate reward vector, namely,
\begin{align}
    & (P_*)_{s,s'} \triangleq p(s'|s,\pi_*(s)) \qquad & s, s' \in \cS \label{eq:optimal_policy_reward} \\
    & (r_*)_s \triangleq r(s,\pi_*(s)) \qquad &  s \in \cS \nonumber
\end{align}
Note that $(r_*)_s=0$ for $s = s_\infty$.
By the MDP transition structure, the transition probability matrix for $\pi_*$ is a upper block-triangular matrix of the form
\newlength{\cdotswidth}
\settowidth{\cdotswidth}{$\cdots$}
\begin{align}
\label{eq:optimal_transition_matrix}
  {P}_* = \begin{blockarray}{cccccc}
    {\scriptstyle \cS_0} & {\scriptstyle \cS_{1}} & \cdots & {\scriptstyle \cS_{d}} & {\scriptstyle s_\infty} \\
    \begin{block}{(c|c|c|c|c)c}
      \makebox[\cdotswidth]{0} & \left(P_*\right)_{0,1} & \cdots &  \left(P_*\right)_{0,d} & \left(P_*\right)_{0,\infty} & {\scriptstyle \cS_0} \vphantom{\vdots} \\ \cmidrule{1-5}
      0 & 0 & \ddots & \vdots & \vdots & {\scriptstyle \vdots} \\ \cmidrule{1-5}
      0 & 0 & 0 & \left(P_*\right)_{d-1,d} & \left(P_*\right)_{d-1,\infty} & {\scriptstyle \cS_{d -1}} \vphantom{\vdots} \\ \cmidrule{1-5}
      0 & 0 & 0 & 0 & \left(P_*\right)_{d,\infty} & {\scriptstyle \cS_{d}} \vphantom{\vdots} \\ \cmidrule{1-5}
      0 & 0 & 0 & 0 & 1 & {\scriptstyle s_\infty} \vphantom{\vdots} \\
    \end{block} 
  \end{blockarray}
\end{align}

\begin{prop}
\label{prop:V*_equation}
${V}^* = r_* + P_*r_* + \ldots + P_*^{d}r_*$.
\end{prop}

\begin{proof} 
Since $V^*$ is a fixed point for $B$, we have
\begin{align*}
    {V}^*(s) & = \max_{a \in \cA} \left\{ r(s,a) + \sum_{s' \in \cS} p(s'|s,a) V^*(s') \right\} \\
    & = r(s,\pi^*(s)) + \sum_{s' \in \cS} p(s'|s,\pi^*(s)) V^*(s'),
\end{align*}
namely,
\begin{align}
\label{eq:optimal_transition_fixed_point}
    {V}^* = r_* + P_*{V}^*.
\end{align}
Let $\tilde{P}_*$ be $P_*$ \eqref{eq:optimal_transition_matrix} with the bottom row and right column associated with $s_\infty$ removed. For $k=1,\ldots,d$, $\tilde{P}_*^k$ is an upper block-triangular matrix with zeros along the main diagonal and (possibly) non-zero blocks $\left(\tilde{P}_*^k\right)_{l,l+k'}$ of size $\abs{\cS_l} \times \abs{\cS_{l+k'}}$ above it, encoding the $k$-step transition probabilities from $\cS_l$ to $S_{l+k'}$ for $l=0,\ldots,d-k$ and $k'=0,\ldots,d-l$. Moreover, $\tilde{P}_*^k = 0$ for all $k > d$.

For any $x \in \R^{\abs{\cS}}$, let $x_{\cS \setminus \Set{s_\infty}} \in \R^{\abs{\cS}-1}$ denote $x$ with the entry associated with  $s_\infty$ removed. If $x_{s_\infty} = 0$, then
\begin{align}
\label{eq:tildePtoP}
    {P}_*^k x = \left[ \begin{array}{c}
        \tilde{P}_*^k x_{\cS \setminus \Set{s_\infty}} \\ \cmidrule{1-1} 0
    \end{array}
    \right] \qquad \forall k \in \N,
\end{align}
Following \eqref{eq:optimal_transition_fixed_point}, we thus have
\begin{align*}
    (I - \tilde{P}_*){V}^*_{\cS \setminus \Set{s_\infty}} & = \left(r_*\right)_{\cS \setminus \Set{s_\infty}},
\end{align*}
where $I\equiv I_{\abs{\cS}-1}$ is the identity matrix in dimension $\abs{\cS}-1$. Using the identity $(I+ \ldots + \tilde{P}_*^k)(I-\tilde{P}_*) = I-\tilde{P}_*^{k+1}$, we get
\begin{align*}
    {V}^*_{\cS \setminus \Set{s_\infty}} & = (I+ \ldots + \tilde{P}_*^{d}) \left(r_*\right)_{\cS \setminus \Set{s_\infty}},
\end{align*}
taking $k=d$ and applying $\tilde{P}_*^{d+1} = 0$, which extends, by \eqref{eq:tildePtoP}, to ${V}^* = (I+ \ldots + {P}_*^{d}) r_*$, completing the proof.
\end{proof}

The undiscounted ($\delta=1$) Bellman map \eqref{eq:B} applied to $V \in \cV_\cS^0$ \eqref{eq:VS0} now becomes
\begin{align*}
    BV(s;c) = \max_{a \in \cA} \left\{ r(s,a;c) + V(\lambda(s,a);c) \right\}, \quad s \in \cS.
\end{align*}
By Theorem~\ref{thm:tseng_conditions}, Assumptions~\ref{assum:absorbing} and \ref{assum:proper} hold for the resulting $\cC$-parametrized family of MDPs. By \cite{Tseng1990Convergence}, Assumptions~\ref{assum:absorbing} and \ref{assum:proper} guarantee the point-wise convergence of $\{B^tV_0(\cdot;c)\}_{t=0}^\infty$ to a unique fixed point for each $c \in \cC$ and  initial $V_0(\cdot;c) \in \cV_\cS^0$ (see \S\ref{sec:undiscounted}). That fixed point is $V^*(\cdot;c)$ by Proposition~\ref{prop:V^*_functional_equations}. The transition probability matrix \eqref{eq:optimal_transition_matrix} associated with its derived optimal policy $\pi_*$ \eqref{eq:optimal_policy} takes the simpler form
\begin{align}
\label{eq:co_mdp_transition_matrix}
  {P}_* = \begin{blockarray}{ccccccc}
    {\scriptstyle \cS_0} & {\scriptstyle \cS_{1}} & {\scriptstyle \cS_{2}} & \cdots & {\scriptstyle \cS_{d}} & {\scriptstyle s_\infty} \\
    \begin{block}{(c|c|c|c|c|c)c}
      \makebox[\cdotswidth]{0} & \left(P_*\right)_{0,1} & 0 & \cdots & 0 & \left(P_*\right)_{0,\infty} & {\scriptstyle \cS_0} \vphantom{\vdots} \\ \cmidrule{1-7}
      0 & 0 & \left(P_*\right)_{1,2} & 0 & \vdots & \left(P_*\right)_{1,\infty} & {\scriptstyle \cS_1} \\\cmidrule{1-7}
      0 & 0 & \ddots & \ddots & 0 & \vdots & {\scriptstyle \vdots} \\ \cmidrule{1-7}
      0 & 0 & \cdots & 0 & \left(P_*\right)_{d-1,d} & \left(P_*\right)_{d-1,\infty} & {\scriptstyle \cS_{d -1}} \vphantom{\vdots} \\ \cmidrule{1-7}
      0 & 0 & 0 & 0 & 0 & \left(P_*\right)_{d,\infty} & {\scriptstyle \cS_{d}} \vphantom{\vdots} \\ \cmidrule{1-7}
      0 & 0 & 0 & 0 & 0 & 1 & {\scriptstyle s_\infty} \vphantom{\vdots}\\
    \end{block} 
  \end{blockarray}
\end{align}

\Vbound*
%\begin{app_prop*}[\ref{prop:V^*_bound}]
%$\tnorm{V^*} \leq \frac{R \sqrt{W} \modulus^\tau}{1-\modulus^\tau} \left( \frac{1}{\tau_\infty} - \frac{1}{\tau_0}\right)$.
%\end{app_prop*}

\begin{proof} 
Let $\pi^*$ \eqref{eq:optimal_policy} denote the optimal policy w.r.t. $B: \cV_{\cS}^0 \rightarrow \cV_{\cS}^0$. Let $P_*$ \eqref{eq:co_mdp_transition_matrix} be its transition probability matrix, specific to the unique CO MDP structure, and $r_*$ \eqref{eq:optimal_policy_reward} its immediate reward vector. By the transition and reward structure of the CO MDP, as well as the optimality of $V^*$, we have ${P}_*$ as in \eqref{eq:optimal_transition_matrix}, only, for $l=0,\ldots,d_\Pi$, we have $\left(P_*\right)_{l,l'} \neq 0$ iff $l'-l=1$ or $l'=d_\Pi+1$.
Let $\tilde{P}_*$ be the $\left(\abs{\cS}-1\right) \times \left(\abs{\cS}-1\right)$ matrix identified with $P_*$ with its bottom row and right column, associated with $s_\infty$, removed. For $k=0,\ldots,d_\Pi$, $\tilde{P}_*^k$ is an upper block triangular matrix with zeros along the main diagonal and non-zero blocks $\left(\tilde{P}_*^k\right)_{l,l+k}$ of size $\abs{\cS_l} \times \abs{\cS_{l+k}}$ above it, encoding the $k$-step transition probabilities from $\cS_l$ to $\cS_{l+k}$ for $l=0,\ldots,d_\Pi-k$. Moreover, $\tilde{P}_*^k = 0$ for all $k > d_\Pi$.

For any $x \in \R^{\abs{\cS}}$, let $x_{\cS \setminus \Set{s_\infty}} \in \R^{\abs{\cS}-1}$ denote $x$ with the entry associated with  $s_\infty$ removed. Following Proposition~\ref{prop:V*_equation}, we have
\begin{align*}
    {V}^*_{\cS \setminus \{s_\infty\}} & = (I+ \ldots + \tilde{P}_*^{d}) \left(r_*\right)_{\cS \setminus \{s_\infty\}}.
\end{align*}
Consequently,
\begingroup
\allowdisplaybreaks
\begin{align*}
    \tnorm{V^*} & \leq \sum_{k=0}^{d_\Pi} \tnorm{\tilde{P}_*^{k} \left(r_*\right)_{\cS \setminus \Set{s_\infty}}} \\
    & = \sum_{k=0}^{d_\Pi} \max_{l = 0, \ldots, d_\Pi-k} \left\{ \frac{1}{\tau_l} \supnorm{\left(\tilde{P}_*^{k} \left(r_*\right)_{\cS \setminus \Set{s_\infty}} \right)_l} \right\} \\
    & = \sum_{k=0}^{d_\Pi} \max_{l = 0, \ldots, d_\Pi-k} \left\{ \frac{1}{\tau_l} \supnorm{ \left(\tilde{P}_*^{k}\right)_{l,l+k} \left( {r}_* \right)_{l+k}} \right\} \\
    & \leq \sum_{k=0}^{d_\Pi} \max_{l = 0, \ldots, d_\Pi-k} \left\{ \frac{1}{\tau_l} \twonorm{\left(\tilde{P}_*^{k}\right)_{l,l+k} \left( {r}_* \right)_{l+k}} \right\} \\
    & \leq \sum_{k=0}^{d_\Pi} \max_{l = 0, \ldots, d_\Pi-k} \left\{ \frac{1}{\tau_l} \twonorm{\left(\tilde{P}_*^{k}\right)_{l,l+k} } \twonorm{\left( {r}_* \right)_{l+k} \vphantom{\tilde{P}_*^{k}}} \right\} \\
    & \leq R \sum_{k=0}^{d_\Pi} \max_{l = 0, \ldots, d_\Pi-k} \left\{ \frac{\sqrt{\abs{\cS_{l+k}}}}{\tau_l} \twonorm{\left(\tilde{P}_*^{k}\right)_{l,l+k} }  \right\} 
\end{align*}
\endgroup
By the Gershgorin circle theorem and the stochasticity of $P_*^k$,\footnote{By our convention for defining $P_*$, its $2$-norm is defined by row multiplication.} we have
\begin{align*}
    \twonorm{P_*^k} = \twonorm{P_* \vphantom{P_*^k}}^k = 1, \qquad \forall k \in \N.
\end{align*}
Since
\begin{align*}
    \twonorm{\left(\tilde{P}_*^k\right)_{l,l+k}} \leq \twonorm{\tilde{P}_*^k \vphantom{\left(\tilde{P}_*^k\right)_{l,l+k}}} \leq \twonorm{P_*^k \vphantom{\left(\tilde{P}_*^k\right)_{l,l+k}}},
\end{align*}
we can conclude that
\begin{align*}
    \tnorm{V^*} \leq R \sum_{k=0}^{d_\Pi} \max_{l = 0, \ldots, d_\Pi-k} \left\{ \frac{\sqrt{\abs{\cS_{l+k}}}}{\tau_l} \right\}.
\end{align*}
Working out each maximum, we have
\begin{align*}
    \max_{l = 0, \ldots, d_\Pi-k} \left\{ \frac{\sqrt{\abs{\cS_{l+k}}}}{\tau_l} \right\} \leq \frac{\sqrt{W}}{\tau_{d_\Pi-k}}, \qquad k=0,\ldots,d_\Pi.
\end{align*}
Thus,
\begin{align*}
    \tnorm{V^*} \leq R \sqrt{W} \sum_{l=0}^{d_\Pi} \frac{1}{\tau_l}.
\end{align*}
Evaluating the sum, 
\begin{align*}
    \Delta = \sum_{l=0}^{d_\Pi} \frac{1}{\tau_l} = \sum_{l=0}^{d_\Pi} \frac{1}{\tau_{l+1}} \cdot \frac{\tau_{l+1}}{\tau_l} \leq \modulus^\tau \sum_{l=0}^{d_\Pi} \frac{1}{\tau_{l+1}} = \modulus^\tau \left(\Delta + \left( \frac{1}{\tau_\infty} - \frac{1}{\tau_0}\right)\right),
\end{align*}
we get
\begin{align*}
    \Delta & \leq \frac{\modulus^\tau}{1-\modulus^\tau} \left( \frac{1}{\tau_\infty} - \frac{1}{\tau_0}\right),
\end{align*}
which is tight for a choice of $\tau_l = \tau_0\modulus^l$ for $\modulus \in (0,1)$. The result now follows.
\end{proof}

\begin{lemma}
\label{lem:sigma_l2_tau_inf}
    For $V \in \cV_{\cS}$ with $V|_{\cS\setminus \supp(\sigma)} \equiv 0$, we have
    \begin{align*}
        c(\sigma,\tau) \tnorm{V} \leq \snorm{V} \leq \tau_0 \tnorm{V},
    \end{align*}
    where
    \begin{align*}
        c(\sigma,\tau) \triangleq \min_{s \in \supp(\sigma)} \left\{ \tau(s) \sqrt{\sigma(s)} \right\}.
    \end{align*}
\end{lemma}

\begin{proof}
    We use the well-known relation between the supremum and $l_2$ norms $\norm{V}_\infty \leq \norm{V}_2$,
    to establish the inequality on the LHS
    \begin{align*}
        \tnorm{V}^2 & = \norm{ \diag(\tau)^{-1} V}_\infty^2 \\
        & \leq \norm{ \diag(\tau)^{-1} V}_2^2 \\
        & = \norm{ \diag(\sigma)^{1/2} \diag(\sigma)^{-1/2} \diag(\tau)^{-1} V}_2^2 \\
        & = \snorm{ \diag(\sigma)^{-1/2} \diag(\tau)^{-1} V}^2 \\
        & \leq \frac{1}{\left(\min_{s \in \supp(\sigma)} \left\{ \tau(s) \sqrt{\sigma(s)} \right\}\right)^2} \snorm{V}^2.
    \end{align*}
    On the other hand, 
    \begin{align*}
        \snorm{V}^2 & = \E_{s\sim\sigma}\left[ V(s)^2 \right] \\
        & \leq \tau_0^2 \E_{s\sim\sigma}\left[ \frac{V(s)^2}{\tau(s)^2} \right] \\
        & \leq \tau_0^2 \max_{s \in \cS} \left\{ \frac{V(s)^2}{\tau(s)^2} \right\} \\
        & = \tau_0^2 \left( \max_{s \in \cS} \left\{ \frac{V(s)}{\tau(s)} \right\} \right)^2 \\
        & = \tau_0^2 \tnorm{V}^2,
    \end{align*}
    which establishes the inequality on the RHS.
\end{proof}

\begin{lemma}
\label{lem:projection_stationarity_condition}
    Assume $\cV_\Theta \subset \cV^0_S$ is convex and let $\tilde{V} \in \cV^0_S$. Then $\Pi^\sigma_\Theta \tilde{V} \in \cV_\Theta$ is the optimal solution of \eqref{eq:projection_min} iff
    \begin{align*}
        \left(\tilde{V}-\Pi^\sigma_\Theta\tilde{V}\right)^\top \diag(\sigma) \left(V_\theta-\Pi^\sigma_\Theta\tilde{V} \vphantom{\tilde{V}}\right) \leq 0
    \end{align*}
    for all $\theta \in \Theta$.
\end{lemma}

\begin{proof}
    By \cite[Theorem 9.7]{Beck2023} applied to the continuously differentiable convex function $g(V) \triangleq \snorm{V - \tilde{V}}^2$
    defined over the compact and convex set $\cV_\Theta \subset \cV^0_S$, we have that $\Pi^\sigma_\Theta\tilde{V}$ is the optimal solution iff it is a stationary point, i.e.,
    \begin{align*}
        \nabla g\left(\Pi^\sigma_\Theta\tilde{V}\right)^\top \left(V_\theta-\Pi^\sigma_\Theta\tilde{V}\right) \geq 0
    \end{align*}
    for all $\theta \in \Theta$. Spelling out the stationarity condition, we get
    \begin{align*}
        \left(\tilde{V}-\Pi^\sigma_\Theta\tilde{V}\right)^\top \diag(\sigma) \left(V_\theta-\Pi^\sigma_\Theta\tilde{V} \vphantom{\tilde{V}}\right) \leq 0
    \end{align*}
    which completes the proof.
\end{proof}

\PVI*
%\begin{app_prop*}[\ref{prop:pvi}]
%Under Assumption~\ref{assum:contraction}, $\sigma$-PVI \eqref{eq:pvi} converges irrespective of initialization and its limit \eqref{eq:pvi_limit} satisfies
%\begin{align}
%    \tnorm{\tilde{V}^* - \Pi^\sigma_\Theta V^*} \leq \frac{\modulus}{1 - \modulus} \cdot \tnorm{V^* - \Pi^\sigma_\Theta V^* \vphantom{\lim_{t \rightarrow \infty} V_{{\theta}_t}}}.
%\end{align}
%In particular, the approximation scheme is fully expressive iff $\tilde{V}^* = V^*$.
%\end{app_prop*}

\begin{proof}
    Since, by Assumption~\ref{assum:contraction}, $\Pi^\sigma_\Theta B$ is a contraction of modulus $\modulus <1$ w.r.t. some $\tnorm{\cdot}$, then, by the Banach fixed point theorem, it has a unique fixed point $\tilde{V}^*=\lim_{t \rightarrow \infty}(\Pi^\sigma_\Theta B)^t V_{{\theta}_0}$ for any choice $V_{\theta_0} \in \cV_\Theta$. In particular, $\sigma$-PVI converges to $\tilde{V}^*$. As for the bound, for any $V_{{\theta}_0} \in \cV_\Theta$,
    \begin{align*}
    \tnorm{\Pi^\sigma_\Theta B V_{{\theta}_0} - \Pi^\sigma_\Theta V^*} & = \tnorm{\Pi^\sigma_\Theta B V_{{\theta}_0} - \Pi^\sigma_\Theta BV^*} \\
    & \leq \modulus \tnorm{V_{{\theta}_0} - V^*} \\
    & \leq \modulus \tnorm{V_{{\theta}_0} - \Pi^\sigma_\Theta V^*} + \modulus \tnorm{V^* - \Pi^\sigma_\Theta V^*}
    \end{align*}
    Recurrent application of $\Pi^\sigma_\Theta B$ results in
    \begin{align*}
        \tnorm{(\Pi^\sigma_\Theta B)^t V_{{\theta}_0} - \Pi^\sigma_\Theta V^*} \leq \modulus^t \tnorm{V_{{\theta}_0} - \Pi^\sigma_\Theta V^*} + \modulus \cdot \frac{1-\modulus^t}{1-\modulus} \tnorm{V^* - \Pi^\sigma_\Theta V^*}.
    \end{align*}
    Taking the limit $t\rightarrow\infty$ establishes the bound. Lastly, when the approximation scheme is fully expressive, i.e., $V^* \in \cV_\Theta$, then $\Pi^\sigma_\Theta V_* = V_*$ and the bound establishes $\tilde{V}^* = V^*$. On the other hand, if $\tilde{V}^* = V^*$ then $V^* \in \cV_\Theta$.
\end{proof}

\estimationConvergence*
\begin{comment}
\begin{app_prop*}[\ref{prop:estimation_convergence}]
    Denote the estimation error by
    \begin{align*}
        \epsilon_t := \tnorm{\tilde{V}_{t+1}-\Pi^\sigma_\Theta B\tilde{V}_t}
    \end{align*}    
    and assume $\epsilon_t \rightarrow 0$. Then, under Assumption~\ref{assum:contraction}, the sequence of estimates $\left\{ \tilde{V}_t\right\}$ converges to $\tilde{V}^*$ \eqref{eq:pvi_limit} in the $\tau$-weighted $\ell_\infty$ norm.
\end{app_prop*}
\end{comment}

\begin{proof}
    Let $\epsilon >0$. By assumption, there exists $t'=t'(\epsilon)$ s.t. $\epsilon_{t} \leq (1-\modulus) \epsilon$ for all $t \geq t'$. At iteration $t'$, we have $\tilde{V}_{t'} \in B_{\rho'}^\tau(\tilde{V}^*)$ for some ball of radius $\rho'=\rho'(\epsilon)$ w.r.t. $\tnorm{\cdot}$. Consequently, at $t'+1$, we have $\tilde{V}_{t'+1} \in \mathcal{B}^\tau_{\modulus \rho' + \epsilon_{t'}}(\tilde{V}^*)$, since 
    \begin{align*}
        \tnorm{\tilde{V}_{t'+1} - \tilde{V}^*} &= \tnorm{\tilde{V}_{t'+1} - \Pi^\sigma_\Theta B \tilde{V}_{t'} + \Pi^\sigma_\Theta B \tilde{V}_{t'} - \tilde{V}^*}\\
        & \leq \tnorm{\tilde{V}_{t'+1} - \Pi^\sigma_\Theta B \tilde{V}_{t'}} + \tnorm{\Pi^\sigma_\Theta B \tilde{V}_{t'} - \Pi^\sigma_\Theta B\tilde{V}^*}\\
        & \leq \epsilon_{t'} + \modulus \rho'.
    \end{align*}
    At the next step, we find $\tilde{V}_{t'+2} \in \mathcal{B}^\tau_{\modulus^2 \rho' + \modulus\epsilon_{t'} + \epsilon_{t'+1}}(\tilde{V}^*)$, since
    \begin{align*}
        \tnorm{\tilde{V}_{t'+2} - \tilde{V}^*} &= \tnorm{\tilde{V}_{t'+2} - \Pi^\sigma_\Theta B \tilde{V}_{t'+1} + \Pi^\sigma_\Theta B \tilde{V}_{t'+1} - \tilde{V}^*}\\
        & \leq \tnorm{\tilde{V}_{t'+2} - \Pi^\sigma_\Theta B \tilde{V}_{t'+1}} + \tnorm{\Pi^\sigma_\Theta B \tilde{V}_{t'+1} - \Pi^\sigma_\Theta B\tilde{V}^*}\\
        & \leq \epsilon_{t'+1} + \modulus \left(\epsilon_{t'} + \modulus \rho'\right).
    \end{align*}    
    and so on. Thus, we have
    \begin{align*}
        \tnorm{\tilde{V}_{T+1}-\tilde{V}^*} \leq \modulus^{T+1-t'}\rho' + \sum_{t=t'}^T \modulus^{T-t} \epsilon_t.
    \end{align*}
    Thus,
    \begin{align*}
        \tnorm{\tilde{V}_{T+1}-\tilde{V}^*} & \leq \modulus^{T+1-t'} \rho' + (1-\modulus^{T+1-t'}) \epsilon \\
        & \leq \modulus^{T+1-t'} \rho' + \epsilon
    \end{align*}
    Letting $T \rightarrow \infty$, we find that  
    \begin{align*}
        0 \leq \liminf_{T\rightarrow\infty} \tnorm{\tilde{V}_T-\tilde{V}^*} \leq \limsup_{T\rightarrow\infty} \tnorm{\tilde{V}_T-\tilde{V}^*} \leq \epsilon
    \end{align*}
    Since $\epsilon>0$ was chosen arbitrarily, $\lim_{T\rightarrow\infty} \tnorm{\tilde{V}_T-\tilde{V}^*} = 0$.
\end{proof}

\BiasVarainace*
\begin{comment}
\begin{app_lemma*}[\ref{lem:bias_var}]
    Let $\sigma \in \Delta_S$ and $p_\sigma$ and $f$
    as defined in \S\ref{sec:bias_variance}, we have
    \begin{multline}
        \E_{p_\sigma}[f(\theta,s,\xi)] = \E_{p_\sigma} \left[ \left(y(s,\xi)-B\tilde{V}(s)\right)^{2}\right] + \\ 2\E_{\sigma} \left[ \left(B\tilde{V}(s)-V_{\theta}(s)\right) \E_{\xi|s}\!\left[\left(y(s,\xi)-B\tilde{V}(s)\right)\right]\right] + \snorm{B\tilde{V}-V_{\theta}}^{2}
    \end{multline}
\end{app_lemma*}
\end{comment}

\begin{proof}
Let $\sigma \in \Delta_S$. As defined in \S\ref{sec:bias_variance}, consider $p_\sigma$ to be the distribution over $\cS \times \cS^{\cA}$ whose marginalization over $\cS^{\cA}$ produces $\sigma$ such that the probability of $\xi \in \cS^{\cA}$ mapping $a$ to $s'$ conditioned on $s \in \cS$ is $\Pr[\xi(a) = s'|s] = p(s'|s,a)$. Following the standard bias-variance decomposition for the loss function $f$ in \eqref{eq:fvi_loss}, we have
\begin{align*}
\E_{p_\sigma}[f(\theta,s,\xi)] 
    & =	\E_{p_\sigma} \left[\left(y(s,\xi)-B\tilde{V}(s)+B\tilde{V}(s)-V_{\theta}(s)\right)^{2}\right] \\
	& =	\E_{p_\sigma} \left[ \left(y(s,\xi)-B\tilde{V}(s)\right)^{2}\right] \\
    & + 2\E_{p_\sigma} \left[ \left(y(s,\xi)-B\tilde{V}(s)\right)\left(B\tilde{V}(s)-V_{\theta}(s)\right)\right] \\
    & +\E_\sigma \left[ \left(B\tilde{V}(s)-V_{\theta}(s)\right)^{2}\right] \\
	& =	\E_{p_\sigma} \left[ \left(y(s,\xi)-B\tilde{V}(s)\right)^{2}\right] \\
    & + 2\E_{\sigma} \left[\E_{\xi|s}\!\left[ \left(y(s,\xi)-B\tilde{V}(s)\right) \left(B\tilde{V}(s)-V_{\theta}(s)\right)\right] \right] \\ 
    & + \E_{\sigma} \left[\left(B\tilde{V}(s)-V_{\theta}(s)\right)^{2}\right] \\
	& =	\E_{p_\sigma} \left[ \left(y(s,\xi)-B\tilde{V}(s)\right)^{2}\right] \\
    & + 2\E_{\sigma} \left[ \left(B\tilde{V}(s)-V_{\theta}(s)\right) \E_{\xi|s}\!\left[y(s,\xi)-B\tilde{V}(s)\right]  \right] \\ 
    & + \snorm{B\tilde{V}-V_{\theta}}^{2},
\end{align*}
where ${\xi|s \sim \prod_{a \in \cA} p(\cdot|s,a)}$, as desired.
\end{proof}

\ShapiroCond*

\begin{proof}
({\romannumeral 1}) Since $S$ is a finite set, the expected value of $f_t(\theta,\cdot)$ w.r.t. any probability measure $\sigma$ over $S$ is finite for any $\theta\in \Theta$. ({\romannumeral 2}) Since $S$ is finite, $\Theta$ is compact, and $v$ is continuous in $\theta$, it follows that 
\begin{align*}
    \tilde{\omega}_t \triangleq \max_{s \in \cS} \max_{\theta\in \Theta} \abs{f_t(\theta,s)-f_t(\theta)}
\end{align*}
is finite,
and, therefore, $\abs{Y^{\theta,\theta'}_t\!(s)}\leq 2\tilde{\omega}_t$.
Since $\E_{\sigma}\left[Y^{\theta,\theta'}_t(s)\right]=0$, by Hoeffding's Lemma
\begin{align*}
    \E_{\sigma} \left[ \exp \left(
    \zeta Y^{\theta,\theta'}_t\!(s)
    \right) \right] \leq 
    \exp \left(\frac{\zeta^2(4\tilde{\omega}_t)^2}{8}\right) = \exp\left(\frac{\zeta^2(2\tilde{\omega}_t)^2}{2}\right)
\end{align*}
implying that $Y^{\theta,\theta'}_t$ is $\omega_t$-sub-Gaussian with 
\begin{align}
\label{eq:subgaussian_omega}
    \omega_t \triangleq 2\tilde{\omega}_t.
\end{align}
({\romannumeral 3}) Since $V_\theta(s)$ is Lipschitz continuous over $\Theta$ for each $s \in \cS$ \eqref{eq:L}, so is $f_t(\cdot,s)$, for some Lipschitz constant $\kappa_t(s)$. Since $S$ is finite, 
\begin{align}
\label{eq:kappa_max}
    \kappa_t^* \triangleq \max_{s \in \cS}\kappa_t(s)
\end{align}
is finite and so is $M_{\kappa_t}(\zeta)$ for all $\zeta\in\R$. ({\romannumeral 4}) This is a direct application of Cram\'er’s Large Deviation Theorem.
\end{proof}

\projectionAngle*
\begin{comment}
\begin{app_lemma*}[\ref{lem:projection_angle}]
    Assume $\cV_\Theta \subset \cV^0_S$ is convex. Then, for any $\theta \in \Theta$, we have
    \begin{align*}
        \snorm{V_{\theta} - V_{\theta_t^*}}^2 \leq f_t(\theta) - f_t^*
    \end{align*}
\end{app_lemma*}
\end{comment}
\begin{proof}
    By definition \eqref{eq:dg_loss_fn}, we have
    \begin{align*}
        f_t(\theta) = \snorm{V_{\theta} - BV_{\tilde{\theta}_t}}^2, \qquad
        f_t(\theta^*_t) = \snorm{V_{\theta^*_t} - BV_{\tilde{\theta}_t}}^2.
    \end{align*}    
    Thus, we need to show
    \begin{align*}
        \snorm{V_{\theta} - V_{\theta_t^*}}^2 &\leq \snorm{V_{\theta} - BV_{\tilde{\theta}_t}}^2 - \snorm{V_{\theta^*_t} - BV_{\tilde{\theta}_t}}^2.
    \end{align*}
    Indeed,
    \begin{align*}
    \snorm{V_{\theta} - BV_{\tilde{\theta}_t}}^2 - \snorm{V_{\theta^*_t} - BV_{\tilde{\theta}_t}}^2&=( V_{\theta}-V_{\theta^*_t})^\top\diag(\sigma)(  V_{\theta}+V_{\theta^*_t}-2BV_{\tilde{\theta}_t})\\
    &\geq ( V_{\theta}-V_{\theta^*_t})^\top\diag(\sigma)(  V_{\theta}+V_{\theta^*_t}-2V_{\theta^*_t})\\
    &= \snorm{V_{\theta}-V_{\theta^*_t}}^2,
    \end{align*}
    where the inequality follows from Lemma~\ref{lem:projection_stationarity_condition} with $\tilde{V} := BV_{\tilde{\theta}_t}$ and $\Pi^\sigma_\Theta \tilde{V} := \Pi^\sigma_\Theta BV_{\tilde{\theta}_t} \equiv V_{\theta_t^*}$.
\end{proof}

\FVI*

\begin{proof}
    Since $f_t(\cdot,s)$ is convex over $\Theta$ for any $s \in \cS$, then so is $f_{t,n_t}$ for any i.i.d. $\sigma$-sample.
    Following \cite[Theorem 9.16]{Beck2023} applied to the function $f_{t,n}$ defined over the compact and convex $\Theta$, after 
    \begin{align*}
        c_t \geq {D_\Theta^2}/{\eta_t \epsilon_t} \geq {d\left(\tilde{\theta}_t,\theta_{t,n}^*\right)^2}/{\eta_t \epsilon_t}
    \end{align*}
    projected GD steps with constant stepsize $\eta_t \in (0,1/\kappa_t^*]$, we get an $\epsilon_t/2$-optimal solution $\tilde{\theta}_{t+1}$ of $f_{t,n_t}$, which serves as our estimate for $\theta_t^*$.
    Note that $(0,1/\kappa_t^*] \subseteq (0,1/\bar{\kappa}_{t,n}]$ w.p.1 since $\Pr[\bar{\kappa}_{t,n} = \frac{1}{n} \sum_{i=1}^n \kappa_t(s_i) >\kappa_t^* = \max_{s \in \cS} \kappa_t(s)] = 0$.
    
    Lemma~\ref{lem:Shapiro_cond} establishes the necessary conditions for \cite[Theorem 5.18]{shapiro2021lectures}. At the $t$ iteration, we pick $\alpha_t \in (0,1)$ and $\hat{\kappa}_t > \E_{\sigma}[\kappa_t(s)]$ with its corresponding $\beta_t = \beta(\hat{\kappa}_t)$ as in Lemma~\ref{lem:Shapiro_cond}({\romannumeral 4}). When \cite[Theorem 5.18]{shapiro2021lectures} is applied to our $\epsilon_t/2$-optimal solution $\tilde{\theta}_{t+1}$ of $f_{t,n_t}$, using $\epsilon\equiv\epsilon_t$ and $\delta\equiv\epsilon_t/2$, we get $\Pr\left[\tilde{\theta}_{t+1} \in \Theta^*_{\epsilon_t}\right] \geq 1-\alpha_t$
    for $n_t$ bounded by 
    \begin{align}
    \label{eq:shapiro_n}
        n_t \geq \max \left\{\frac{32\omega_t^2}{\epsilon_t^{2}} \left[N \ln\left(\frac{16\nu D_{\Theta} \hat{\kappa}_t }{\epsilon_t}\right)+\ln\left(\frac{2}{\alpha_t}\right)\right],\beta_t^{-1}\ln\left(\frac{2}{\alpha_t}\right)\right\},
    \end{align}
    with $\omega_t$ given in \eqref{eq:subgaussian_omega}. Thus, 
    \begin{align*}
        & \E\left[ d\left(\tilde{\theta}_{t+1},\Theta^*_{\epsilon_t}\right) \right] \leq \Pr\left[\tilde{\theta}_{t+1} \in \Theta^*_{\epsilon_t}\right] \cdot 0 + \Pr\left[\tilde{\theta}_{t+1} \notin \Theta^*_{\epsilon_t}\right] \cdot D_{\Theta} \leq \alpha_t D_{\Theta}.
    \end{align*}
    With $\Theta^*_{\epsilon_t}$ compact, let us pick $\tilde{\theta}_{t,\epsilon_t} \in \Theta^*_{\epsilon_t}$ such that $d\left(\tilde{\theta}_{t+1},\tilde{\theta}_{t,\epsilon_t}\right)=d\left(\tilde{\theta}_{t+1},\Theta^*_{\epsilon_t}\right)$.
    Then, applying the Lipschitz constant \eqref{eq:L} and Lemma~\ref{lem:sigma_l2_tau_inf}
    \begin{align*}
        \tnorm{V_{\tilde{\theta}_{t+1}} - V_{\theta_t^*}} & \leq \tnorm{V_{\tilde{\theta}_{t+1}} - V_{\tilde{\theta}_{t,\epsilon_t}}} + \tnorm{V_{\tilde{\theta}_{t,\epsilon_t}} - V_{\theta_t^*}} \\ 
        & \leq \tnorm{L} d\left(\tilde{\theta}_{t+1},\Theta^*_{\epsilon_t}\right) + \frac{1}{c(\sigma,\tau)} \snorm{V_{\tilde{\theta}_{t,\epsilon_t}} - V_{\theta_t^*}} 
    \end{align*}
    By Lemma~\ref{lem:projection_angle} applied to $\theta \equiv \tilde{\theta}_{t,\epsilon_t}$, we know that $\snorm{V_{\tilde{\theta}_{t,\epsilon_t}} - V_{\theta_t^*}}^2 \leq f_t(\tilde{\theta}_{t,\epsilon_t}) - f_t^* \leq \epsilon_t$, and, therefore, 
    \begin{align*}
        \tnorm{V_{\tilde{\theta}_{t+1}} - V_{\theta_t^*}} & \leq \tnorm{L} d\left(\tilde{\theta}_{t+1},\Theta^*_{\epsilon_t}\right) + \frac{\sqrt{\epsilon_t}}{c(\sigma,\tau)}.
    \end{align*}
    We, thus, have, for the estimation error \eqref{eq:estimation_error},
    \begin{align*}
        \E\left[\tnorm{V_{\tilde{\theta}_{t+1}} - V_{\theta_t^*}}\right] & \leq \tnorm{L} \alpha_t D_\Theta + \frac{\sqrt{\epsilon_t}}{c(\sigma,\tau)}.
    \end{align*}
    Assigning $\alpha_t := 2\sqrt{\epsilon_t} \in (0,1)$ in \eqref{eq:shapiro_n}, we get the desired sample size bound in \eqref{eq:fvi_n}. Let us denote the resulting expected estimation error by
    \begin{align*}
        \tilde{\epsilon}_t \triangleq \sqrt{\epsilon_t}\left(2\tnorm{L} D_\Theta + \frac{1}{c(\sigma,\tau)}\right),
    \end{align*}
    which decreases asymptotically to zero as $O(\sqrt{\epsilon_t})$.
    
    Proposition~\ref{prop:estimation_convergence} can be applied in expectation as well. In particular, for the projected improvement fixed point $\tilde{V}^*$ \eqref{eq:pvi_limit}, we have, for each $t=0,\ldots,T-1$,
    \begin{align*}
        \E\left[\tnorm{V_{\tilde{\theta}_{T}} -\tilde{V}^*}\right] 
        & \leq \modulus^{T-t} \rho + \tilde{\epsilon}_t/(1-\modulus),
    \end{align*}
    where $\rho = \frac{2 R \sqrt{W} \modulus^\tau}{1-\modulus^\tau} \left( \frac{1}{\tau_\infty} - \frac{1}{\tau_0}\right)$ comes from choosing $\tilde{\theta}_0$ since $V_{\tilde{\theta}_0} \in \mathcal{B}_\rho^\tau(0)$. In particular,
    \begin{align*}
        \E\left[\tnorm{V_{\tilde{\theta}_{T}} -\tilde{V}^*}\right] 
        & \leq \min_{t=0,\ldots,T-1}\left\{\modulus^{T-t} \rho + \tilde{\epsilon}_t/(1-\modulus)\right\}
    \end{align*}    
    and, consequently, 
    \begin{align*}
        \lim_{T \rightarrow \infty} \E\left[\tnorm{V_{\tilde{\theta}_{T}} -\tilde{V}^*}\right] 
        & = 0.
    \end{align*}
\end{proof}

\VthetaStateBound*

\begin{proof} 
    We have
    \begin{align*}
        \abs{V_\theta(s) - BV_{\tilde{\theta}_t}(s)} & \leq \abs{V_\theta(s) - \tilde{V}^*(s)} + \abs{\tilde{V}^*(s) - V^*(s)} +
        \abs{V^*(s) - BV_{\tilde{\theta}_t}(s)} \\
        & \leq L(s)D_\Theta + \tau_0\tnorm{\tilde{V}^* - V^*} + \tau_0 \tnorm{BV^* - BV_{\tilde{\theta}_t}}\\
        & \leq \norm{L}_\infty D_\Theta + \tau_0\tnorm{\tilde{V}^* - V^*} + \tau_0 \modulus^\tau\tnorm{V^* - V_{\tilde{\theta}_t}}\\
        & \leq \norm{L}_\infty D_\Theta + \tau_0\tnorm{\tilde{V}^* - V^*}  + \tau_0 \modulus^\tau \left[\tnorm{V^* - \tilde{V}^*} + \tnorm{\tilde{V}^* - V_{\tilde{\theta}_t}}\right]\\
        & \leq \left(1+\frac{\tau_0 \modulus^\tau}{\tau_{d_\Pi}}\right)\norm{L}_\infty D_\Theta + \tau_0\left(1+\modulus^\tau \vphantom{\frac{\tau_0}{\tau_{d_\Pi}}}\right) \tnorm{\tilde{V}^* - V^*} \\
        & \leq \left(1+\frac{\tau_0 \modulus^\tau}{\tau_{d_\Pi}}\right)\norm{L}_\infty D_\Theta + \frac{\tau_0 \modulus \left(1+\modulus^\tau\right)}{1-\modulus} \tnorm{V^* - \Pi^\sigma_\Theta V^*},
    \end{align*}
    where the last inequality was established by Proposition~\ref{prop:pvi} and the second to last was established by \eqref{eq:L}. 
\end{proof}

\CorFVI*
\begin{comment}
\begin{app_coro*}[\ref{coro:fvi}]
    Let $\epsilon>0$. Following the global choice of constants $\omega_t \equiv \omega, \hat{\kappa}_t \equiv \hat{\kappa}, \kappa_t^* \equiv \kappa^*, \beta_t \equiv \beta$, and $\eta_t \equiv \eta \in (0,1/\kappa^*]$ in Remark~\ref{rem:fvi_consts}, while taking $\epsilon_t = {\epsilon_0}/{(1+\delta)^t}$, for $\epsilon_0 < 1/4$ and $\delta > 0$, and
    \begin{align*}
        T = \left\lceil a \log_{1+\delta}\left(\frac{\rho+\rho'}{\epsilon}\right) \right\rceil,
    \end{align*}
    for $\rho$ as in Theorem~\ref{thm:fvi} and
    \begin{align*}
        a & = - \left(\frac{2}{\log_{1+\delta}\modulus^*}\right) > 0, \quad
        \modulus^* = \max \left\{\modulus,\frac{1}{\sqrt{(1+\delta)}}\right\} < 1, \\
        \rho' &= \frac{\sqrt{\epsilon_0}}{(1-\modulus)} \left(2\tnorm{L} D_\Theta + \frac{1}{c(\sigma,\tau)}\right),
    \end{align*}
    Algorithm~\ref{alg:fvi} will output an $\epsilon$-approximation $V_{\tilde{\theta}_{T}}$ of $\tilde{V}^*$ \eqref{eq:pvi_limit} w.r.t. $\tnorm{\cdot}$ in expectation with sampling complexity
    \begin{align*}
        \sum_{t=0}^{T-1} n_t = O\left(\log_{1+\delta}\left(\frac{1}{\epsilon}\right)
        \left(\frac{1}{\epsilon}\right)^{2a}\right) 
    \end{align*}
    and first order complexity
    \begin{align*}
        \sum_{t=0}^{T-1} c_t 
        = O\left(\left(\frac{1}{\epsilon}\right)^{a}\right),
    \end{align*}        
\end{app_coro*}
\end{comment}

\begin{proof}    
    Following the choice of global constants in Remark~\ref{rem:fvi_consts}, the total number of samples drawn in Algorithm~\ref{alg:fvi} over all iterations $t=0,\ldots,T-1$ is given by
    \begin{align*}
        \sum_{t=0}^{T-1} n_t & {\leq} %= 
        \sum_{t=0}^{T-1} \frac{32 \omega^2}{\epsilon_t^{2}} \left[N \ln\left(\frac{16 \nu D_{\Theta} \hat{\kappa}}{\epsilon_t}\right) + \frac{1}{2}\ln\left(\frac{1}{\epsilon_t}\right)\right] {+1}
    \end{align*}
    Choosing $\epsilon_t = {\epsilon_0}/{(1+\delta)^t}$, with $\epsilon_0 < 1/4$ and $\delta > 0$, we get {
    \begin{align}
        \sum_{t=0}^{T-1} n_t 
        & \leq \sum_{t=0}^{T-1} (1+\delta)^{2t} \left(\tilde{a}_1 t + \tilde{a}_2\right)\nonumber\\
        &\leq (\tilde{a}_1T+\tilde{a}_2)\sum_{t=0}^{T-1}(1+\delta)^{2t}\nonumber\\
        &=(\tilde{a}_1T+\tilde{a}_2)\frac{(1+\delta)^{2(T-1)}-1}{(1+\delta)^2-1}\nonumber\\
        &\leq (\tilde{a}_1T+\tilde{a}_2)\frac{(1+\delta)^{2T-1}}{\delta},\label{eq:sum_n_t}
    \end{align}}
    where
    \begin{align*}
        \tilde{a}_1 = \frac{32 \omega^2}{\epsilon_0^{2}} \left(N+\frac{1}{2}\right)\ln\left(1+\delta\right)+1,\qquad
        \tilde{a}_2 = \frac{32 \omega^2}{\epsilon_0^{2}} N\ln\left( \frac{16 \nu D_{\Theta} \hat{\kappa}}{\epsilon_0^{3/2}} \right).
    \end{align*}
    Now, if we require the expected error \eqref{eq:fvi_error} to be bounded by $\epsilon$, then it is sufficient to require that
    \begin{align*}
        \modulus^{T/2} \rho + \frac{\sqrt{\epsilon_{T/2}}}{(1-\modulus)} \left(2\tnorm{L} D_\Theta + \frac{1}{c(\sigma,\tau)}\right) \leq \epsilon,
    \end{align*}
    where we picked $t=T/2$ (assuming for the moment $T$ is even). Applying our choice of $\epsilon_t$, we have $\modulus^{T/2} \rho + \left(\modulus'\right)^{T/2} \rho' \leq \epsilon$,
    where
    \begin{align*}
        \modulus' = \frac{1}{\sqrt{(1+\delta)}} < 1,\qquad
     \rho' = \frac{\sqrt{\epsilon_0}}{(1-\modulus)} \left(2\tnorm{L} D_\Theta + \frac{1}{c(\sigma,\tau)}\right).
    \end{align*}    
    Taking $\modulus^* = \max \{\modulus,\modulus'\} < 1$, it is sufficient to require $(\rho+\rho')\left(\modulus^*\right)^{T/2} \leq \epsilon$. When $T$ is odd, we should, similarly, require $(\rho+\rho')\left(\modulus^*\right)^{(T-1)/2} \leq \epsilon$, which implies, either way, that
    \begin{align*}
        & T \geq \log_{1+\delta}\left(\frac{\epsilon}{\rho+\rho'}\right) \left(\frac{2}{\log_{1+\delta}\modulus^*}\right) \geq 0
    \end{align*}
    Thus, taking $T = \left\lceil {a_0} \log_{1+\delta}\left(\frac{\rho+\rho'}{\epsilon}\right) \right\rceil$ for ${a_0} = - \left(\frac{2}{\log_{1+\delta}\modulus^*}\right) > 0$ guarantees $\E\left[\tnorm{V_{\tilde{\theta}_{T}} -\tilde{V}^*}\right] 
        \leq \epsilon$ for a given $\epsilon>0$. 
    
    Back to sampling complexity, applying the above choice of $T$
    to \eqref{eq:sum_n_t}, we obtain that for all sufficiently small $\epsilon$
    \begin{align*}
        (\tilde{a}_1 T + \tilde{a}_2)\frac{(1+\delta)^{2T-1}}{\delta} {\leq  \left(a_1 \log_{1+\delta}\left(\frac{\rho+\rho'}{\epsilon}\right) + a_2\right) \left(\frac{\rho+\rho'}{\epsilon}\right)^{2a_0}}
    \end{align*} 
    with
    {
    \begin{align*}
        a_1 = \frac{{a_0} \cdot \tilde{a}_1(1+\delta)}{\delta}, \qquad
        a_2 = \frac{(\tilde{a}_2+\tilde{a}_1)(1+\delta)}{\delta},
    \end{align*} }
    to get the sampling complexity as stated.
    
    The case for the total sum of projected GD steps is more straight forward. Applying the given choice of a monotonically decreasing sequence $\epsilon_t$ and total number of iterations $T$, we have
    {\begin{align*}
        \sum_{t=0}^{T-1} c_t &\leq \frac{D_\Theta^2\kappa^*}{\mu} \sum_{t=0}^{T-1} \frac{1}{\epsilon_t}+T\leq \frac{D_\Theta^2\kappa^*}{\mu\epsilon_0} \frac{(1+\delta)^T}{\delta}+T \\
        &\leq \frac{D_\Theta^2(1+\delta)}{\eta\epsilon_0\delta}\left(\frac{\rho+\rho'}{\epsilon}\right)^{a_0}+a_0\log_{1+\delta}\left(\frac{\rho+\rho'}{\epsilon}\right)+1
    \end{align*}}  
    as stated.
\end{proof}

%%%%%%%%%%%%%%%%%%%%%%%%%%%%%%%%%%%%
%%%%%%%%%%%%%%%%%%%%%%%%%%%%%%%%%%%%

\section{Examples in Detail}

%%%%%%%%%%%%%%%%%%%%%%%%%%%%%%%%%%%%
%%%%%%%%%%%%%%%%%%%%%%%%%%%%%%%%%%%%

\subsection{Traveling Salesman Problem}
\label{app:TSP}

For Theorem~\ref{thm:KH} to apply to the TSP \S\ref{sec:TSP}, we require an equivalence relation $\sim$ over $\cA^*$ that satisfies conditions ({\romannumeral 1}) and ({\romannumeral 2}). First, however, we need to consider $\sim_\Pi$ of Definition~\ref{def:simPi}. As noted, it has two distinguished classes $s_e$ \eqref{eq:state_e} and $s_\infty$ \eqref{eq:state_infinity}, where $s_\infty$ consists of all routes that are either of length $> d+1$ or correspond to illegal (partial or full) routes. By definition, it also has an additional distinguished class consisting of $\Pi$ in its entirety, which we denote by $s_{d+1}$. As in the KSP Example~\ref{ex:ksp}, we define $\sim$ by specifying equivalence classes that refine those of $\sim_\Pi$, apart from $s_e, s_{d+1}$ and $s_\infty$. 

To complete our calculation of $\sim_\Pi$, let $x,y \in \cA^* \setminus s_e \sqcup s_{d+1} \sqcup s_\infty$. If $x \sim_\Pi y$ then there exists $u \in \cA^*$ such that $xu, yu \in \Pi$ and, in particular, $\lng(x) = \lng(y) = l+1$ for $0 \leq l < d$. By definition, denoting $x=x_0\cdots x_{l}$, $y=y_0\cdots y_{l}$, and $u=u_{l+1}\cdots u_d$, we must have: (1) $x_0 = y_0 = 0$, (2) $\emptyset \subseteq \{x_1,\ldots,x_l\} = \{y_1,\ldots,y_l\} \subseteq \cA \setminus \{0\}$, and (3) $\abs{\{x_1,\ldots,x_l\}} = \abs{\{y_1,\ldots,y_l\}} = l$. One can easily see that (1)--(3) are both necessary and sufficient, for if $x$ and $y$ are such that (1)--(3) hold, then any $u$ for which $xu \in \Pi$ will satisfy $\{u_{l+1},\ldots,u_d\} = \cA \setminus \{x_1,\ldots,x_l\}$ and this will also guarantee $yu \in \Pi$. We can thus conclude that the equivalence classes of $\sim_\Pi$ that partition $\cA^* \setminus s_e \sqcup s_{d+1} \sqcup s_\infty$ are characterized by subsets $\emptyset \subseteq \cB \subseteq \cA \setminus \{0\}$. 

For $\sim$ to satisfy condition ({\romannumeral 1}) it should be preserved under right concatenation, and it should finitely refine $\sim_\Pi$. If we only care to refine classes apart from $s_e, s_{d+1}, s_\infty$, then we only need consider $x,y \in \cA^* \setminus s_e \sqcup s_{d+1} \sqcup s_\infty$ and their respective $\sim$ classes. Moreover, condition ({\romannumeral 2}) is straightforward for $x,y \in s_e$ or $x,y \in s_{d+1}$ and does not apply to $x,y \in s_\infty$. To establish condition ({\romannumeral 2}) in all other cases, let $x \sim y$, which should imply $x \sim_\Pi y$, and let $u,v \in \cA^*$ such that $xu,xv \in \Pi$. As before, $\lng(x)=\lng(y)=l+1$ for $0 \leq l < d$. Denote $x=x_0\cdots x_{l}$, $y=y_0\cdots y_{l}$, $u=u_{l+1}\cdots u_d$, and $v=v_{l+1}\cdots v_d$. We know that $x_0=y_0=u_d=v_d=0$. Then,
\begin{align*}
    g(xu;c)-g(yu;c) &= - \left(c[{0,x_1}] + \ldots + c[{x_{l},u_{l+1}}] + \ldots + c[u_{d-1},0] \vphantom{\sum}\right) \\
    & \hspace{2.6ex} + \left(c[{0,y_1}] + \ldots + c[{y_{l},u_{l+1}}] + \ldots + c[u_{d-1},0] \vphantom{\sum}\right) \\
    &= - \left(c[{0,x_1}] + \ldots + c[{x_{l},u_{l+1}}] \vphantom{\sum}\right) \\
    & \hspace{2.6ex} + \left(c[{0,y_1}] + \ldots + c[{y_{l},u_{l+1}}] \vphantom{\sum}\right).
\end{align*}
If we wish $g(xu;c)-g(yu;c) = g(xv;c)-g(yv;c)$, then, by the calculation above, we need
\begin{align}
\label{eq:3.4ii_tsp}
    c[{y_{l},u_{l+1}}] - c[{x_{l},u_{l+1}}] = c[{y_{l},v_{l+1}}] - c[{x_{l},v_{l+1}}].
\end{align}
This will hold, irrespective of $u$ and $v$, when $x_l = y_l$. Thus, to establish $\sim$, it is sufficient to finitely refine $\sim_\Pi$ by characterizing equivalence classes (apart from $s_e, s_{d+1}, s_\infty$) by pointed sets, i.e., pairs $(\cB,b)$, where $\emptyset \subsetneq \cB \subseteq \cA \setminus \{0\}$ and $b \in \cB$. The set $\cB=\emptyset$ characterizes the $\sim_\Pi$ equivalence class $[0]=\{0\}$ -- it cannot and does not need to be refined as \eqref{eq:3.4ii_tsp} holds trivially. To conclude,
\begin{align*}
    \begin{array}{l}
         x \in \cA^*\setminus s_e \sqcup s_{d+1} \sqcup s_\infty \vspace{1ex}\\
         \left[ x \right] = s_{(\cB,b)} \text{ or } s_\emptyset \vspace{1ex}\\
        \emptyset \subsetneq \cB \subseteq \cA \setminus \{0\}, b \in \cB    
    \end{array}
     \iff 
     \begin{array}{l}
          x = x_0 \cdots x_l, l = \abs{\cB} \text{ or } l=0 \vspace{1ex}\\
          x_0 = 0  \vspace{1ex}\\
          \{x_1,\ldots,x_l\} = \cB, x_l = b, l>0
     \end{array}
\end{align*}

The additive reward structure coming from Theorem~\ref{thm:KH} is defined through the help function $\tg:\cA^* \times \cC \rightarrow \R$ \eqref{eq:g_tilde_equals_g}-\eqref{eq:g_def} determined here (up to a choice of scalar), for a legal (partial or full) route, by
\begin{align*}
    \tg(x;c) &= - \sum_{j=0}^{\min\{d,l\}-1} c[{x_j, x_{j+1}}], \quad x=x_0 \cdots x_l \in \cA^*
\end{align*}
with $\tg(e;c)=\tg(a;c)=0$. By \eqref{eq:iota_def}, the incremental reward function $\iota: \cS \times \cA \times \cC \rightarrow \R$ is given for $s=s_{(\cB,b)}$ or $s_\emptyset$ by
\begin{align}
\label{eq:iota_tsp_appendix}
    \iota(s,a;c) = \tg(xa;c) - \tg(x;c) = - c[x_l,a], \quad s=[x]
\end{align}
whereas
\begin{align*}
    \iota(s_{e},a;c) = 0, \quad
    \iota(s_{d+1},a;c) = 0, \quad \iota(s_{\infty},a;c) = 0.
\end{align*}
 
For Theorem~\ref{thm:tseng_conditions} to apply, we require the equivalence relation $\sim$ over $\cA^*$ to satisfy Assumptions~\ref{assum:x_extension} and \ref{assum:s_infinity}. Both hold trivially, Assumption~\ref{assum:x_extension} since $\lng(x)=d_\Pi$ for each $x \in \Pi$ and Assumption~\ref{assum:s_infinity} by definition. The modified reward structure $r: \cS \times \cA \times \cC \rightarrow \R$ \eqref{eq:reward_modified} maintains $\iota$ \eqref{eq:iota_tsp_appendix}, apart from  $s \in \cS$ and $a \in \cA$ such that $\lambda(s,a)=s_\infty$. Thus,
\begin{align*}
    r(s,a;c) = - M \iff \begin{cases}
        s=s_e, a\neq 0\\
        s=s_\emptyset, a=0\\
        s=s_{(\cB,b)}, \cB \cup \{a\} = \cB
        \text{ or } |\cB|<d-1, a=0 \end{cases}
\end{align*}
As noted, we can pick, for example, $M = \sum_{i,j=0}^d c[i,j]>0$. 
For the approximation bound in Theorem~\ref{thm:approximation} to hold, we require Assumption~\ref{assum:partial_to_feasible}, and, indeed, for any $x=x_0\cdots x_l \in \cA^* \setminus s_e \sqcup s_{d+1} \sqcup s_\infty$, we can choose $u \in \cA^*$ to be any permutation of $\cA \setminus \{x_0,\ldots,x_l\}$ concatenated with $\text{`}0\text{'}$.

%%%%%%%%%%%%%%%%%%%%%%%%%%%%%%%%%%%%
%%%%%%%%%%%%%%%%%%%%%%%%%%%%%%%%%%%%

\subsection{Shortest Path Problem}
\label{app:SPP}

We follow the SPP DDP formulation provided in \S\ref{sec:SPP}.
By Theorem~\ref{thm:KH}, we require  any choice of an equivalence relation $\sim$ over $\cA^*$ to satisfy conditions ({\romannumeral 1}) and ({\romannumeral 2}). This means $\sim$ should be a right congruence -- if $[x]=[y]$ then $[xu]=[yu]$ for any $u\in A^*$ -- of finite rank that refines $\sim_\Pi$ and, in addition, for $x \sim y$ and $u,v \in \cA^*$ s.t. $xu,xv \in \Pi$, we require
\begin{align*}
    g(xu;c)-g(yu;c) = g(xv;c)-g(yv;c), \quad \forall c \in \cC,
\end{align*}
for $g$ defined in \eqref{eq:g_spp}. We can calculate directly what this means in terms of $\sim$. Since $\sim$ refines $\sim_\Pi$, if $x,y \in s_\infty$ condition ({\romannumeral 2}) holds vacuously, since there are no $u,v$ s.t. $xu, xv \in \Pi$. If $x,y \in s_e$ then $x=y=e$ and condition ({\romannumeral 2}) holds trivially with zero on both sides. If $x,y \in \cA^* \setminus s_e \sqcup s_\infty$ and $\lng(x) = \lng(y) = d$, then the only $u,v$ for which $xu,xv \in \Pi$ is $u=v=e$ and condition ({\romannumeral 2}) again holds trivially. We are left with the only non-trivial case: $x,y \in \cA^* \setminus s_e \sqcup s_\infty$, $\lng(x) = \lng(y) = l$ and $l < d$. Denote $x = x_1 \cdots x_l$ and $y = y_1 \cdots y_l$ and pick $u = u_{l+1} \cdots u_{l'} \in \cA^*$ so that $l+1 \leq l' \leq d$. Then, as in \S\ref{sec:TSP},
\begin{align*}
  g(xu;c) - g(yu;c)
  &= -\Big( c[\vin,x_1] + \cdots + c[x_\ell,u_{\ell+1}] \Big) \\
  &\quad + \Big( c[\vin,y_1] + \cdots + c[y_\ell,u_{\ell+1}] \Big).
\end{align*}
For condition ({\romannumeral 2}) to hold, we will need
\begin{align*}
    c[{y_l,u_{l+1}}] - c[{x_l,u_{l+1}}] = c[{y_l,v_{l+1}}] - c[{x_l,v_{l+1}}]
\end{align*}
irrespective of $u,v$. This will be guaranteed iff $x_l = y_l$. The coarsest finite rank refinement to guarantee condition ({\romannumeral 2}) thus has the following equivalence classes, on top of $s_e$ and $s_\infty$,
\begin{align*}
    \begin{array}{l}
         x \in \cA^*\setminus s_e \sqcup s_\infty \vspace{1ex}\\
         \left[ x \right] = s_{(l,a)} \text{ or } s_d \vspace{1ex}\\
        0 < l < d, a \in \cA
    \end{array}
     \iff 
     \begin{array}{l}
          x = x_1 \cdots x_l \vspace{1ex}\\
          0< l < d, x_l=a \text{ or } l=d, x_l=\vter \vspace{1ex}\\
          \\
     \end{array}
\end{align*}
and it is clearly a right congruence. 

%%%%%%%%%%%%%%%%%%%%%%%%%%%%%%%%%%%%
%%%%%%%%%%%%%%%%%%%%%%%%%%%%%%%%%%%%

\section{PVI Experiment in Detail}
\label{app:experiments}

\begin{table}[t]
\centering
\footnotesize
\setlength{\tabcolsep}{6pt}
\renewcommand{\arraystretch}{1.15}
\begin{tabular}{llcccc}
\toprule
\textbf{COP} & \textbf{d} & \textbf{K Low} & \textbf{K High} & $\bm{n_{\text{low}}}\,/\,\bm{n_{\text{high}}}$ & \textbf{PS} \\
\midrule
\multirow{3}{*}{KSP} & 10 & 5  & 10 & $112007\,/\,93434$                        & 0.508 \\
\cmidrule(lr){2-6}
                     & 14 & 7  & 14 & $113534\,/\,100020$ & 0.513 \\
\cmidrule(lr){2-6}
                     & 18 & 9  & 18 & $110795\,/\,95459$ & 0.493 \\
\midrule
\multirow{3}{*}{TSP} &  8 & 4  &  8 & $118990\,/\,111646$ &                           0.499 \\
\cmidrule(lr){2-6}
                     & 10 & 5  & 10 & $123548\,/\,122058$ & 0.498 \\
\cmidrule(lr){2-6}
                     & 12 & 6  & 12 & $124227\,/\,124636$ & 0.495 \\
\bottomrule
\end{tabular}
\vspace{1ex}
\caption{Probability of Superiority (PS) for the relative optimality gap comparing higher $K$ to the baseline lower $K$ for each $(\mathrm{COP}, d)$. Entries are point estimates. Sample sizes vary by after filtering to contractive PVI runs ($\gamma<1$) out of a total of $125,000$; the column $n_{\text{low}}/n_{\text{high}}$ reports the respective group sizes. PS equals the normalized Mann--Whitney $U$ (ties receive half credit).}
\label{tab:experiment_optimality_gap}
\end{table}

First, we spell out the specifics of instance generation $(\mathrm{COP},d,K,\sigma,\Phi,\tau,\theta_0)$ for the PVI experiments with the affine approximation scheme \S\ref{sec:pvi_experiments}. We generated $50$ KSPs \eqref{eq:KSP} with a single constraint ($m=1$) by sampling coefficients $c_j \sim \cN\left({1},4\right)$ (thus allowing for some negative values), $w_{1j} \sim \mathrm{Poisson}(1)$, and $b_1 \sim \mathrm{Poisson}(d)$, while restricting to $w_1>0$. In each dimension, we assumed $n=d$. We generated $50$ TSPs \S\ref{sec:TSP} by sampling $d$ points uniformly in $[0,1]^2$. 

For all COPs, we made the following sampling choices for the state distribution $\sigma$, the affine embedding matrix $\Phi$, the weights $\tau$, and the initial $\theta_0$: $\sigma$ was sampled uniformly from $\{(\sigma_s)_{s \in \cS \setminus \{s_\infty\}} \mid \sigma_s \geq 0,\, \sum_{s \in \cS \setminus \{s_\infty\}} \sigma_s = 1\}$; the state embedding $\phi(s)$ for $s \in \cS \setminus \{s_\infty\}$ was sampled uniformly in $[-1,1]^K \subset \R^K$; for the weights $\tau_l$, $l \in <d_\Pi+2> = \Set{0,\ldots,d_\Pi+1}$, we first sampled $\{\tau'_l\}_{l=0}^{d_\Pi+1}$ uniformly from $\{(\tau'_l)_{l \in <d_\Pi+2>} \mid \tau'_l > 0,\, \sum_{l=0}^{d_\Pi+1} \tau'_l = 1\}$ and then set $\tau_l = \sum_{l'=l}^{d_\Pi+1} \tau'_{l'}$; finally, the entries $(\theta_0^1,\ldots,\theta_0^{K-1})$ of the initial $\theta_{0}$ were sampled uniformly in $[-1,1]$.

Recall that we let $\chi$ be the random variable defined by the chance to sample contractive $(\Phi,\tau,\theta_0)$ per instance $(\mathrm{COP},d,K,\sigma)$. As a generative description of $\chi$, we assessed a Beta--binomial model using randomized probability integral transforms with the Kolmogorov-Smirnov (KS), Cramer-von Mises (CvM), and Anderson–Darling (AD) empirical distribution function tests and a parametric bootstrap with refitting. With $n=2500$, the tests rejected the model for KSP despite visually small deviations (e.g., for (KSP, $10$, $5$) KS $=0.022$, $p=0.003$; CvM $0.352$, $p<10^{-3}$; AD $=2.102$, $p<10^{-3}$) but accepted it for TSP.\footnote{We therefore chose to present non-parametric summaries (Mean with CI, Min, Mode$_\text{KDE}$ with CI) without imposing a parametric form.}

As noted, we did not witness a decisive benefit for higher embedding dimensions $K$ in terms of the relative optimality gap, defined via
\begin{align}
\label{eq:rel_opt_gap}
    \abs{(g(x^*)-g(x_{t_*}))/g(x^*)}
\end{align}
calculated for the approximate solution $x_{t_*}$ derived from $V_{\theta_{t_*}}$ for each contractive PVI run. This insight is captured in the Probability of Superiority (PS) for the relative optimality gap comparing the higher dimension embedding $K$ against the lower $K$ baseline. Here $\displaystyle \text{PS}=\Pr(X_{K\,\text{high}}<X_{K\,\text{low}})+\tfrac12\Pr(X_{K\,\text{high}}=X_{K\,\text{low}})$ (so $\text{PS}>0.5$ favors higher $K$), which is equivalent to the normalized Mann--Whitney $U$. In our case, $X_{K\,\text{high}}$ and $X_{K\,\text{low}}$ designate the r.v.s of the relative optimality gap for high and low $K$ values respectively. To begin with, we never expected the affine approximation scheme to provide quality solution approximates. Yet, we hypothesize that a decisive factor for $X_K$ would be $O(\abs{\cS}/K)$, which, in our experiments, is identical for both high and low choices of $K$.

% KSP is weakly NP-hard, while TSP is NP-hard. We witness an echo of this difference in the quality of solutions we get using the affine approximation scheme.

As a validation of the per–run contraction modulus $\gamma$ in~\eqref{eq:gamma_experiment}, we evaluated the slack of the residual bound from Proposition~\ref{prop:pvi}. For each contractive PVI run ($\gamma<1$) we computed the slack
\begin{align}
    \mathrm{Slack} = \;\frac{\gamma}{1-\gamma} - \frac{\tnorm{{V_{\theta_{T-1}}}-\Pi^\sigma_\Theta V^*}}{\tnorm{V^*-\Pi^\sigma_\Theta V^*}}
\label{eq:empirical_slack}
\end{align}
as implied by~\eqref{eq:proj_vi_rate}. Figure~\ref{fig:experiment_proposition} shows, for each $(\mathrm{COP},d)$, the histogram of $\mathrm{Slack}$ values (right $5\%$ trimmed). Across all panels we observe $\mathrm{Slack} \approx 0$ with mass near zero, indicating that the bound is respected and often tight, and supporting the correctness of our $\gamma$ calculation \eqref{eq:gamma_experiment}.

\begin{figure}[t]
\centering

% -------- Row 1 --------
\subfloat{
  \includegraphics[width=0.32\linewidth]{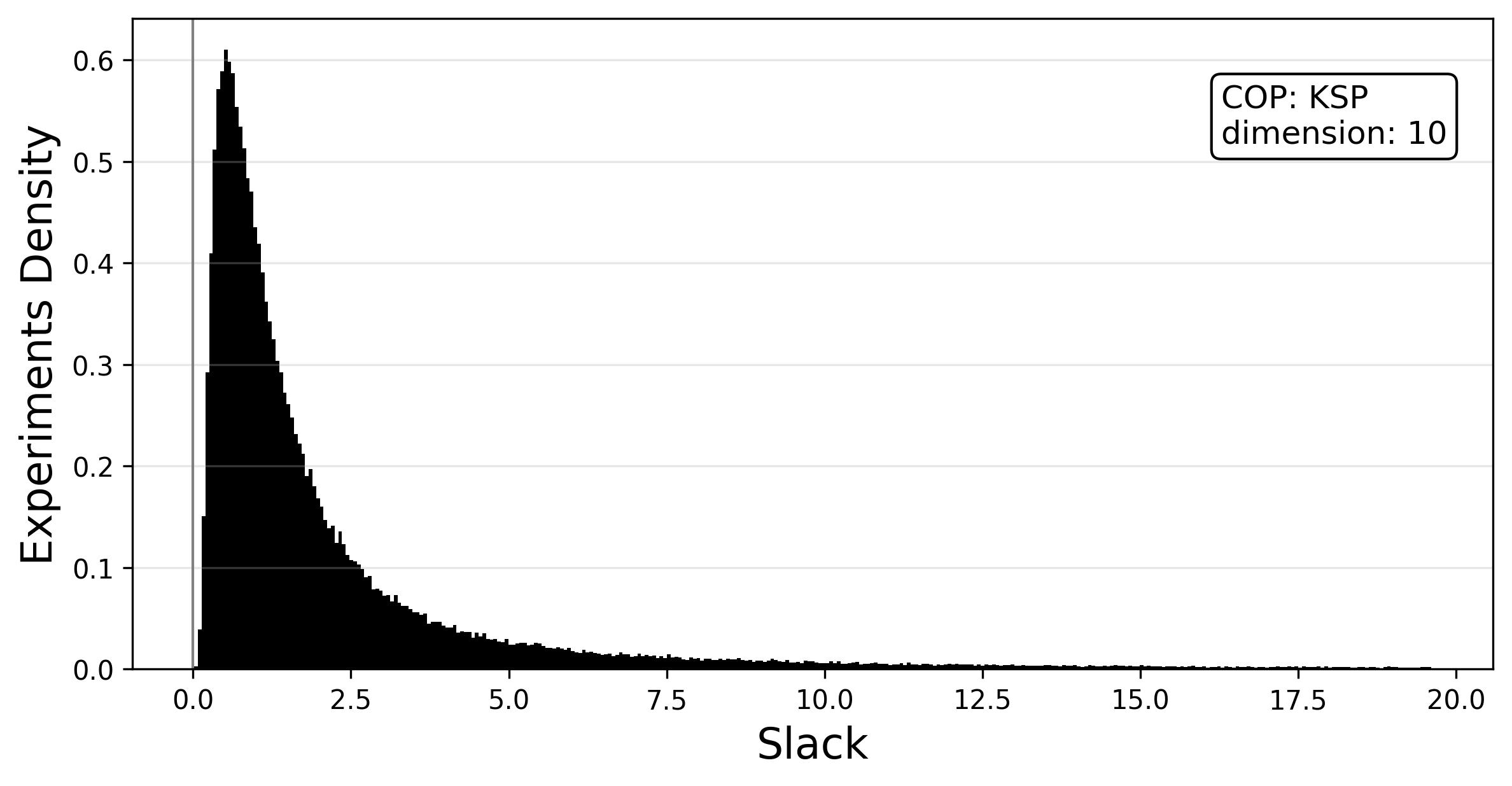}
}\hfill
\subfloat{
  \includegraphics[width=0.32\linewidth]{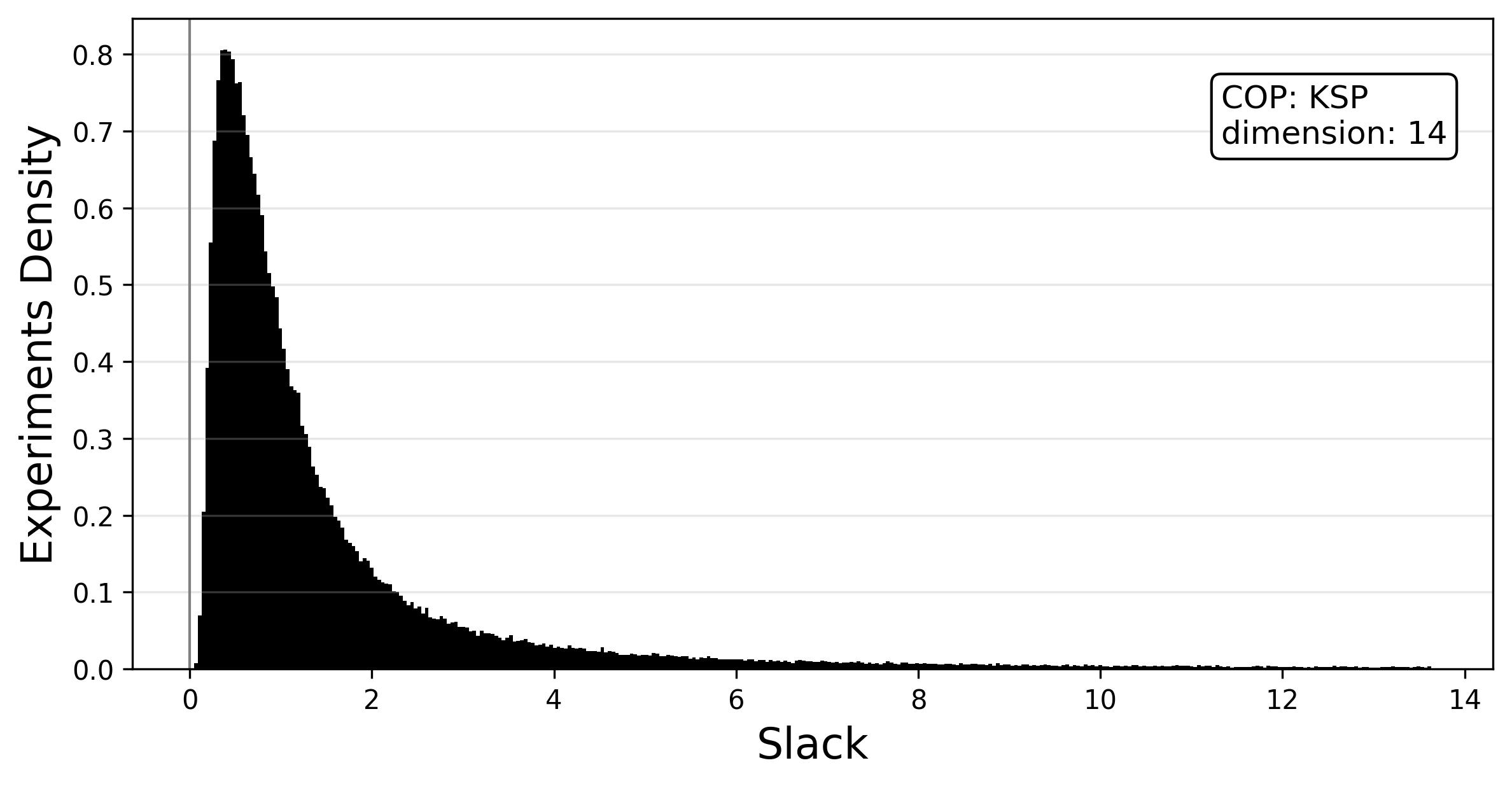}
}\hfill
\subfloat{
  \includegraphics[width=0.32\linewidth]{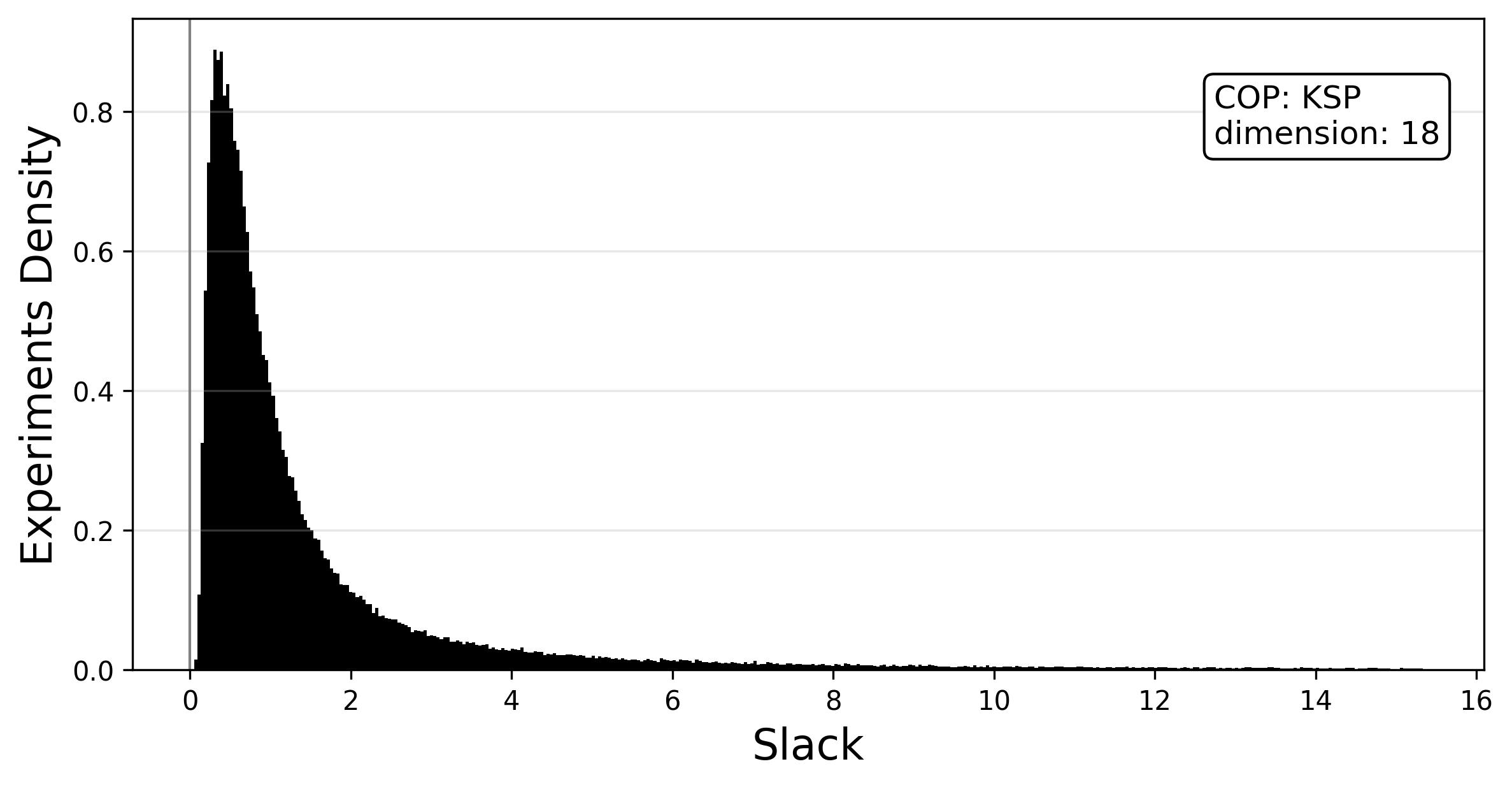}
}

\vspace{0.6em}

% -------- Row 2 --------
\subfloat{
  \includegraphics[width=0.32\linewidth]{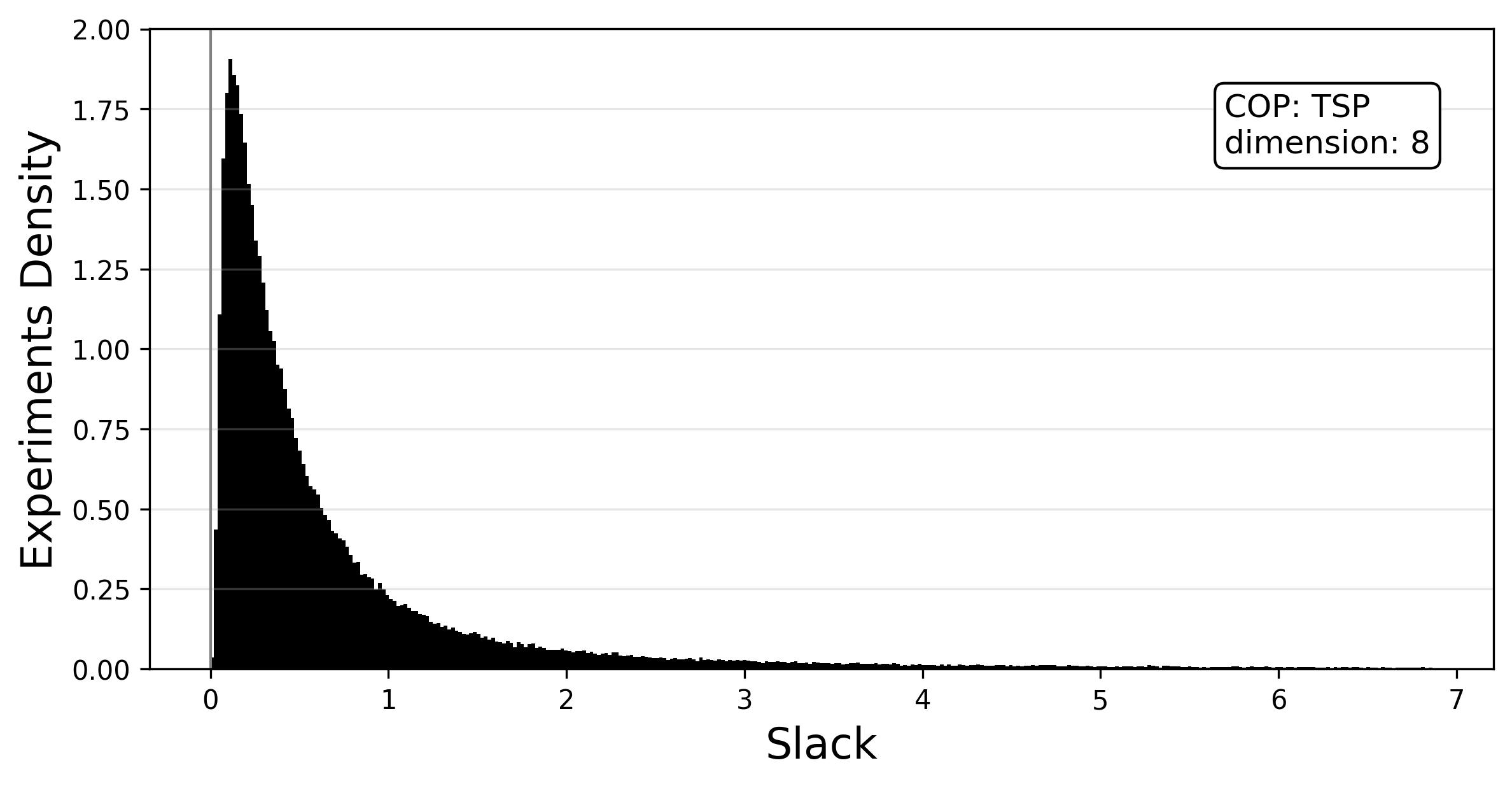}
}\hfill
\subfloat{
  \includegraphics[width=0.32\linewidth]{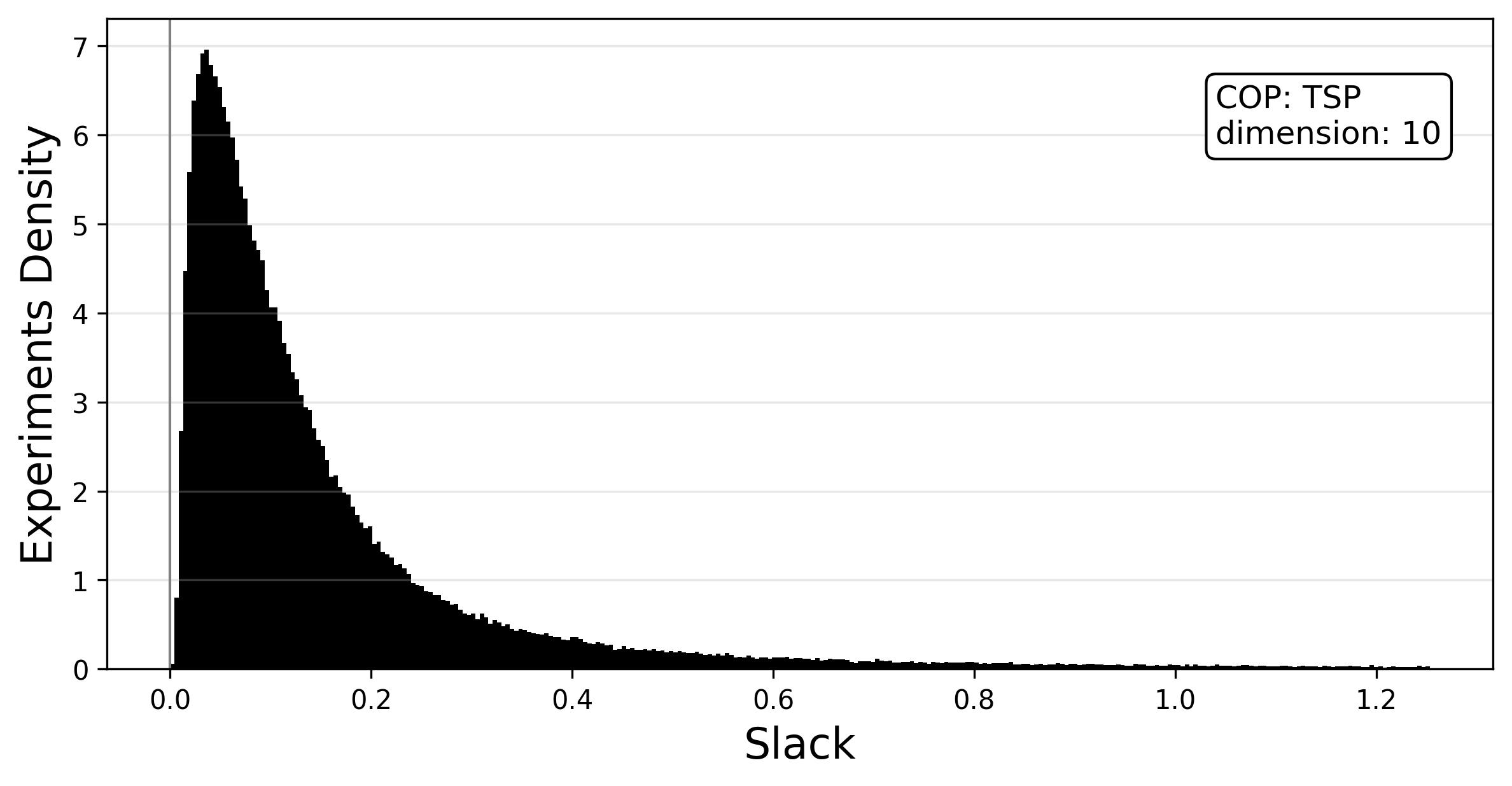}
}\hfill
\subfloat{
  \includegraphics[width=0.32\linewidth]{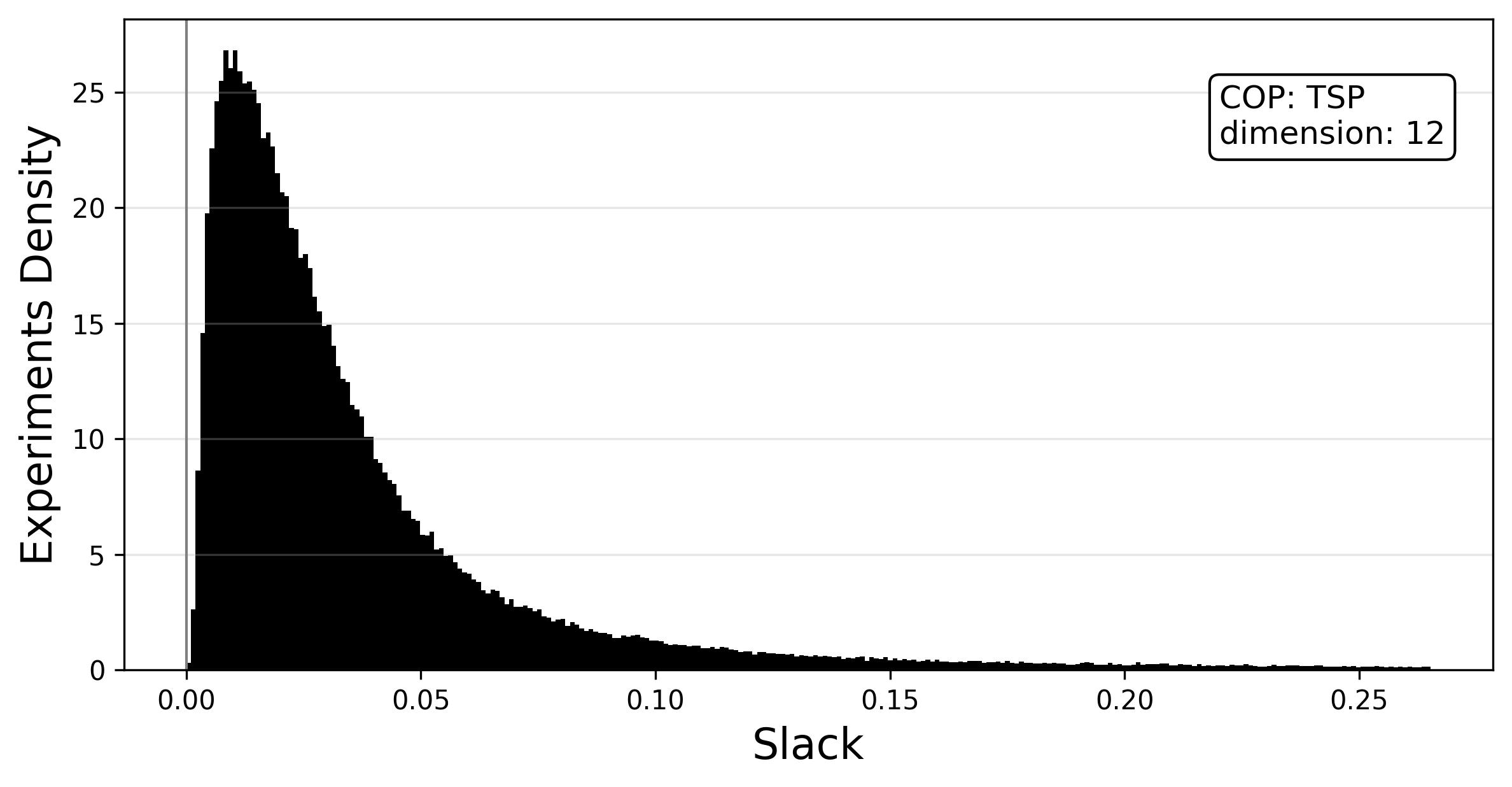}
}

\caption{Each panel corresponds to a single $(\mathrm{COP},d)$ instance and summarizes all \emph{contractive} PVI runs ($\gamma<1$). Shown is the histogram of $\mathrm{Res}$ \eqref{eq:empirical_slack}
derived from Proposition~\ref{prop:pvi}; the $y$-axis reports relative frequency. The lower $5\%$ of values (right tail) were trimmed prior to plotting. Group sizes for each setting appear in Table~\ref{tab:experiment_optimality_gap}. TSP12 statistics are derived from 30 COPs per $K$.}
\label{fig:experiment_proposition}
\end{figure}

\end{document}